%% file: neurips_2025.tex
\documentclass{article}

\PassOptionsToPackage{numbers, compress,sort}{natbib}

\usepackage[final]{neurips_2025}

\usepackage[utf8]{inputenc} 
\usepackage[T1]{fontenc}    
\usepackage{hyperref}       
\usepackage{url}            
\usepackage{booktabs}       
\usepackage{amsfonts}       
\usepackage{nicefrac}       
\usepackage{microtype}      
\usepackage{xcolor}         
\input{packagesImports}

\input{definitions}
\usepackage[capitalize,noabbrev]{cleveref}

\title{CHiQPM: Calibrated Hierarchical Interpretable Image Classification}

\author{Thomas Norrenbrock, Timo Kaiser \& Bodo Rosenhahn  \\
 Institute for Information Processing (tnt)\\
  L3S - Leibniz Universität Hannover, Germany \\
\texttt{\{norrenbr,kaiser,rosenhahn\}@tnt.uni-hannover.de} \\
\And
Sovan Biswas \& Neslihan Kose\\
Intel  Labs, Germany  \\
\texttt{\{sovan.biswas,neslihan.kose.cihangir\}@intel.com} \\
\And
Ramesh Manuvinakurike \\
Intel  Labs, USA  \\
\texttt{ramesh.manuvinakurike@intel.com} \\
}

\begin{document}

\maketitle

\input{secs/abstract}
\input{secs/intro}
\input{secs/relatedWork}
\input{secs/method}

\input{secs/results}
\input{secs/conclusion}

\FloatBarrier
\subsection*{Acknowledgments}
This work was supported by the Federal Ministry of Education and Research (BMBF), Germany, under the AI service center KISSKI (grant no. 01IS22093C), the Deutsche Forschungsgemeinschaft (DFG) under Germany’s Excellence Strategy within the Cluster of Excellence PhoenixD (EXC2122), 
the MWK of Lower Sachsony within Hybrint (VWZN4219),
the European Union  under grant agreement no. 101136006 – XTREME.
The work has been done in collaboration and partially funded by the Intel Corporation.
This work was partially supported by the German Federal Ministry of the Environment, Nature Conservation, Nuclear Safety and Consumer Protection (GreenAutoML4FAS project no. 67KI32007A).
\bibliography{iclr2025_conference}
\bibliographystyle{icml2025}

\FloatBarrier
\newpage
\input{secs/checklist}
\FloatBarrier
\newpage
\input{secs/appendix}


\end{document}

%% file: packagesImports.tex
\usepackage[utf8]{inputenc} 
\usepackage[T1]{fontenc}    
\usepackage{url}            
\usepackage{booktabs}       
\usepackage{amsfonts}       
\usepackage{nicefrac}       
\usepackage{microtype}      
\usepackage{amssymb}
\usepackage{pifont}
\usepackage{tikz}
\usetikzlibrary{shapes.geometric, arrows, fit,positioning, fit,calc}
\usepackage{algorithm}
\usepackage{placeins}
\usepackage{algorithmic}
\usepackage[algo2e]{algorithm2e} 
\usepackage{times}
\usepackage{epsfig}
\usepackage{graphicx}
\usepackage{amsmath}
\usepackage{numprint}
\usepackage{amssymb}
\usepackage{rotating}
\usepackage{makecell}
\usepackage{svg}
\usepackage{lineno}
\usepackage{multirow}
\usepackage{wrapfig}
\usepackage{smartdiagram}
\usepackage{csquotes}
\usepackage{caption}
\usepackage{arydshln}
\usepackage{subcaption}
\usepackage{rotating}

\usepackage[acronym,nogroupskip]{glossaries}
\glsdisablehyper

%% file: definitions.tex
\newcommand{\newt}[1]{#1}
\newcommand{\cindex}{\ensuremath{ c}} 
\newcommand{\findex}{\ensuremath{d}} 
\newcommand{\cmark}{\textcolor{green}{\ding{51}}}%
\newcommand{\xmark}{\textcolor{red}{\ding{55}}}%

\newcommand{\stanfordheader}{\gls{stanfordheader}}
\newcommand{\cubheader}{\gls{cubheader}}

\newcommand{\imgnetheader}{\gls{imgnetheader}}

\newglossaryentry{customLoss}
{
name={\ensuremath{\mathcal{L}_{\mathrm{div}}}},
description={Custom Loss für unterschiedliche Features}
}
\newglossaryentry{cLW}
{
name={\ensuremath{\beta}},
description={Gewichtung für customLoss}
}
\newglossaryentry{elaWeight}
{
name={\ensuremath{\lambda}},
description={Gewichtung für elasticNet}
}
\newglossaryentry{elaW}
{
name={\ensuremath{\alpha}},
long = {\ensuremath{\alpha\in[0,1]}},
description={Gewichtung zwischen l1 und l2 für elasticNet}
}
\newglossaryentry{trainDataset}
{
name={\ensuremath{\boldsymbol{D}_t}},
first = {\ensuremath{\boldsymbol{D}_t\in \mathbb{R}^{\gls{nTrainImages}\times 3\times w\times h}}},
description={LokalisierungMaps mit Missingness}
}
\newglossaryentry{nFeatures}
{
name={\ensuremath{n_{f}}},
description={Anzahl der verwendeten Features}
}
\newglossaryentry{nReducedFeatures}
{
name={\ensuremath{\gls{nFeatures}^*}},
description={Anzahl der verwendeten Features im Sparse Decision Layer}
}
\newglossaryentry{outputVector}
{
name={\ensuremath{\boldsymbol{y}}},
long = {\ensuremath{\boldsymbol{y}\in \mathbb{R}^{\gls{nClasses}}}},
description={Finaler Ausgang des Netzes}
}
\newglossaryentry{features}
{
name={\ensuremath{\boldsymbol{f}}},
description={Aus Bild berechnete Features}
}
\newcommand{\ours}{\textbf{Ours}}
\newcommand{\tabOurs}{\hqpm{} (\ours)}
\newglossaryentry{LocalizationMaps}
{
name={\ensuremath{\boldsymbol{L}}},
long = {\ensuremath{\boldsymbol{L}_p}\in \mathbb{R}^{\gls{nReducedFeatures}\times \frac{w}{p}\times \frac{h}{p}}},
description={LokalisierungMaps mit Missingness}
}
\newglossaryentry{featuresMapwidth}
{
name={\ensuremath{ w_M}},
description={Weite der FeatureMap}
}
\newglossaryentry{featuresMapheigth}
{
name={\ensuremath{ h_M}},
description={Weite der FeatureMap}
}
\newglossaryentry{featureMaps}
{
first ={\ensuremath{\boldsymbol{M} \in \mathbb{R}^{\gls{nFeatures} \times \gls{featuresMapwidth}\times \gls{featuresMapheigth}}}},
name={\ensuremath{m}},
plural={\ensuremath{\boldsymbol{M}}},
description={Aus Bild berechnete Features}
}
\newglossaryentry{featureMapsSmall}
{
first ={\ensuremath{\boldsymbol{M} \in \mathbb{R}^{\gls{nReducedFeatures} \times \gls{featuresMapwidth}\times \gls{featuresMapheigth}}}},
name={\ensuremath{m}},
plural={\ensuremath{\boldsymbol{M}}},
description={Aus Bild berechnete Features}
}
\newcommand{\activeMean}{\ensuremath{\mu_2^\findex}}
\newglossaryentry{singleMap}
{
first ={\ensuremath{\boldsymbol{M}^\findex \in \mathbb{R}^{\gls{featuresMapwidth}\times \gls{featuresMapheigth}}}},
name={\ensuremath{\boldsymbol{M}^\findex }},
plural={\ensuremath{\boldsymbol{M}}},
description={Aus Bild berechnete Features}
}
\newglossaryentry{featureMapsTrain}
{
first ={\ensuremath{\boldsymbol{M}_\mathrm{train} \in \mathbb{R}^{\gls{nTrainImages}\times\gls{nFeatures} \times \gls{featuresMapwidth}\times \gls{featuresMapheigth}}}},
name={\ensuremath{m}},
plural={\ensuremath{\boldsymbol{M}}_\mathrm{train}},
description={Aus Bild berechnete Features}
}
\newglossaryentry{trainFeatures}
{
first ={\ensuremath{\boldsymbol{F} \in \mathbb{R}^{\gls{nTrainImages} \times \gls{nFeatures}}}}, 
name={\ensuremath{\boldsymbol{F}^{\mathrm{train}}}},
plural = {\ensuremath{F^{\mathrm{train}}}},
description={Aus Bild berechnete Features}
}
\newglossaryentry{denseNet}
{
first = {DenseNet121~\citep{huang2017densely}},
name ={DenseNet121},
description={Final layer in the neural network}
}
\newglossaryentry{resNet}
{
first = {Resnet50~\citep{he2016deep}},
name ={Resnet50},
description={Final layer in the neural network}
}
\newglossaryentry{incv}
{
first = {Inception-v3~\citep{szegedy2016rethinking}},
name ={Inception-v3},
description={Final layer in the neural network}
}
\newglossaryentry{birdsheader}
{
first = {NABirds~\citep{7298658}},
name = {NABirds},
description={Final layer in the neural network}
}
\newglossaryentry{fgvcheader}
{
first = {FGVC-Aircraft~\citep{FGVCAircraft}},
name={FGVC-Aircraft},
description={Final layer in the neural network}
}
\newglossaryentry{stanfordheader}
{
first = {Stanford Cars~\citep{StanfordCars}},
name={Stanford Cars},
description={Final layer in the neural network}
}
\newglossaryentry{cubheader}
{
first = {CUB-2011~\citep{wah2011caltech}},
long = {CUB-2011~\citep{wah2011caltech}},
name={CUB-2011},
description={Final layer in the neural network}
}
\newglossaryentry{travelingheader}
{
first = {TravelingBirds~\citep{koh2020concept}},
long = {TravelingBirds~\citep{koh2020concept}},
name={TravelingBirds},
description={Final layer in the neural network}
}
\newglossaryentry{decisionLayer}
{
name={decision layer},
description={Final layer in the neural network}
}
\newglossaryentry{fittingLossTarget}
{
name={\ensuremath{\mathcal{L}_{\mathrm{target}}}},
description={Main goal of fitting}
}
\newglossaryentry{layerName}
{
name ={\textit{SLDD-Model}},
first ={\textit{SLDD-Model}~\citep{norrenbrocktake}},
description={The proposed benchmark}
}
\newglossaryentry{NewlayerName}
{
name ={QPM},
description={The proposed benchmark}
}
\newcommand{\hqpm}{CHiQPM}
\newcommand{\indicator}[2]{\ensuremath{\delta_{#1}^{#2}}}
\newcommand{\floss}{\ensuremath{\mathcal{L}_{\mathrm{feat}}}}

\newcommand{\confDots}{The stars denote different $\alpha\in\{0.12,0.1,0.075,0.05\}$ used for calibration.}

\newcommand{\gtns}{GTE}

\newcommand{\upscore}{\ensuremath{s_\mathrm{up}}}
\newcommand{\finscore}{\ensuremath{s_\mathrm{sel}}}
\newcommand{\adalvl}{\ensuremath{n^\mathrm{limit}}}

\newcommand{\predclass}{\ensuremath{\hat{c}}}
\newcommand{\allClassSet}{\ensuremath{\mathbb{C}}}
\newcommand{\classfeatures}{\ensuremath{\mathbb{F}}}
\newcommand{\orderedClassFeatures}{\ensuremath{\hat{\classfeatures}}}
\newcommand{\predset}{\ensuremath{\mathbb{Y}}}
\newcommand{\setmetricsim}{Set Coherence}
\newcommand{\datacal}{\ensuremath{\mathbb{D}_\mathrm{cal}}}
\newcommand{\noclassfeatures}{\ensuremath{\overline{\classfeatures}}}

\newcommand{\ccsim}{\ensuremath{\textbf{K}}}
\newcommand{\treedense}{\ensuremath{\rho}}
\newcommand{\pairsTarget}{\ensuremath{\mathbb{K}}}

\newcommand{\CpairsSet}{\ensuremath{\mathbb{P}}}
\newcommand{\iterPairs}{\ensuremath{\mathbb{T}}}
\newcommand{\achieved}{\ensuremath{m}}
\newcommand{\muf}{\ensuremath{\boldsymbol{\mu}^{\gls{nReducedFeatures}}}}
\newcommand{\sigmaf}{\ensuremath{\boldsymbol{\sigma}^{\gls{nReducedFeatures}}}}

\newglossaryentry{denseLayer}
{
name={{dense high-dimensional \gls{decisionLayer}}},
description={The layer that results from training}
}
\newglossaryentry{correlationMatrix}
{
name={\ensuremath{\boldsymbol{Q}}},
first = {\ensuremath{\boldsymbol{Q}\in \mathbb{R}^{\gls{nFeatures}\times\gls{nFeatures}}}},
long = {\ensuremath{q}},
description={Correlation Matrix}
}
\newglossaryentry{featureVector}
{
name={\ensuremath{f}},
long={\ensuremath{\boldsymbol{f}}},
first ={\ensuremath{\boldsymbol{f} \in \mathbb{R}^{\gls{nFeatures}}}},
description={The features of the dense alyer}
}
\newglossaryentry{RedfeatureVector}
{
name={\ensuremath{\boldsymbol{f^*}}},
first ={\ensuremath{\boldsymbol{f^*} \in \mathbb{R}^{\gls{nReducedFeatures}}}},
description={The selected feature Vector}
}
\newglossaryentry{OnlyInteractionVector}
{
name={\ensuremath{\boldsymbol{P}}},
long={\ensuremath{\boldsymbol{P} \in \mathbb{R}^{ \gls{nInteractions}}}},
description={The Interaction Vector}
}
\newglossaryentry{InteractionVector}
{
name={\ensuremath{\boldsymbol{f^*_{\phi}}}},
first ={\ensuremath{\boldsymbol{f^*_{\phi}} \in \mathbb{R}^{\gls{nReducedFeatures} + \gls{nInteractions}}}},
description={The Extended Interaction Vector}
}
\newglossaryentry{nInteractions}
{
name={\ensuremath{n_I}},
description={number of interaction term}
}
\newglossaryentry{dnn}
{
name={\ensuremath{f_\theta(x})},
description={Deep neural network}
}

\newglossaryentry{bias}
{
name={\ensuremath{\boldsymbol{b}}},
long ={\ensuremath{\boldsymbol{b} \in \mathbb{R}^{\gls{nClasses}}}},
description={The bias in the decison layer}
}
\newglossaryentry{classifyFunc}
{
name={\ensuremath{C}},
description={The classifier on the featuers}
}
\newglossaryentry{WeightMatrix}
{
name={\ensuremath{\boldsymbol{W}}},
long = {\ensuremath{\boldsymbol{W}\in \mathbb{R}^{\gls{nClasses}\times \gls{nReducedFeatures} }}},
plural = {\ensuremath{w}},
description={The Weight matrix in the decision layer}
}
\newglossaryentry{qWeightMatrix}
{
name={\ensuremath{\boldsymbol{W}^{Q}}},
first ={\ensuremath{\boldsymbol{W}^{Q}\in \{-\alpha,0,\alpha\}^{\gls{nClasses}\times \gls{nReducedFeatures} }}},
long = {\ensuremath{\boldsymbol{W}^{Q}\in \{-\alpha,0,\alpha\}^{\gls{nClasses}\times \gls{nReducedFeatures} }}},
plural = {\ensuremath{w}},
description={The Weight matrix in the decision layer}
}

\newglossaryentry{nClasses}
{
name={\ensuremath{n_c}},
description={Number of Classes}
}
\newglossaryentry{nTrainImages}
{
name={\ensuremath{n_T}},
description={Number of Train Images}
}
\newglossaryentry{nWeights}
{
name={\ensuremath{n_w}},
description={Number of Entries != 0 in \gls{WeightMatrix}}
}
\newglossaryentry{nperClass}
{
name={\ensuremath{n_{\mathrm{wc}}}},
description={Number of Entries != 0 in \gls{WeightMatrix} per Class}
}
\newglossaryentry{interpTrans}
{
name={\ensuremath{\phi}},
description={Interpretable Transformation}
}
\newglossaryentry{quantizedValue}
{
name={\ensuremath{\alpha}},
description={Interpretable Transformation}
}
\newglossaryentry{targetVector}
{
name={\ensuremath{\hat{\boldsymbol{y}}}},
description={Target Vector in Training}
}
\newglossaryentry{glmsaga}
{
first = {\mbox{\textit{glm-saga}~\citep{wong2021leveraging}}},
long = {\mbox{\textit{glm-saga}~\citep{wong2021leveraging}}},
name={\mbox{\textit{glm-saga}}},
description={Target Vector in Training}
}
\newglossaryentry{cbm}
{
name={CBM},
first ={\textit{Concept Bottleneck Model}~(CBM)~\citep{koh2020concept}},
long = {CBM~\citep{koh2020concept}},
description={Target Vector in Training}
}
\newglossaryentry{ProtoPNet}
{
name={\textit{ProtoPNet}},
first ={\textit{ProtoPNet}~\citep{chen2019looks}},
description={Target Vector in Training}
}
\newglossaryentry{ProtoPShare}
{
name={\textit{ProtoPShare}},
first ={\textit{ProtoPShare}~\citep{rymarczyk2021protopshare}},
description={Target Vector in Training}
}
\definecolor{myGreen}{RGB}{34, 139, 34}
\newglossaryentry{ProtoPool}
{
name={\textit{ProtoPool}},
first ={\textit{ProtoPool}~\citep{rymarczyk2022interpretable}},
long ={\textit{ProtoPool}~\citep{rymarczyk2022interpretable}},
description={Target Vector in Training}
}
\newcommand{\pipnettable}{PIP-Net} 

\newcommand{\protopooltable}{ProtoPool} 
\newcommand{\qsenntable}{Q-SENN}
\newcommand{\slddtable}{SLDD-Model} 
\newglossaryentry{protopnet}
{
name={\textit{ProtoPNet}},
first ={\textit{ProtoPNet}~\citep{chen2019looks}},
description={Target Vector in Training}
}
\newglossaryentry{qsenn}
{
name={\textit{Q-SENN}},
first ={\textit{Q-SENN}~\citep{norrenbrock2024q}},
description={Target Vector in Training}
}
\newglossaryentry{PIP-Net}
{
name={\textit{PIP-Net}},
first ={\textit{PIP-Net}~\citep{nauta2023pipnet}},
description={Target Vector in Training}
}

\newglossaryentry{imgnetheader}
{
first = {ImageNet-1K~\citep{imagenet15russakovsky}},
long = {ImageNet-1K~\citep{imagenet15russakovsky}},
name={ImageNet-1K},
description={Final layer in the neural network}
}
\newglossaryentry{labelfreecbm}
{
name={\textit{Label-free CBM}},
first ={\textit{Label-free CBM}~\citep{oikarinen2023label}},
description={Target Vector in Training}
}
\newglossaryentry{ProtoTree}
{
name={\textit{Prototree}},
first ={\textit{ProtoTree}~\citep{nauta2021neural}},
description={Target Vector in Training}
}
\newglossaryentry{ImageSample}
{
name={\ensuremath{\boldsymbol{I}}},
first ={\ensuremath{\boldsymbol{I} \in \mathbb{R}^{3\times w\times h}}},
description={The classifier on the featuers}
}
\newglossaryentry{SaliencyOverlay}
{
name={\ensuremath{\boldsymbol{I}^s}},
first ={\ensuremath{\boldsymbol{I}^s \in \mathbb{R}^{3\times w\times h}}},
description={The classifier on the featuers}
}
\newglossaryentry{SaliencyMap}
{
name={\ensuremath{\boldsymbol{S}}},
first ={\ensuremath{\boldsymbol{S} \in \mathbb{R}^{w\times h}}},
description={The classifier on the featuers}
}

\newcommand{\resnet}{\gls{resNet}}

\newcommand{\attributeset}[1]{\ensuremath{\rho_{#1}}}

\newcommand{\loc}[1]{\textrm{SID@#1}}
\newcommand{\glm}{\gls{glmsaga}}
\newcommand{\glmtable}{glm-saga\textsubscript{5}}

\newcommand{\eg}{\textit{e}.\textit{g}. }
\newcommand{\ie}{\textit{i}.\textit{e}. }

\newcommand{\gtmatrix}{\ensuremath{A}^{\mathrm{gt}}}

\newcommand{\arrowDown}{\ensuremath{\boldsymbol{\downarrow}}}

\newcommand{\arrowUp}{\ensuremath{\boldsymbol{\uparrow}}} 

\tikzstyle{process} = [rectangle, minimum width=1cm, minimum height=1cm, text centered, text width=1.8cm, draw=black]
\tikzstyle{decision} = [diamond, minimum width=1cm, minimum height=1cm, text centered, draw=black]
\tikzstyle{arrow} = [thick,->,>=stealth]

\newcommand{\ClassSim}{\Psi}
\newcommand{\classSim}{\psi}
\newcommand{\simmat}{\ensuremath{\boldsymbol{A}}} 

\newcommand{\infvec}[1]{\ensuremath{s_{#1}}}

\newcommand{\FSimeaturemat}{\ensuremath{\boldsymbol{R}}}
\newcommand{\BSimeaturemat}{\ensuremath{\boldsymbol{b}}} 
\newcommand{\gurobi}{\textit{Gurobi}~\citep{gurobi}}

\newcommand{\fvecgurobi}{\ensuremath{\boldsymbol{s}}}
\newcommand{\wgurobi}{\ensuremath{\boldsymbol{W}}}

\newcommand{\generality}{\ifmmode
    \mathrm{Class\text{-}Independence}
  \else
    Class-Independence
  \fi}
\newcommand{\contrastiveness}{Contrastiveness}
\newcommand{\cubsim}{Structural Grounding}

\newcommand{\boldnessstatement}{Among more compact models, the best result is marked in bold, second best underlined.}
\newcommand{\cboldnessstatement}{Among more compact ($\circ$ or above) models, the best result is marked in bold, second best underlined.}

\newcommand{\interpmetricstablestart}[1]{Comparison on QPM ~\cite{qpmPaper} Interpretability metrics with #1. Requiring annotations, \cubsim{} can only be computed for \cubheader{}. 
 \boldnessstatement}
\newcommand{\accmetricstablestart}[1]{Accuracy and Compactness using #1: \hqpm{} shows the highest accuracy among interpretable models.}

%% file: secs/abstract.tex
\begin{abstract}
 Globally interpretable models are a promising approach for trustworthy AI in safety-critical domains.
Alongside global explanations, detailed local explanations are a crucial complement to effectively support human experts during inference.
 This work proposes the Calibrated Hierarchical QPM (\hqpm{}) which offers uniquely comprehensive global and local interpretability, paving the way for human-AI complementarity. 
 \hqpm{} achieves  superior global interpretability by contrastively explaining the majority of classes and offers novel hierarchical explanations that are more similar to how humans reason and can be traversed to offer a built-in interpretable Conformal prediction (CP) method.
Our comprehensive evaluation shows that \hqpm{} achieves state-of-the-art accuracy as a point predictor, maintaining 
 $99\%$ accuracy of non-interpretable models. This demonstrates a substantial improvement, where interpretability is incorporated without sacrificing overall accuracy. Furthermore, its calibrated set prediction is competitively efficient to other CP methods, while providing interpretable predictions of coherent sets along its hierarchical explanation.
%
\end{abstract}

%% file: secs/intro.tex

\begin{figure}
  \centering 
  \includegraphics[width=.9\linewidth]{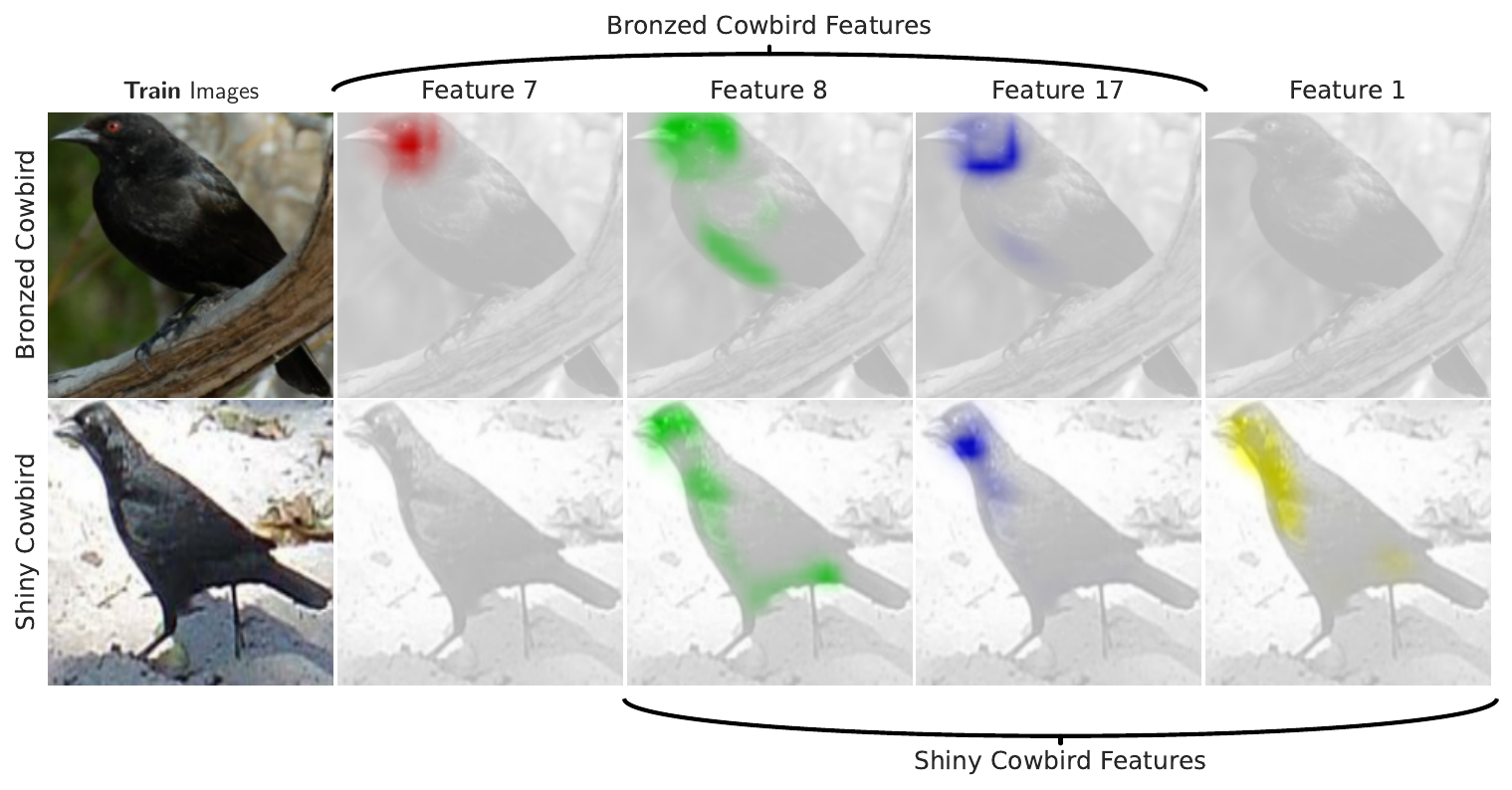} 
  \caption{Contrastive global Explanation, comparing the class representations of Shiny and Bronzed Cowbirds for \hqpm{} that represents every class with 3 of 30 features. The cowbirds are differentiated based on the red eye.}
  \label{fig:TeaserGlob} 
\end{figure}



\section{Introduction}
\begin{figure}[h!t]
    \centering 

    \begin{minipage}[b]{0.69\textwidth} 
        \centering
        \includegraphics[width=\linewidth]{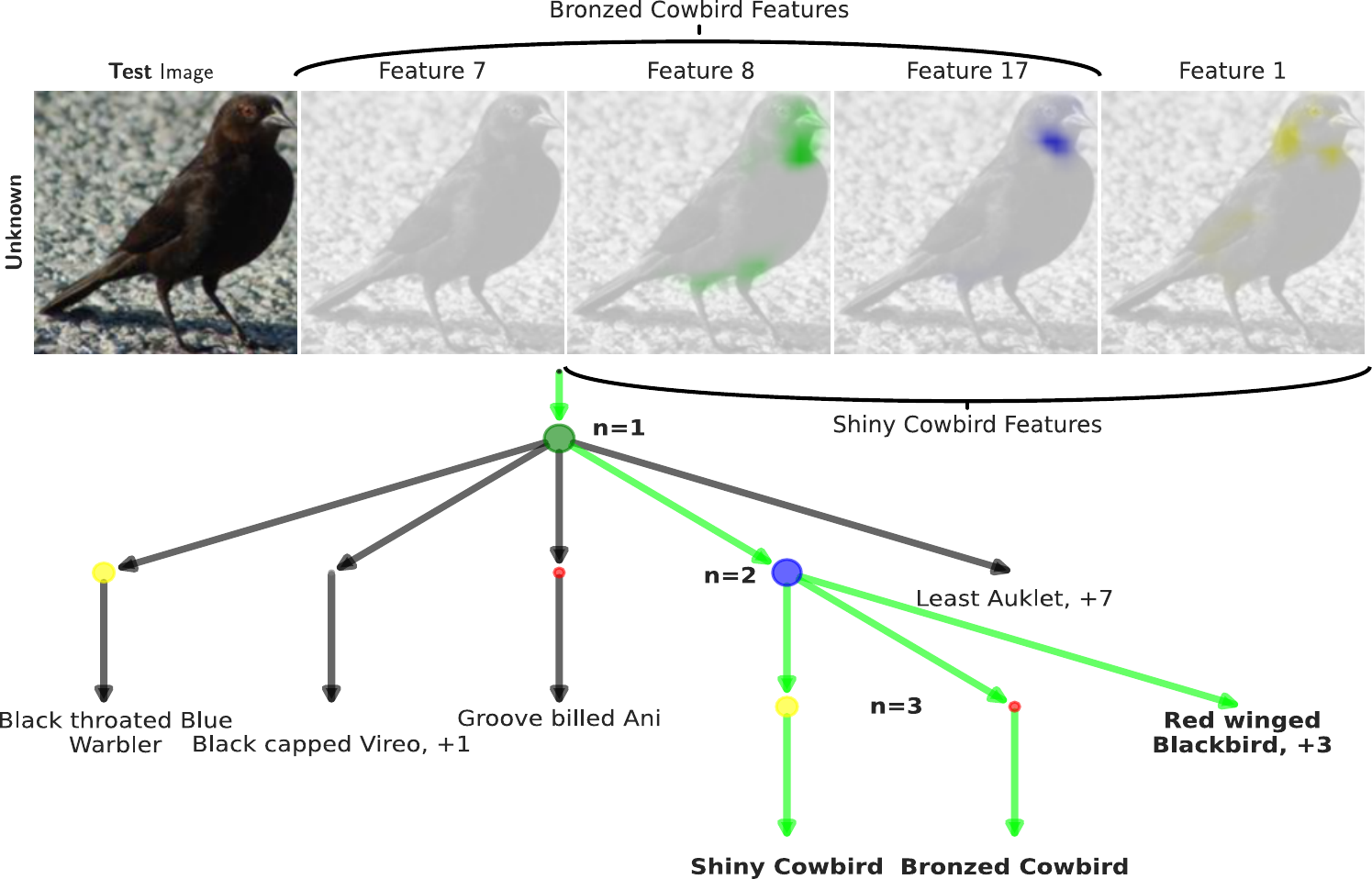} 
         \captionof{figure}
         {Exemplary local explanation provided by our \hqpm{}, with the global explanation in \cref{fig:TeaserGlob}, for a difficult test image of a Bronzed Cowbird with a pale red eye that is not clearly visible. This leads to negligible activation of the red-eye detecting Feature 7. The calibrated \hqpm{} provides a hierarchical explanation that communicates clear evidence for the predicted coherent set of black birds (marked in bold with green edges, including the correct label), but no sufficient evidence to differentiate between them.}
        \label{fig:TeaserFull} 
    \end{minipage}
    \hfill 
    \begin{minipage}[b]{0.3\textwidth} 
        \centering
        \includegraphics[width=\linewidth]{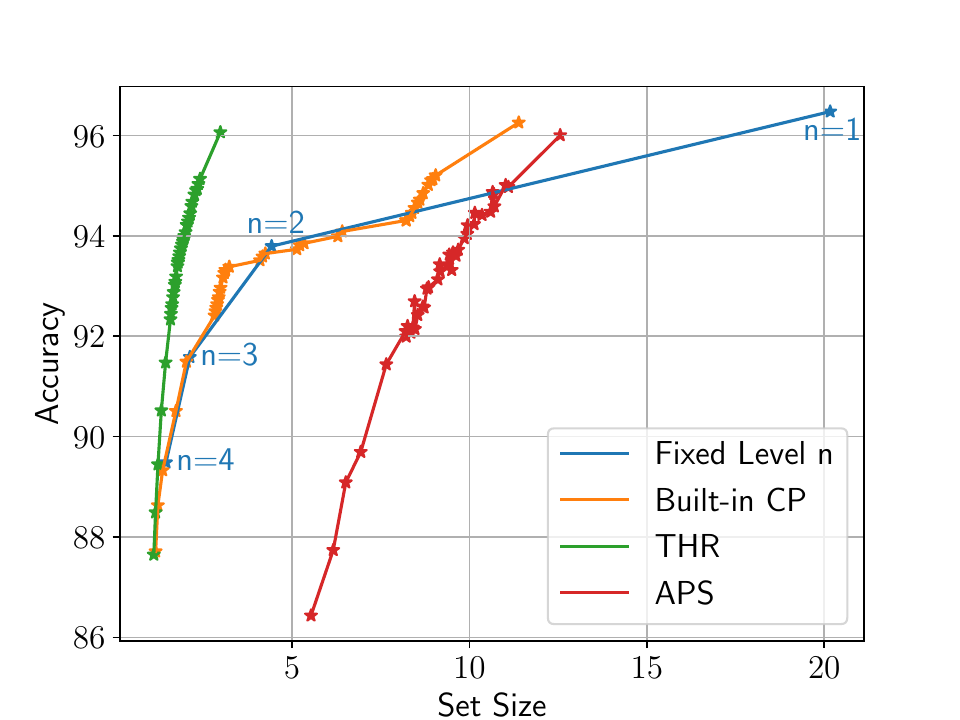} 
        \captionof{figure}{Coverage relative to the set size on \cubheader{} for various set prediction methods applied to \hqpm{} with $5$ out of a total of $50$ assigned features per class. The stars denote different calibration or hierarchy levels and are linearly interpolated.
 Traversing the hierarchical explanations (\cref{fig:TeaserFull}),
the built-in conformal prediction method predicts coherent sets with competitive efficiency to CP methods
THR~\cite{thrpaper} or APS~\cite{apspaper}.}
        \label{fig:introComp} 
    \end{minipage}


\end{figure}
Deep Learning has made remarkable advances and is being used more widely, including in high-stakes domains like medicine~\cite{ahsan2022machine} or autonomous driving~\cite{kuutti2020survey}.
Using more transparent models, \eg{} those that are interpretable by-design, is a promising approach to facilitate safety, robustness, and \newt{trust~\cite{schier2025explainable}} and is even required by law for some applications~\citep{veale2021demystifying}.
For domains like autonomous driving, with no expert present during inference, models with built-in global interpretability, that can generally explain their behavior, are valuable as their reasoning can be robustly tested and verified before deployment.
QPM and Q-SENN~\cite{norrenbrock2024q,qpmPaper} follow that goal by enforcing very compact class representations, that are made up of general, diverse and contrastive features, which are properties of human-friendly explanations~\cite{miller2019explanation}. 
\newt{The generality of these features is in contrast to prototypical networks like \gls{ProtoTree}, which seem inherently interpretable, as they use the similarity between image patches as crucial element of their computation. However, the space in which the similarity is computed is freely learnt which leads to class-specific prototypes~\cite{qpmPaper} and similarities humans can frequently not predict~\cite{hoffmann2021looks,kim2021hive}. Due to these shortcomings, this work follows the goal of models like QPM~\cite{qpmPaper} to represent classes using general features. }\\ 
Considering the cognitive limitations of average humans~\cite{miller1956magical}, QPM represents every class with the binary assignment of very few broadly shared features, usually $\leq5$ features.
A consequence is the emergence of highly contrastive class representations. 
Similar to \cref{fig:TeaserGlob}, the difference between two classes in the model's representation space can be concretely pointed out, like differentiating the two birds via their eye, just like humans do~\cite{wah2011caltech}.
While the contrastive representations are very helpful for QPM, they are also fairly rare, \eg{}on average just $0.13\%$ pairs per class on \cubheader.\\
Other domains, \eg{}medicine or science, can profit off of additional interpretability.
When a human expert is present, they should be supported rather than replaced, a notion known as human-AI complementarity.
While global interpretability is still beneficial in this scenario, the value of explaining the decision for a single sample rises, known as local explanation.
These typically have the form of saliency maps, such as GradCAM~\cite{selvaraju2017grad}, that visualize where the explainee saw support for its decision.
They can also be meaningfully computed for individual features if the globally interpretable model can be decomposed into detecting general human understandable concepts.
This crucial property is a key feature of our proposed \hqpm{}, as demonstrated in \cref{fig:TeaserFull,fig:TeaserGlob}.
However, those heatmaps generally do not transport a notion of certainty.
Therefore, predicting a set of classes with configurable guarantees on the accuracy using Conformal Prediction (CP)~\cite{vovk2005algorithmic} has emerged as a promising direction for supporting experts~\cite{10.5555/3692070.3693971,straitouri2023improving}.
Intuitively, more classes are predicted for uncertain or less conform samples, whereas a point-predictor always predicts just one class.
However, these sets typically contain a larger variety of classes, that resemble the misalignment between human and machine representations.\\
This work introduces the Calibrated Hierarchical QPM (\hqpm{}).
It improves the global interpretability of QPM while maintaining or improving the state-of-the-art accuracy by enforcing more pairs of  classes with highly contrastive class representations and adapting the training pipeline to ensure class representations made of interpretable features via the proposed Feature Grounding Loss \floss{} combined with ReLU activation. 
\hqpm{} is the first model with a built-in interpretable set prediction that can be calibrated via CP\newt{, inheriting all its robust guarantees}.
Intuitively, \hqpm{} predicts sets of classes by predicting all classes that share the dominant $n$ features with the most likely class, \eg{}predicting all the black birds in the hierarchical explanation  in \cref{fig:TeaserFull} below the blue feature at the tree level $n=2$.
\Cref{fig:introComp} demonstrates the already competitive efficiency of this set predictor compared to CP methods, while \hqpm{} can be calibrated using CP to dynamically select the appropriate level for a concrete sample.
This results in a novel way of providing hierarchical local explanations and traversing them to dynamically and understandably construct coherent prediction sets similar to how a human would reason.
Considering the graph in \cref{fig:TeaserFull}, the \hqpm{} found the green and blue feature, that identify black birds, but no sufficient evidence to differentiate between them.
Therefore, it predicts the coherent set of various black birds, including the correct class.
\newt{Holistically, the novel hierarchical local explanations answer an unprecedented range of questions simultaneuosly: 1. What meaningful features of which classes are found in this image? 2. How does each feature narrow down the set of potential predictions into increasingly similar classes? 3. Which set shall be predicted to guarantee a configurable average accuracy? 4. Which features would have needed to activate stronger in order to predict a smaller set with sufficient certainty?}

Our main \textbf{contributions}\footnote{The code is published: \url{https://github.com/ThomasNorr/CHiQPM/}.} are:
\begin{itemize}
\item We present the Calibrated Hierarchical QPM (\hqpm{}). 
It is based on a heavily constrained discrete quadratic problem (QP), that selects features from a black-box model and assigns them to classes.
The features of \hqpm{} then adapt to the 
optimal solution,
resulting in a globally and locally interpretable model.
\item \hqpm{} offers novel  hierarchical local
explanations and can be calibrated to reach a target coverage with competitive efficiency while ascending through its dynamically constructed interpretable class hierarchy and selecting the appropriate level. 
Thus, \hqpm{} can be considered an interpretable conformal predictor.
\item We present the Feature Grounding Loss \floss, which, alongside an additional ReLU, leads to learning more grounded and sparser features that facilitate compact hierarchical explanations along more human concepts.
\item The state-of-the-art performance of \hqpm{} as point- and built-in interpretable calibrated coherent set-predictor is evaluated across multiple architectures and datasets, including \imgnetheader, where the gap to the black-box baseline is more than halved.
\end{itemize}


%% file: secs/relatedWork.tex
\section{Background}
\subsection{Interpretable Machine Learning}
\label{sec:rlworkInt}
The field of Interpretable Machine learning can be split into models with interpretability by design
and methods that aim to explain models post-hoc~\citep{kim2018interpretability,bau2017network,Fel_2023_CVPR,pmlr-v202-kalibhat23a,oikarinen2023clipdissect,kaiser2025uncertainsam}.
This paper introduces a model that offers interpretability by design, which is why we focus on that part.
Two directions of interpretability can be defined: 
Explaining the decision for a single sample is  called \textit{local}, whereas a \textit{global} explanation is concerned with the general behavior of  a model~\cite{molnar2020interpretable}.
Measuring interpretability is an unsolved task on its own, but several desired criteria have been determined.
A human-friendly explanation should be diverse, general, compact and contrastive~\cite{miller2019explanation,lipton1990contrastive,read1993explanatory}.
The SENN~\cite{alvarez2018towards} framework, which \hqpm{} can be considered as, further adds grounding as criteria, which refers to the alignability of learned representation with human concepts. 
One issue with measuring grounding is the prevalence of polysemantic neurons~\citep{scherlis2022polysemanticity, elhage2022toy, templeton2024scaling}, which refers to features that detect multiple human concepts simultaneously. 
Nevertheless, multiple metrics have been proposed to quantify the desirable aspects~\cite{qpmPaper,norrenbrock2024q}, which we evaluate in this work.\\
Globally interpretable models typically make their classification using a simple interpretable model applied to more interpretable features.
These features are either freely learned with priors or losses to induce interpretability~\cite{norrenbrocktake,norrenbrock2024q,qpmPaper,nauta2023pipnet}, supervised with human concepts in the family of Concept Bottleneck Models (CBM)~\cite{koh2020concept,sawada2022concept,margeloiu2021concept,marconato2022glancenets,dominici2025counterfactual} or restricted to closely resemble parts of the training data, so-called prototypes~\cite{rymarczyk2021protopshare,nauta2021neural,ma2023looks,chen2019looks}.
Recently, the simple interpretable model is typically a very compact linear layer~\cite{nauta2023pipnet,norrenbrocktake,norrenbrock2024q,qpmPaper}, but another example is a decision tree.
In its most interpretable form, \gls{ProtoTree} learns a static deep decision tree to classify based on similarities to learned prototypes.
Like all prototype models, these similarities are typically not predictable for a human~\cite{kim2021hive,hoffmann2021looks} and the interpretability of a very deep decision tree is debatable.
Apart from the static tree in \gls{ProtoTree}, \hqpm{} with its sample-specific low-depth 
hierarchical explanation is most comparable to \glsname{layerName}, \glsname{qsenn} and QPM~\cite{norrenbrocktake,norrenbrock2024q,qpmPaper}.
They follow a pipeline of training a black-box model, making it compact, i.e. selecting features from the black box and sparsely connecting them to classes, and then fine-tuning the compact model to achieve a superior accuracy-compactness tradeoff.
\glsname{layerName} and \glsname{qsenn} used \glm{} for feature selection and their class assignment, which includes normalizing the features and is a very local optimization.
QPM improves upon that by formulating the selection and assignment as quadratic problem and solving it globally optimal, improving upon accuracy and interpretability via the binary comparable class representations with a fixed number of features.
\newt{
    The incorporated constraints on compactness enable a quantitative measurement of a metric that relates to interpretability. While prior work~\cite{glandorf2025p3b,glandorf2023hypersparse,rosenhahn2023optimization} primarily optimized compactness for efficiency, this and other work in interpretable machine learning
\cite{qpmPaper,norrenbrock2024q,nori2019interpretml, nauta2023pipnet} focus on it for interpretability. 
}
\subsection{QPM}
\label{sec:qpm}
\input{secs/methodoverview}
QPM~\cite{qpmPaper} follows a pipeline similar to~\cref{fig:OverviewAppraoch}.
The first step is training a black-box dense model with an auxiliary Feature Diversity Loss~\gls{customLoss}~\cite{norrenbrocktake} to ensure that the initial $\gls{nFeatures}$ feature maps activate on distinct locations of the image. 
\paragraph{QP}
Afterwards, a quadratic problem is formulated to jointly find an optimal selection of $\gls{nReducedFeatures}$ 
 features  $\fvecgurobi~\in\{0,1\}^{\gls{nFeatures}}$ and 
assignment between features and $\gls{nClasses}$ classes $\wgurobi~\in\{0,1\}^{\gls{nClasses}\times \gls{nFeatures} }$, with 
$\wgurobi^*~\in\{0,1\}^{\gls{nClasses}\times \gls{nReducedFeatures} }$ denoting \wgurobi{} for the selected features.
These variables are then optimized so that the selected features are assigned to classes they are maximally correlated to, described in the class-feature similarity matrix $\simmat{}~\in \mathbb{R}^{\gls{nClasses}\times \gls{nFeatures} }$. 
Additionally, the problem formulation steers the feature selection towards distinct and local features, encoded in the feature-feature similarity matrix $~\FSimeaturemat~\in\mathbb{R}^{\gls{nFeatures}\times \gls{nFeatures} }$ or linear bias term $~\BSimeaturemat~\in\mathbb{R}^{\gls{nFeatures} }$ respectively.
Additionally, the quadratic problem is constrained to result in a solution with the desired level of compactness:
\begin{align}
  \sum_{\findex=1}^{\gls{nFeatures}}\infvec{\findex} &= \gls{nReducedFeatures} &&
   \sum_{\findex=1}^{\gls{nFeatures}} w_{\cindex,\findex} \infvec{\findex} = \gls{nperClass} \quad \forall \cindex \in \{1, \dots,\gls{nClasses}\}
   \label{eq:5perInit}
\end{align} 
The even sparsity 
can cause duplicates to arise.
QPM prohibits these with an iterative optimization, that also includes an efficient incorporation of \cref{eq:5perInit}.
For more details, we refer the reader to~\cite{qpmPaper}. 
Finally, the features are fine-tuned with a fixed $\wgurobi^*$, so that they adapt to their assigned classes and detect shared general concepts, 
detailed in \cref{ssec:Tune}.

\subsection{Conformal Prediction}
\label{sec:introconf}
We calibrate our model using the Conformal Prediction~(CP)~\cite{vovk2005algorithmic} framework.
For a more thorough introduction, we refer the reader to~\cite{conformalINfo} which we summarize shortly.
Conformal Prediction is a mathematically robust framework with minimal assumptions that 
can guarantee a desired error rate $\alpha$ is not exceeded. 
The most popular framework and the one we use is called split conformal prediction~\cite{papadopoulos2002inductive}.
It is based on splitting off calibration data \datacal{} to calibrate any model and guarantees an error rate of up to $\alpha$ on data $(x_\mathrm{test},y_\mathrm{test})$ that is exchangeable with \datacal:
\begin{equation}
    1-\alpha \leq P(y_\mathrm{test} \in \predset{}(x_{test}))
\end{equation}
Here, $\predset{}(x_{test})$ describes the predicted classes for the test sample.
It is constructed using a nonconformity score $s(x,c)$, that should relate to how well the predicted label $c$ fits to the input $x$.
This calibration score is computed on the entire calibration set \datacal{}, resulting in a set of scores $\mathbb{S}$ and its values are used to construct the test prediction sets:
\begin{equation}
    \predset{}(x_{test}) = \{ c \in\allClassSet: s(x_{test},c) \leq \text{Quantile}(1-\alpha, \mathbb{S})\}
\end{equation}
Simply put, if $(1-\alpha)$ of the true labels on the calibration data were less nonconform, i.e. had a lower nonconformity score, than a candidate class $c$ during prediction, $c$ will not be included in \predset{}.
The nonconformity score
$s(x_{test},c)$ can be fairly arbitrarily defined and just needs to capture some notion of conformity and have sufficiently many distinct values, so that the quantile has the desired resolution.
CP is typically evaluated under two aspects: a)~Unconditional coverage that guarantees a minimum average accuracy on an equally distributed test set. b)~Conditional coverage that aims for the desired expected accuracy for all test samples, even difficult ones. 
The simple method THR~\cite{thrpaper} is fairly ideal for unconditional coverage.
Therefore, research has focused on conditional coverage, \eg{}APS~\cite{apspaper}, also in difficult settings~\cite{ding2024class}, or conformal training~\cite{stutz2022learning}, which involves directly optimizing for the size of predicted sets during training.
\citet{cortes-gomez2025utilitydirected} even steer the prediction sets towards more coherence.
To the best of our knowledge, this is the first work that 
proposes the notion of interpretable CP.
The predicted sets are constructed by traversing hierarchical explanations, leading to coherent sets of similar classes by design, while 
efficiently 
ensuring unconditional coverage. 




%% file: secs/methodoverview.tex
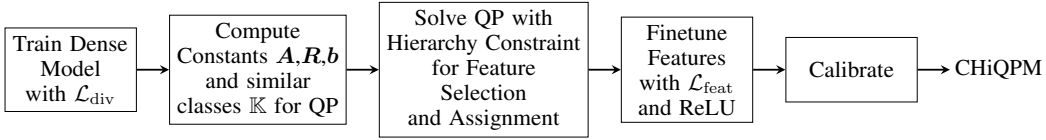
\begin{figure*}[t]
    \centering
    \resizebox{\linewidth}{!}{\begin{tikzpicture}[node distance=.01cm]
        \node (denseTraining) [process] {Train Dense Model with \gls{customLoss}};
          \node (ComputeMatrics) [process, right=0.5cm of denseTraining, text width=2.5cm] {Compute Constants \simmat{},\FSimeaturemat{},\BSimeaturemat{} and similar classes \pairsTarget{} for QP};
        \node (featureSelAssign) [process, right=0.5cm of ComputeMatrics, text width=3cm] {Solve QP with\\Hierarchy Constraint\\for Feature Selection \\and Assignment};
        \node (finetune) [process, right=0.5cm of featureSelAssign] {Finetune Features\\with \floss\\ and ReLU};
        \draw [arrow] (denseTraining) -- (ComputeMatrics);
        \draw [arrow] (ComputeMatrics) -- (featureSelAssign);

        \node (calibrate) [process, right=0.5cm of finetune] {Calibrate};
       \draw [arrow] (featureSelAssign) -- (finetune);
        \draw [arrow] (finetune) -- (calibrate);
         \node(end) [right=0.5cm of calibrate] {};
        \draw [arrow] (calibrate) -- (end) node[midway, right, xshift=.2cm] {\hqpm};
   \end{tikzpicture}
    }
\caption{Overview of our proposed pipeline to obtain a \hqpm}
    \label{fig:OverviewAppraoch}
\end{figure*}

%% file: secs/method.tex
\section{Method}
\label{sec:metho}

The proposed \hqpm{} is designed to classify an input image, denoted as \gls{ImageSample} into one or more of \gls{nClasses} classes, represented as $c \in \allClassSet= \{c_1, c_2, ..., c_{\gls{nClasses}}\}$.
\hqpm{} consists of a deep feature extractor, $\Phi$, which processes \gls{ImageSample} to generate a low-dimensional feature vector \gls{RedfeatureVector} via 
feature maps \Gls{featureMapsSmall} of dimensions $\gls{featuresMapwidth} \times \gls{featuresMapheigth}$, and a sparse interpretable final assignment $\wgurobi^*~\in\{0,1\}^{\gls{nClasses}\times \gls{nReducedFeatures} }$ of these features to classes.
Thus, the prediction of \hqpm{} can be expressed as
$\gls{outputVector} = \wgurobi^*\gls{RedfeatureVector}$, with $\hat{c} = \mathrm{argmax}(\gls{outputVector})$ being the predicted class.
Compared to QPM, we apply an additional ReLU to the features   $\textbf{f}^* = \mathrm{ReLU}( \hat{\textbf{f}})$,
so that \hqpm{} only reasons positively, and negligible activations are suppressed.
Notably, all classes in \hqpm{} are represented with the same number \gls{nperClass} of features per class, which leads to easily comparable class representations. 
It is easiest, when two classes share exactly $\gls{nperClass}-1$ features, as
 the differentiating factor can then be concretely pointed out, as shown in \cref{fig:TeaserGlob}.
We denote the set of these pairs with 
\begin{equation}
     \CpairsSet = \{ 
    (i,j) : 1 \leq i < j \leq \gls{nClasses}
    \text{ and }\boldsymbol{w}^*_i\boldsymbol{w}{^*}_{j}^{T } = \gls{nperClass} -1\} 
\end{equation}
\hqpm{} follows a similar pipeline to QPM, shown in \cref{fig:OverviewAppraoch}. 
\hqpm{} improves upon \gls{NewlayerName} (\cref{sec:qpm}) 
with easier interpretable class representations for more of the classes via an increased $\lvert\CpairsSet\rvert$, directly enforced in the QP  (\cref{sec:constr}),  
alongside a built-in interpretable conformal prediction method traversing the novel local hierarchical explanations  (\cref{sec:confMeth})  and an improved fine-tuning with a new Feature Grounding Loss \floss{} that improves grounding and compactness of \hqpm{}'s explanations (\cref{sec:MethLoss}).
\subsection{Hierarchical Constraint}
\label{sec:constr}

    

In order to ensure a higher cardinality $\lvert\CpairsSet\rvert$ resulting in more easily comparable class
representations,
an additional constraint is added to the quadratic problem. 
\newt{
A set \pairsTarget{} of pairs of classes that are highly similar in our dense model is determined, and we ensure that most of these are part of the resulting \CpairsSet{} via a two-step process.
Specifically, we calculate the class-class similarity $\ccsim\in\mathbb{R}^{\gls{nClasses}\times \gls{nClasses} }$ based on the similarities computed for the QP 
\simmat{} and set \pairsTarget{} as its $\treedense \cdot \gls{nClasses}$ most similar class pairs:
\begin{align}
    \ccsim &= \simmat{} \simmat{}^T - \textbf{I}_{\gls{nClasses}} &&
    \pairsTarget = \{ 
   (i,j) : 1 \leq i < j \leq\gls{nClasses}
    \text{ and }\ccsim_{i,j}  \geq \theta\}    
\end{align}
}
Here, $\theta = \text{sort}(\ccsim)_{2\cdot\treedense \cdot \gls{nClasses}}$ describes
the $(2\cdot\treedense \cdot \gls{nClasses})$-th highest value in $\ccsim{}$, 
$\textbf{I}_{\gls{nClasses}}$ is the identity matrix and
\treedense{} is a hyperparameter that 
controls how many classes each class should be very similar to in the representations of the resulting \hqpm{}, with $\rho=0.5$ enforcing on average one very similar class each. 
A higher \treedense{} increases the number of classes that can be contrastively globally explained as in \cref{fig:TeaserGlob}, thus improving global interpretability.
Forcing similar classes to be represented very similarly by \hqpm{} further induces an efficient use of the $\gls{nReducedFeatures}$ features, as shared concepts need only be detected by shared features.
This improves the grounding of the hierarchical explanations and thus the efficiency of our built-in set prediction.
However, with increasing \treedense{}, the risk of adding classes to  \CpairsSet{} that do not share $\gls{nperClass}-1$ general concepts rises.
Those pairs would cause the shared features to be less robust and thus harm performance.
Thus, there can be a tradeoff between point and set prediction.
Therefore, \treedense{} 
should be set using calibration data to efficiently achieve a target coverage while balancing point prediction, which is ablated in \cref{sec:abls,ssec:rhodense}.
After obtaining \pairsTarget, the similarity of each 
included
pair
is directly added as constraints to the discrete optimization:
\begin{equation}
   (\boldsymbol{w}_\cindex \circ \boldsymbol{w}_{\cindex'})^T \fvecgurobi{} = \gls{nperClass} -1 \quad \forall (\cindex,\cindex') \in \pairsTarget
   \label{eq:o4sim}
\end{equation}
We relax this constraint after finding the initial global solution to find the most optimal way for the \hqpm{} to have sufficient, \ie$\lvert\CpairsSet\rvert\geq\lvert\pairsTarget\rvert$ , highly similar classes in its representations. This is detailed in \cref{ssec:Relax}.
\subsection{Set Prediction}
\label{sec:confMeth}
This section describes how the proposed \hqpm{} can be used to predict interpretable sets \predset{} along its local hierarchical explanation.
The construction of these explanations and how they can be used to predict 
sets at a fixed hierarchy level is first formalized. 
Then, we show how \hqpm{} is calibrated using Conformal Prediction to 
predict coherent sets \predset{} that contain the target label with an error rate $\alpha\in(0,1)$ while still traversing the explaining hierarchy.

\subsubsection{Hierarchical Explanation}
\label{sec:hierexp}
Our \hqpm{} enables the construction of hierarchical local explanations for a concrete test sample, as shown in \cref{fig:TeaserFull}.
It contains nodes for all nonzero feature activations and indicates the presence of all classes that are assigned to at least one of the shown feature-nodes.
For every class $c$ with its assigned features $\classfeatures^{c}\in\{1, \dots, \gls{nReducedFeatures}\}^{\gls{nperClass}}$, the features are shown in the order of their activation, which can be interpreted as reasoning from the more clearly visible feature like the neck in \cref{fig:TeaserFull} to the less certain features, such as the pale red eye.
The class node is then attached to its last activating feature and the class would be predicted if \hqpm{}'s calibration or fixed level determines that this feature and all its descendants should be predicted. 
To formalize predicting at a fixed level, we define 
the order of activations 
$\orderedClassFeatures^{c}$
of the assigned features for every class $c$, where 
$\orderedClassFeatures^c $ 
is ordered so that:
\begin{equation}
    f^*_{\orderedClassFeatures^c_i} \geq 
 f^*_{\orderedClassFeatures^c_j} \quad \forall \quad i,j  \in\{1,\dots,\gls{nperClass}\}\text{ where } i < j
\end{equation}
Additionally, we introduce the indicator function \indicator{n}{c} at depth $n\in\{1,\dots,\gls{nperClass}\}$, which indicates if the class $c$ shares the same top $n$ features with the predicted class \predclass{}:
\begin{equation}
    \indicator{n}{c} = \begin{cases}
  1, & \text{if} \quad \orderedClassFeatures^c_i  = \orderedClassFeatures^{\predclass{}}_i \quad \forall i  \in\{1,\dots, n\}\ \\
  0, & \text{otherwise}
\end{cases}
\end{equation}
Thus, predicting at a fixed depth of $n$ features shared with the predicted class \predclass{} is:
\begin{equation}
        \predset{}^n = \{ c \in\allClassSet:\indicator{n}{c}=1\}\label{eq:fixPred}
\end{equation}
For example in the graph in \cref{fig:TeaserFull},  
$\predset{}^1$ contains all 
shown class nodes, $\predset{}^2$ 
the five classes that share the green and blue feature and finally 
$\predset{}^3 = \predset{}^{\gls{nperClass} }=\{\predclass{}\}$, 
as only the predicted class \predclass{} shares all features with itself.
Note that the stars for the fixed depth line in \cref{fig:introComp} indicate predicting with a fixed $n\in\{ 1, \dots, \gls{nperClass}\}$.
For more compact explanations, we restrict the graphs in the main paper to include only those classes which share the most activating feature with the predicted class.
However,
even for large datasets, full graphs are still informative and included in the appendix, along further visualization details in \cref{ssec:VizDetails}.
\subsubsection{Nonconformity Score}
\label{sec:nonconformscore}
As introduced in \cref{sec:introconf}, the nonconformity score $s$ is a crucial aspect of any Conformal Prediction method.
In order to construct the prediction sets for \hqpm{} by going up the class hierarchy, we compute the nonconformity score based on how similar the class is in the hierarchy to the initially predicted class \predclass{}, as we are ascending the hierarchy that led to \predclass{}.
\newt{
Note that the introduced method offers all the guarantees of CP, as only a different nonconformity score is used.
}
For every class $c$, we propose to use the activations of the features in the shared path down the tree as nonconformity score:
\begin{align}
    \upscore(c) =   - \sum_{i=1}^{\gls{nperClass}} \indicator{i}{c} f^*_{\orderedClassFeatures^c_i}\label{eq:upscore}
\end{align}
 Note that \indicator{i}{c} ensures that all $i$ features are shared and the dependency on the input sample of all variables is omitted for brevity.
 We call this simple nonconformity score \textit{up} as it goes strictly up the tree.
 \paragraph{Subtree Selection}
The conformal predictor can predict with more granularity and achieve its guarantees when the prediction can also go down towards only some subtrees below the feature node determined by \upscore(c).
That also allows \hqpm{} to predict those descendants preferably that have some support beyond the shared path, \eg choosing to predict only the cowbirds in \cref{fig:TeaserFull}.
Therefore, we extend the nonconformity score to also account for the activation of the feature at the point of diversion after sharing $k$ features:
\begin{align}
    i^{\mathrm{div}} &= \orderedClassFeatures^c_{k+1} \quad \text{with} \quad k =\sum_{j=1}^{\gls{nperClass}-1}\indicator{j}{c}\label{eg:finscore1}\\
     \finscore(c) &=  -  f^*_{i^{\mathrm{div}}}  - \sum_{j=1}^{\gls{nperClass}-1} \indicator{j}{c} f^*_{\orderedClassFeatures^c_j}\label{eg:finscore}
\end{align}
\paragraph{Limited Level}
Finally, the maximum number of levels the set is constructed from is limited to ensure efficient sets.
Towards that goal, the minimum reachable error rate $\alpha^n_{cal}$ on the calibration data for each fixed level $n$ is calculated according to \cref{eq:fixPred}.
The conformal prediction is then limited to the highest level \adalvl{} that still reaches the target coverage defined by $\alpha$.
To ensure the limitation, we limit \upscore{} to \adalvl{}, and
multiply the score with the indicator function \indicator{n}{c} indexed at \indicator{\adalvl}{c}.
This
ensures all classes that were not correctly predicted under \adalvl{} get the most nonconform score of $0$
\footnote{\Cref{eq:adalvl,eg:finscore,eg:finscore1} have been corrected from the NeurIPS 2025 proceedings version to accurately reflect the mathematical logic of the published code.}:
 \vspace{-.35cm}
\begin{align}
s(c) = \underbrace{\indicator{\adalvl}{c}}_{\text{Limitation}} \cdot \Biggl( \underbrace{- f^*_{i^{\mathrm{div}}} - \sum_{j=1 +\adalvl{}}^{\gls{nperClass} - 1} \indicator{j}{c} f^*_{\orderedClassFeatures^c_j} }_{\text{Limited \finscore{}}} \Biggr)\label{eq:adalvl}
\end{align}
 \vspace{-.7cm}

\subsection{Feature Grounding Loss}
\label{sec:MethLoss}

\begin{wrapfigure}{R}{0.6\textwidth} 
  \centering 
  \includegraphics[width=\linewidth]{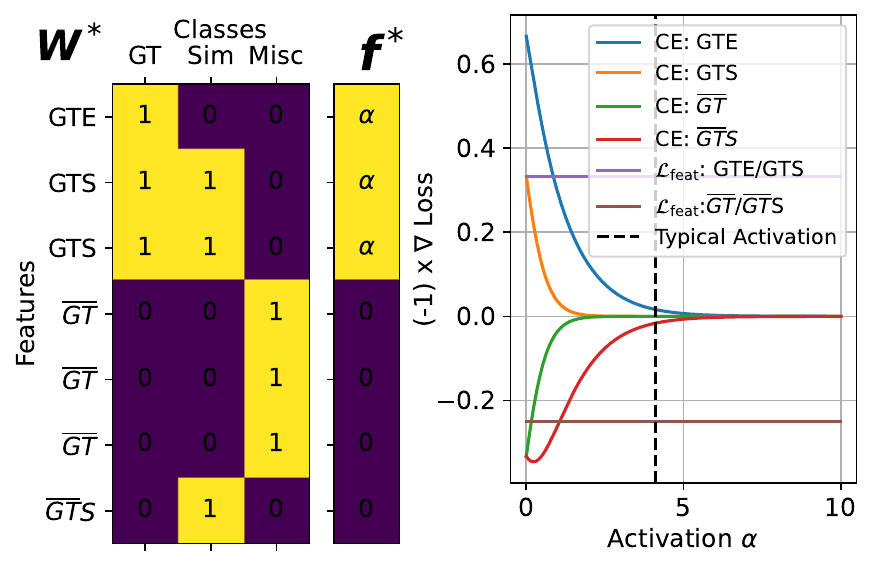} 
  \caption{Gradient on features $\boldsymbol{f}^*$ for a train sample labeled \textit{GT} for a toy example with 3 classes and 7 
  features, with $\wgurobi{}^*$ shown left. 
  At the average activation on the CUB dataset, the  Ground Truth Exclusive (GTE) feature  has a 
  roughly $4000$
  times higher gradient than the other assigned features, which are shared with \textit{Sim}, Ground Truth Shared (GTS). 
  }
  \label{fig:gradLoss} 
\end{wrapfigure}

This section presents our novel Feature Grounding Loss \floss. 
Its motivation is shown in the example in \cref{fig:gradLoss}.
In this 
example, two similar classes share $2$ of their $3$ features, hence they are in \CpairsSet.
This is relevant, as most classes in each \hqpm{} have a similar class due to the constraints described in \cref{sec:constr}.
For a concrete training example, where one of these is the ground truth, Cross-Entropy loss only causes a significant gradient on the differing Ground Truth Exclusive (\gtns{}) feature.
Therefore, the distinguishing feature \gtns{} is pushed to also activate on other concepts of the ground truth class, instead of just activating on the general concept that would differentiate them.
To alleviate that issue, we propose the Feature Grounding Loss \floss:
\begin{align}
     \floss = -\frac{\sum_{i \in \classfeatures} \frac{f^*_i}{\lvert\classfeatures\rvert} - \sum_{i \in \noclassfeatures} \frac{f^*_i}{\lvert\noclassfeatures\rvert}}{\mathrm{max}(f^*)}\label{eq:scaleFloss}
\end{align}
Here, $\classfeatures$ and $\noclassfeatures\in\{1, \dots, \gls{nReducedFeatures}\}^{\gls{nReducedFeatures}-\gls{nperClass}}$ indicate the indices of the features of the ground truth class or those not assigned to it, respectively.
\floss{} describes the difference in average activation between \classfeatures{} and \noclassfeatures, scaled by the maximum activation to prevent increasing activations.
As shown in \cref{fig:gradLoss}, $\Delta$ has the same gradient for all positively assigned features and also has a nonzero gradient for all features in \noclassfeatures.
Therefore, it encourages all the features of the ground truth class to detect more general concepts while also inducing sparsity in the feature activations, enabling the visualization of the entire hierarchical explanation.
We add \floss{} to our overall training loss during fine-tuning with weighting  $\lambda_\mathrm{feat}$.

%% file: secs/results.tex
\section{Experiments}
\label{sec:exp}

Following QPM~(\cref{sec:qpm}),
we evaluate our method on \cubheader{}, \stanfordheader{} and \imgnetheader.
\cubheader{} and \stanfordheader{} are the most commonly used datasets for interpretability, while \imgnetheader{} is suitable to demonstrate how the method scales to larger problems with more real-world applications.
\cubheader{} 
includes human annotations of relevant concepts for every image, which makes it suitable for evaluating the alignment between human representations and the ones learned.
As the proposed method can be applied to any backbone, we show results on \resnet{}, Resnet34, \gls{incv} and Swin-Transfomer-Small~\cite{liu2021swin} to allow an easy comparison with previous work.
The main paper focuses on results for \resnet{}, always reporting the mean across $5$ random seeds, with $3$ for \imgnetheader{}.
More results are included in
\cref{stab:res50Interp,stab:aCCproto-table,stab:r34aCCproto-table,stab:r34Interpproto-table,stab:incaCCproto-table,stab:incInterpproto-table,stab:inc,stab:r34,sstab:swInterpproto-table-adapted,stab:sw,sstab:swcaCCproto-table},
demonstrating the robustness of our method across random seeds and 
architectures.
For implementation details, we generally followed QPM~\cite{qpmPaper}, but report details and slight deviations in \cref{ssec:impdeta}.
Specifically, all models that follow a pipeline similar to ours, \glsname{layerName}, Q-SENN and QPM, use the same parameters for the dense training.
As usual in literature, 
$\gls{nperClass}=5$ and $\gls{nReducedFeatures}=50$ 
are set
if not reported otherwise.
Further, we generally set the density parameter for our class hierarchy to $\treedense=0.5$, as it is sufficient to demonstrate the improvements in built-in set prediction without sacrificing accuracy as point-predictor.
Finally, $\lambda_\mathrm{feat}=3$ is set as higher values cause reduced accuracy.


\subsection{Metrics}
\label{sec:metrics}
\begin{table}[t]
\caption{Comparison on Accuracy, Compactness, \contrastiveness{} and \cubsim{}. 
Compact describes the number of features \gls{nReducedFeatures} and features per class \gls{nperClass}.
It is binned into very compact + ($\gls{nperClass}=5$ and $\gls{nReducedFeatures} =50$), medium $\circ$, and the baseline, denoted - ($\gls{nperClass}=2048 $ and $\gls{nReducedFeatures} =2048)$, exact figures in \cref{stab:aCCproto-table}. 
\cref{fig:dimVar} shows the increasing gap when further raising compactness. \cboldnessstatement}
\label{tab:MainPointRes}
  \centering
  \begin{tabular}{lcccccccc}
    \hline
    \multirow{2}{*}{Method} & \multicolumn{3}{c}{Accuracy \arrowUp} & Com- & \multicolumn{3}{c}{Contrastiveness \arrowUp} & SG \arrowUp \\
    & CUB & CAR & IN & pact & CUB & CAR & IN & CUB \\
    \hline
    Dense Resnet50 & 86.6 & 92.1 & 76.1 & - & 74.4 & 75.1 & 71.6 & 34.0 \\
    \hline
    \glmtable{} & 78.0 & 86.8 & 58.0 & $\circ$  & 74.0 & 74.5 & 71.7 & 2.5 \\
    \pipnettable{} & 82.0 & 86.5 & - & $\circ$  & \underline{99.5} & \underline{99.5} & - & 6.7 \\
    \protopooltable{} & 79.4 & 87.5 & - & $\circ$ & 76.7 & 78.9 & - & 13.9 \\
    \midrule
    \slddtable{} & 84.5 & 91.1 & 72.7 & \textbf{+} & 87.2 & 89.7 & \underline{93.4} & 29.2 \\
    \qsenntable{} & 84.7 & 91.5 & \underline{74.3} & \textbf{+} & 93.0 & 94.2 & 92.6 & 23.4 \\
    \gls{NewlayerName} & \underline{85.1} & \underline{91.8} & 74.2 & \textbf{+} & 96.0 & 97.7 & 89.3 & \underline{47.9} \\
    \hline
    \tabOurs & \textbf{85.3} & \textbf{91.9} & \textbf{75.3} & \textbf{+} & \textbf{99.9} & \textbf{100} & \textbf{99.9} & \textbf{75.0} \\
    \hline
  \end{tabular}
\end{table}
\begin{table*}[t]
\caption{Average Set Size $\lvert\predset{}\rvert$ of \hqpm{} calibrated to reach various coverages $1-\alpha$ comparing different conformal prediction methods. All methods are very close or reach the desired coverage.
}
\label{tab:confCComparison}
\centering
\resizebox{\linewidth}{!}{
\begin{tabular}{lcccccccccccc}
\toprule
 \multicolumn{1}{c}{$\lvert\predset{}\rvert$\arrowDown} & \multirow{2}{*}{\makecell{Inter-\\pretable}}  & \multicolumn{4}{c}{CUB} & \multicolumn{3}{c}{CARS} & \multicolumn{4}{c}{INET} \\  
 \cmidrule(lr){1-1} \cmidrule(lr){3-6} \cmidrule(lr){7-9} \cmidrule(lr){10-13}
Method &  &$\alpha$=0.12 & $\alpha$=0.1 & $\alpha$=0.075 & $\alpha$=0.05 & 
$\alpha$=0.075 & $\alpha$=0.05 & $\alpha$=0.0025 & $\alpha$=0.22 & $\alpha$=0.2 & $\alpha$=0.175 & $\alpha$=0.15 \\
\midrule
\ours & \cmark& 1.22 & 1.73 & 2.94 & 9.05 & 1.05 & 1.25 & 8.25 &  1.10 & 1.42 & 3.25 & 4.58 \\ 
$s = \finscore$ & \cmark& 4.62 & 6.15 & 9.53 & 29.4 & 3.62 & 5.95 & 28.4 & 8.14 & 11.3 & 17.9 & 30.5\\ 
$s = \upscore$ &\cmark&  3.03 & 3.91 & 8.87 & 18.7 & 2.32 & 3.27 & 17.9 & 4.36 & 6.23 & 11.8 & 31.4\\ 
\midrule
THR& \xmark& 1.16 & 1.32 & 1.67 & 2.41 & 1.02 & 1.15 & 2.09 & 1.05 & 1.16 & 1.40 & 1.87\\
APS & \xmark&6.30 & 7.20 & 8.54 & 11.3 & 5.64 & 6.83 & 9.61 &  16.7 & 18.9 & 22.1 & 26.8 \\
\bottomrule
\end{tabular}
}
\end{table*}

\hqpm{} is designed to improve upon 
QPM~(\cref{sec:qpm}),
primarily via more easily interpretable class representations, being able to produce meaningful hierarchical explanations and by offering the built-in interpretable calibrated set prediction.
Therefore, we evaluate \hqpm{} across all QPM metrics in addition to the accuracy as point predictor in relation to its compactness.
For evaluating the performance as set predictor, we report the size of the predicted sets, when calibrated to reach a specific accuracy or coverage.
Finally, the annotations in \cubheader{} are used to compute the \setmetricsim{}  of the predicted sets and also measure the grounding of the features via the Alignment~\cite{norrenbrock2024q} metric, alongside QPM's \cubsim{}.
Following QPM, we compute the ground truth class-class similarity matrix $\boldsymbol{\mathrm{\ClassSim}}^{gt} = \boldsymbol{\Lambda}\boldsymbol{\Lambda}^T$ using the annotated 
average class attributes $\boldsymbol{\Lambda}\in[0,1]^{\gls{nClasses}\times312}$ with columns
$\boldsymbol{\lambda}_\cindex\in[0,1]^{312}$, where $\lambda_{\cindex,j}$ indicates the fraction of images with label $\cindex$, in which attribute $j$ is annotated to be present.
$\boldsymbol{\mathrm{\ClassSim}}^{gt}$ enables quantifying the similarities of classes that are predicted together, the novel \setmetricsim{} $\mathrm{sc}$:
\begin{align}
\mathbb{K}(\predset{}(x_\mathrm{test})) &= \{ \mathrm{\classSim}^{gt}_{\cindex,\cindex'} \mid \cindex, \cindex' \in \predset{}(x_\mathrm{test}), \cindex < \cindex' \}&&
 \mathrm{sc} &= \text{mean} (\bigcup_{x_\mathrm{test}\in \mathbb{D}_\mathrm{test}} \mathbb{K}(\predset{}(x_\mathrm{test})))\label{eq:setcoh}
\end{align}
It is defined as the mean over all the class similarities of classes that the predictor predicts jointly.
Thus, a higher value indicates sets with more similar classes in reality.\\
For brevity, we focus on a subset of more strongly affected metrics in the main paper but report results on all of them in the appendix alongside every metrics formulation in \ref{ssec:metrics}.

\subsection{Results}

\label{sec:pointRes}
This section discusses the main quantitative results of our proposed method.
Further qualitative examples are included in the
appendix.
\Cref{fig:globRed,fig:globNeck,sfig:C2Glob,sfig:C1Glob,sfig:I1Glob,sfig:I2Glob,sfig:sparrows} showcase global explanations, \cref{sfig:TeaserFull,sfig:window2,fig:globgraph1,fig:globgraph2,sfig:neckloc2,sfig:necklock2,sfig:spar1,sfig:spar2,sfig:mount1,sfig:mount2} include the novel local hierarchical explanations and \cref{sfig:BigTern,sfig:SmallTern,sfig:FeaturesBlack,sfig:FeaturesCover} demonstrate how the features of \hqpm{} are general concept detectors.
The accuracy as point predictor along the generally preferable qualities of Compactness, \contrastiveness{} and \cubsim{} is shown in \cref{tab:MainPointRes}.
%
\hqpm{} shows state-of-the-art accuracy for compact point predictors.
Further, it scores nearly perfectly on \contrastiveness{}.
\hqpm{} learns features that can be more clearly separated between active and inactive than even the class detectors of~\gls{PIP-Net}, indicating a gap between the ReLU-induced minimum of $0$ and the activations where a relevant concept is found.
The clear distinction between active and inactive enables our saliency maps, like in \cref{fig:TeaserFull,fig:TeaserGlob}, 
to also transport \textit{activation} rather than just \textit{location} without a reference test image and therefore enables extensive local explanations in practice.
The details are explained in \cref{ssec:SaliencyViz}.
Finally, \cubsim{}
quantifies that the additionally added pairs via \cref{eq:o4simCon} are also similar in reality and thus lead to more grounded class representations.
The state-of-the-art accuracy as point predictor paves the way for accurate set prediction along the hierarchical explanation, as the sets are conditioned on the predicted class. \\
For calibrating Conformal Prediction methods, the first $10$ test examples per class are split off into the calibration data \datacal.
That way, we can use the same models for evaluating point and set prediction.
\newt{Notably, applying Split Conformal Predictions requires exchangeability between calibration and test data. Our experimental setup is designed to ensure this exchangeability, as detailed in ~\cref{ssec:Limits}. }
As comparable CP methods, THR~\cite{thrpaper} and APS~\cite{apspaper} are used, as they are applicable without hyperparameters and broadly used~\cite{conformalINfo}.
\Cref{tab:confCComparison} compares our built-in CP method with these and also with the two simpler nonconformity scores \finscore{} and \upscore.
Evidently, our
proposed nonconformity score that restricts the sets to be constructed by going up the hierarchical local explanations shows competitive efficiency to THR for higher error rate $\alpha$ and approaches APS for lower values.
The reason can be seen when comparing our approach with the simpler \finscore{} that does not restrict the tree level to \adalvl{}:
With lower $\alpha$, the gap decreases, as \adalvl{} has to be set more loosely, allowing larger and therefore inefficient sets, which can be gauged from \cref{fig:introComp}.
\Cref{fig:SimPlot} visualizes the \setmetricsim{} (\cref{eq:setcoh}) of multiple models and CP methods and also indicates the set size for a desired $\alpha{}$.
As expected, \hqpm{} predicts the most coherent sets, as it moves up the hierarchical local explanation constructed from class representations with \cubsim{}.
\hqpm{} further clearly surpasses QPM in efficiency and \setmetricsim{}, making it the only model with efficient, calibrated, built-in, and interpretable coherent set prediction.

\subsection{Ablation Studies}
\label{sec:abls}
\begin{table}[t]
\caption{Impact of \floss{} on \cubheader{}. It effectively increases the Feature Alignment while 
reducing the fraction of nonzero features on the test data.
}
\label{tab:flossAbl}
\centering
\begin{tabular}{lccc}
\toprule
Method & Acc \arrowUp& \makecell{Feature\arrowUp\\Alignment}& \makecell{Feature \arrowDown\\Sparsity}  \\
\midrule
\tabOurs & \textbf{85.3}& \textbf{3.8} & \textbf{22.3} \\
w/o \floss & \textbf{85.3}  & 3.0  & 39.9  \\
and L1-Regularization &\textbf{85.3} & 3.4 & 31.3 \\
and L1 for  $\noclassfeatures$ & 85.1 & 3.6 & 26.4 \\
\bottomrule
\end{tabular}
\end{table}

\begin{wrapfigure}{R}{0.6\textwidth}
\vskip -0.38in
\includegraphics[width=\linewidth] {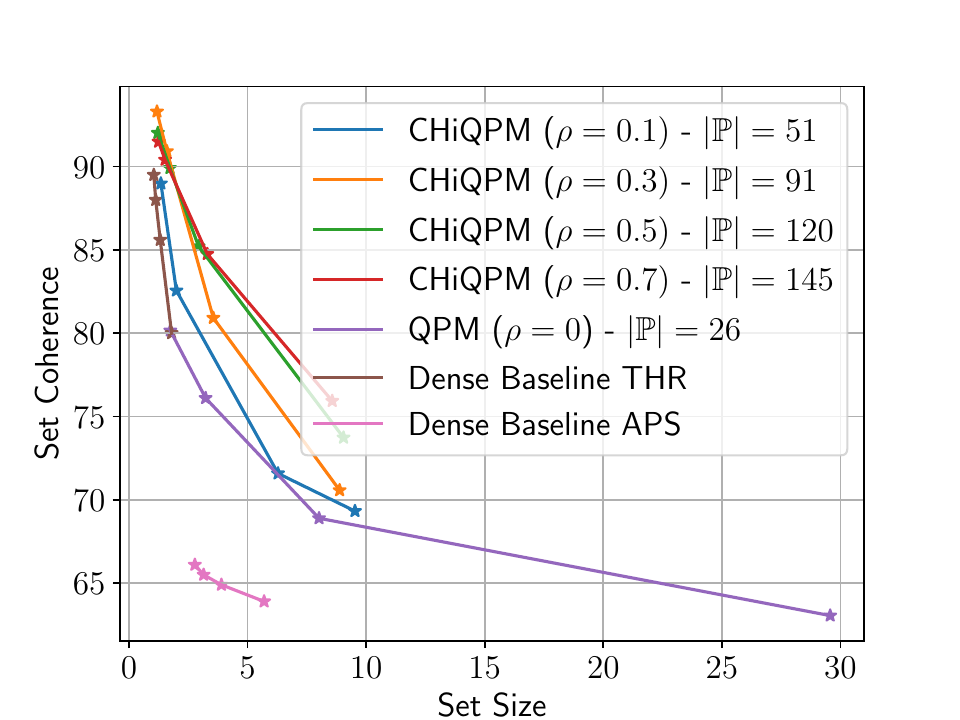} %
\caption{\setmetricsim{} on \cubheader{} for all classes predicted together. Across the set sizes, the built-in set prediction produces the most coherent sets. \confDots{} Note that an increased set size corresponds to a lower $\alpha$. Therefore, the competitive efficiency is also visible.}
\label{fig:SimPlot}
\end{wrapfigure}
This section investigates the impact of \treedense{}, \floss{} and summarizes the ablations shown in the appendix.
\Cref{fig:SimPlot} demonstrates how a higher \treedense{} leads to an increased $\lvert\CpairsSet\rvert$, more \setmetricsim{} and efficient sets.
For example, $\alpha = 0.075$ can be reached with 6.3 classes per prediction with $\treedense=0.1$ or 2.9 with $\treedense=0.5$, as \adalvl{} can be higher. 
Note that \treedense{} can be optimized on calibration data, steering the \hqpm{} towards efficiently reaching a desired $\alpha$, demonstrated in \cref{ssec:rhodense}.
Further visualizations that explicitly relate \treedense{} to the efficiency are included in the appendix, \cref{fig:conftreeacc,fig:hieraacc}. 
\newt{
The impact of \floss{} is shown in \cref{tab:flossAbl} and also contrasted with a $\mathcal{L}1$-regularization on the features $\textbf{f}^*$ as another form to induce sparse features.
The degree of regularization was chosen so that a further increase greatly reduces the accuracy.
Without sacrificing any accuracy, the proposed \hqpm{} reaches the highest Feature Alignment and best sparsity and thus validates 
\floss{}
as a suitable formulation to induce compact grounded hierarchical local explanations, with \cref{ssec:impactlambda} visualizing the impact of $\lambda_\mathrm{feat}$.
As further validation for our method, \cref{ssec:DynHiera} shows the benefit of using a sample-specific hierachy instead of a dynamic one. 
Additionally, \cref{ssec:GradCam} quantifies that CHiQPM's features quantifiably localize more on relevant regions in the image.
}
Finally, further ablations show a larger gap between \hqpm{} and QPM at even higher compactness (\cref{fig:dimVar}) and the strong positive impact of ReLU (\cref{ssec:ReluAbs}).

%% file: secs/conclusion.tex
\section{Conclusion}
This work introduces the Calibrated Hierarchical QPM (\hqpm{}). 
Faithfully following its
grounded globally interpretable class representations, \hqpm{} provides hierarchical local explanations. 
\hqpm{} is 
calibrated as a form of built-in interpretable Conformal Prediction to traverse the hierarchy at test time and predict a set of coherent classes, similar to how 
a human reasons,
which 
can be 
a step towards human-AI complementarity.
Finally, \hqpm{}'s improved global and additional novel form of local interpretability 
come with state-of-the-art accuracy as compact point predictor and efficiency on par with non-coherent set predictors even on \imgnetheader{}, ensuring broad applicability.

%% file: secs/checklist.tex
\section*{NeurIPS Paper Checklist}

\begin{enumerate}

\item {\bf Claims}
    \item[] Question: Do the main claims made in the abstract and introduction accurately reflect the paper's contributions and scope?
    \item[] Answer: \answerYes{} 
    \item[] Justification: The Built-In Hierarchical Explanations, \eg{} \cref{fig:TeaserFull}, are indeed novel. \Cref{sfig:avgsim} showcases improved set coherence and improved global interpretability via more class pairs, denoted in the legend. \cref{tab:MainPointRes} shows improvements in  interpetability metrics and performance as point predictor and \cref{tab:confCComparison} the competitive performance as interpretable conformal predictor. \Cref{tab:flossAbl} demonstrates the positive impact of the novel Feature Grounding Loss.

\item {\bf Limitations}
    \item[] Question: Does the paper discuss the limitations of the work performed by the authors?
    \item[] Answer: \answerYes{} 
    \item[] Justification: That is discussed in \cref{ssec:Limits}.

\item {\bf Theory assumptions and proofs}
    \item[] Question: For each theoretical result, does the paper provide the full set of assumptions and a complete (and correct) proof?
    \item[] Answer: \answerNA{} 
    \item[] Justification: There is no theoretical result. Conformal Prediction only assumes exchangeability which we mention in \cref{sec:introconf}.

    \item {\bf Experimental result reproducibility}
    \item[] Question: Does the paper fully disclose all the information needed to reproduce the main experimental results of the paper to the extent that it affects the main claims and/or conclusions of the paper (regardless of whether the code and data are provided or not)?
    \item[] Answer: \answerYes{} 
    \item[] Justification: The paper extends QPM, which offers published code (\url{https://github.com/ThomasNorr/QPM}). The methodical extensions of \cref{eq:o4sim} are straightforward to integrate into QPM and all parameters are clearly described. Similarly, the proposed nonconformity score of \cref{eq:adalvl} can be directly used with \textit{torchcp} for use in conformal prediction. The code will be published after acceptance and all used parameters are clearly described in this work.

\item {\bf Open access to data and code}
    \item[] Question: Does the paper provide open access to the data and code, with sufficient instructions to faithfully reproduce the main experimental results, as described in supplemental material?
    \item[] Answer: \answerNo{} 
    \item[] Justification: Unfortuntately, we are unable to submit the code alognside the paper, but will release it after acceptance. We want to emphasize that the experiments are clearly described modifications to published code.

\item {\bf Experimental setting/details}
    \item[] Question: Does the paper specify all the training and test details (e.g., data splits, hyperparameters, how they were chosen, type of optimizer, etc.) necessary to understand the results?
    \item[] Answer: \answerYes{} 
    \item[] Justification: The main paper includes the novel parameters \treedense{} and $\lambda_\mathrm{feat}$ and why their values are chosen in \cref{sec:exp} and describes that all other parameters mirror the previous work QPM, with the only differences being outlined in \cref{ssec:impdeta}. The datasets all have a default split which is used.

\item {\bf Experiment statistical significance}
    \item[] Question: Does the paper report error bars suitably and correctly defined or other appropriate information about the statistical significance of the experiments?
    \item[] Answer: \answerYes{} 
    \item[] Justification: Due to space limitations, we do not report standard deviations in the main paper, but include standard deviations for the tables with main results in the appendix. \Cref{ssec:Results} details what variability they capture and how they are computed.

\item {\bf Experiments compute resources}
    \item[] Question: For each experiment, does the paper provide sufficient information on the computer resources (type of compute workers, memory, time of execution) needed to reproduce the experiments?
    \item[] Answer: \answerYes{} 
    \item[] Justification: This is discussed in \cref{sec:computeRes}.
    
\item {\bf Code of ethics}
    \item[] Question: Does the research conducted in the paper conform, in every respect, with the NeurIPS Code of Ethics \url{https://neurips.cc/public/EthicsGuidelines}?
    \item[] Answer: \answerYes{} 
    \item[] Justification: We adhered to the Code of Ethics.

\item {\bf Broader impacts}
    \item[] Question: Does the paper discuss both potential positive societal impacts and negative societal impacts of the work performed?
    \item[] Answer: \answerYes{} 
    \item[] Justification: That is discussed in \cref{ssec:BroaderImpact}.
    
\item {\bf Safeguards}
    \item[] Question: Does the paper describe safeguards that have been put in place for responsible release of data or models that have a high risk for misuse (e.g., pretrained language models, image generators, or scraped datasets)?
    \item[] Answer: \answerNA{} 
    \item[] Justification: Our model does not pose such risks, but its interpretability enables a thorough inspection of it before release.

\item {\bf Licenses for existing assets}
    \item[] Question: Are the creators or original owners of assets (e.g., code, data, models), used in the paper, properly credited and are the license and terms of use explicitly mentioned and properly respected?
    \item[] Answer: \answerYes{} 
    \item[] Justification: We follow standard practices and cite every  paper whose datasets we use, as well as the code packages we used. We were unable to find the correct license for any of the used datasets.

\item {\bf New assets}
    \item[] Question: Are new assets introduced in the paper well documented and is the documentation provided alongside the assets?
    \item[] Answer: \answerNA{} 
    \item[] Justification: No new assets.

\item {\bf Crowdsourcing and research with human subjects}
    \item[] Question: For crowdsourcing experiments and research with human subjects, does the paper include the full text of instructions given to participants and screenshots, if applicable, as well as details about compensation (if any)? 
  \item[] Answer: \answerNA{} 
    \item[] Justification: No research with human subjects.

\item {\bf Institutional review board (IRB) approvals or equivalent for research with human subjects}
    \item[] Question: Does the paper describe potential risks incurred by study participants, whether such risks were disclosed to the subjects, and whether Institutional Review Board (IRB) approvals (or an equivalent approval/review based on the requirements of your country or institution) were obtained?
    \item[] Answer: \answerNA{} 
    \item[] Justification: No research with human subjects.

\item {\bf Declaration of LLM usage}
    \item[] Question: Does the paper describe the usage of LLMs if it is an important, original, or non-standard component of the core methods in this research? Note that if the LLM is used only for writing, editing, or formatting purposes and does not impact the core methodology, scientific rigorousness, or originality of the research, declaration is not required.
    \item[] Answer: \answerNA{} 
    \item[] Justification: Only used for writing, editing, or formatting purposes. 

\end{enumerate}

%% file: secs/appendix.tex
\newpage
\appendix
\section{Implementation Details}
\label{ssec:impdeta}
For implementing CP, we utilized the \textit{torchcp}~\cite{wei2024torchcp} package and implemented the model using  \textit{Pytorch}~\cite{Pytorch}.
Note that all details will also be clear in the published code.

The QP is solved as described, with the additional constraints from \cref{sec:constr}, but with two relaxations based on observations:
First, the MIP-Gap of the discrete optimization in \gurobi{} can be relaxed without significant effect on the resulting metrics, hence we set it to $1\%$.
Second, due to the high number of classes, $\gls{nClasses} = 1000$, the resulting assignment for \imgnetheader{} contains more pairs $\lvert\CpairsSet\rvert$ than desired anyway. Therefore, we do not add any pairwise constraints to the QP for \imgnetheader{}.
After solving the QP, we further fine-tune the model for $70$ epochs using the learning rate schedule from Q-SENN, as we observed significantly larger changes in the features during the fine-tuning as opposed to QPM\newt{, as quantified in \cref{ssec:QPSolution}}. 
The longer training with a higher learning rate facilitates that the features can be reorganized according to the imposed hierarchical structure.
We set $\lambda_\mathrm{feat}=3$, but skip this loss during the final $10$ epochs, as its desired effects are induced without reducing top-1 Accuracy. All other steps of the pipeline equal QPM~\cite{qpmPaper}. \newt{By mistake, the subtrahend in \cref{eq:scaleFloss} was scaled by a factor of $2$ in all experiments}. We repeated the experiments in \cref{ssec:impactlambda} with the correct implementation and observed the expected results with negligible impact of the mistake. Therefore, we did not rerun every experiment. 

Note that for accuracy as point-predictor and for \predclass{} for conformal prediction, the $\mathrm{argmax}$ of \gls{outputVector} is considered the predicted class, leading to the lower index in case of a tie.

In order to apply our novel CP to QPM, we subtract the minimum of all features of the features, ensuring that reasoning is restricted down the hierarchical explanation and unaffected by slightly varying \muf.

\section{Metrics}
\label{ssec:metrics}
This section defines the metrics that are not introduced in this work for the first time.

Following QPM~\cite{qpmPaper}, we measure the general qualities of learning diverse, general and contrastive features via \loc{5},
    \begin{align}
\hat{M}^{\findex}_{i,j} = \frac{M^{\findex}_{i,j}}{\frac{1}{\gls{featuresMapwidth}\gls{featuresMapheigth}} \sum |\textbf{M}^{\findex}|} \quad \hat{S}^{\findex}_{i,j} = \frac{e^{\hat{M}^{\findex}_{i,j}}}{\sum_{m,n} e^{\hat{M}^{\findex}_{m,n}}} \\
   \loc{5} = \frac{\sum_{i=1}^{\gls{featuresMapheigth}}\sum_{j=1}^{\gls{featuresMapwidth}}\max(\hat{S}^{1}_{i,j},\hat{S}^{2}_{i,j}, \dots, \hat{S}^{5}_{i,j})}{5} , 
\end{align}
\generality{} $\tau$,
\begin{align}
    \tau = 1-\frac{1}{\gls{nReducedFeatures}} \sum_{\findex=1}^{\gls{nReducedFeatures}} \max_\cindex
&\frac{\sum_{j=1}^{\gls{nTrainImages}} l^\cindex_j (f_{j,\findex} - \min \boldsymbol{f}_{:,\findex})}{\sum_{j=1}^{\gls{nTrainImages}} (f_{j,\findex}- \min \boldsymbol{f}_{:,\findex})}
\end{align}
and \contrastiveness{}:
\begin{equation}
  \mathrm{\contrastiveness{}} =\sum_{\findex=1}^{\gls{nReducedFeatures}} 1 - \mathrm{Overlap}(\mathcal{N}_1^\findex, \mathcal{N}_2^\findex)\label{eq:contra}
  \end{equation}
\loc{5} describes how diversely the top 5 weighted features activate for an average test sample. 
\generality{} measures which fraction of each features' activation does not activate on its most related class and a low \generality{} indicates that features are class-detectors.
Finally, \contrastiveness{} measures the overlap between two Gaussian distributions, parametrized by $\mu_1^\findex{},\sigma_1^\findex{}, \mu_2^\findex{},\sigma_2^\findex{},  $ fit into the feature distribution $\boldsymbol{f}_{:,\findex}$.
A high \contrastiveness{} indicates features that can be split into activating and not activating.

Using the computed $\boldsymbol{\mathrm{\classSim}^{gt}_{\cindex,\cindex'}}$, \cubsim{}~\cite{qpmPaper} quantifies how similar the top $25$ most similar classes in reality, $\boldsymbol{\mathrm{\ClassSim}}^{gt} = \boldsymbol{\Lambda}\boldsymbol{\Lambda}^T$, 
are represented in the class representations $\boldsymbol{\mathrm{\ClassSim}}^{Model} = \wgurobi \wgurobi^T$:
\begin{equation}
  \mathrm{\cubsim} = 
  \frac{\sum_{\cindex,\cindex' \in C_\mathrm{Sim} } \mathrm{\classSim}^{Model}_{\cindex,\cindex'} }{\sum_{\cindex,\cindex' \in C_\mathrm{Sim} } \mathrm{\classSim}^{gt}_{\cindex,\cindex'}}.
\end{equation}

The Feature Alignment $r$~\cite{norrenbrock2024q} between the learned features and these attributes is also measured to estimate if the features \glspl{trainFeatures} are more human-like:
\begin{align}
\gtmatrix_{a,j}&=
\frac{1}{\lvert\attributeset{a+}\rvert }\sum_{i\in\attributeset{a+}}\glspl{trainFeatures}_{i,j}- \frac{1}{\lvert\attributeset{a-}\rvert}\sum_{i\in\attributeset{a-}}\glspl{trainFeatures}_{i,j} \\
    r &= \frac{1}{\gls{nReducedFeatures}}\sum_{j=1}^{\gls{nReducedFeatures}} \frac{\gls{nTrainImages}}{\sum_{l=1}^{\gls{nTrainImages}}\glspl{trainFeatures}_{l,j}- \min_{l}\glspl{trainFeatures}_{l,j} }\max_{i} \gtmatrix_{i,j}\quad.
\end{align}
\section{Visualization Details}
\label{ssec:VizDetails}

\subsection{Graph Visualization}
\label{ssec:GraphViz}
This section discusses the details of how the hierarchical explanations in the form of graphs are visualized, \eg for \cref{fig:TeaserFull}, 
and also the global hierarchical explanations in \cref{fig:globgraph1,fig:globgraph2}.
Generally, the content of the graphs is explained in \cref{sec:hierexp}.
To indicate the activation of each feature $i$, the radius of each feature node scales with its fraction of the maximum activation $r_i = \frac{f_i^*}{\mathrm{max}(f^*)} r_\mathrm{scale}$, where $r_\mathrm{scale}$ is a constant.

Using colors, we add additional information to the graphs.
The nodes are colored according to the saliency maps, that are used in the global explanation and also as part of the local explanation.
All nodes, that correspond to features not visualized as saliency map, are represented in gray.
Further, we use color to indicate the actual prediction done. Only those classes are predicted by \hqpm{}, for which a colored path coming from the root of the graph connects to the class nodes.


Finally, the class labels at each node are summarised if multiple classes are at the same node, i.e.\, sharing the same top-$n$ features, but not having any activation on their other assigned features.
To summarise the class names, the text description "\textit{ClassName, + x}" is used.
Here \textit{ClassName} is chosen from the classes at that node, usually the name with the shortest words to reduce horizontal overlap, and \textit{x} refers to the number of classes not called \textit{ClassName} at this node.
\subsection{Visualizing saliency maps}
\label{ssec:SaliencyViz}
This section explains how the saliency maps for the individual features as part of global and local explanation are visualized.

Typical saliency map methods and available implementations~\cite{jacobgilpytorchcam}, like GradCAM~\cite{selvaraju2017grad}, only focus on showing where support for a decision is found.
As they are used to showing why a decision was made, they do not have to transport activation.
Therefore, they overlay the computed saliency map $\gls{SaliencyMap}$ on the image \gls{ImageSample} scaled to the full range of values irrespective of activation to get the Saliency explanation \gls{SaliencyOverlay}:
\begin{align}
   \gls{ImageSample}^\text{NoActivation} &= \text{ColorMap}(\frac{\gls{SaliencyMap}}{\text{max}(\gls{SaliencyMap})} )\beta+ (1-\beta) \gls{ImageSample} \\
     \gls{SaliencyOverlay} &= \frac{\gls{ImageSample}^\text{NoActivation}}{\text{max}(\gls{ImageSample}^\text{NoActivation})} 
\end{align}
Here, \text{ColorMap} converts the spatial information into RGB values and $\beta$ defines a weighting between image and saliency map.
While typically \textit{jet} is used, we use individual colors in this work, scaled using gamma correction to match the impression of \textit{jet}, as we decompose the decision process into detecting individual features or concepts.
Notably, different saliency methods differ in how they compute \gls{SaliencyMap}.

Similar to GradCAM, we use the feature maps \glsfirst{featureMapsSmall} as the basis for our visualization, as they directly cause the activations our model uses and are also evaluated and steered to be diverse.
However, we visualize the map \gls{singleMap} for every feature \findex{} individually rather than a weighted sum because every feature can and should be interpreted on its own.

Different to prior work, \hqpm{} has near perfect contrastive features, as measured by \contrastiveness{}.
That means that for a concrete sample, the feature \findex{} can be clearly sorted into the inactive distribution $\mathcal{N}_1^\findex$ or 
 the active distribution $\mathcal{N}_2^\findex$.
We make use of that distinction and especially the mean activation of active samples \activeMean{} to produce saliency maps that indeed already transport whether the feature is activated or not. 
Specifically, the Saliency explanation for a single feature \findex{} is computed following:
\begin{align}
   \gls{ImageSample}^\text{Activation}_\findex &= \text{ColorMap}(\frac{\gls{singleMap}}{\text{max}(\gls{singleMap})}min(\frac{f^*_\findex}{\activeMean}, 1))\beta+ (1-\beta) \gls{ImageSample} \\
     \gls{SaliencyOverlay}_\findex&= \frac{\gls{ImageSample}^\text{Activation}_\findex}{\text{max}(\gls{ImageSample}^\text{Activation}_\findex)} 
\end{align}
Here, the term $min(\frac{f^*_\findex}{\activeMean}, 1)$ scales the saliency map based on how active the feature is. 
Due to the perfect \contrastiveness{}{}, the threshold \activeMean{} is a good indicator how objectively active the feature is.
Notably, this form of transporting confidence can be applied to a single test image, as the active threshold \activeMean{} is computed on training data.
This is in contrast to how QPM~\cite{qpmPaper} scaled activations for the same features across images.

In our formulas, we omit how all feature maps are linearly scaled from their feature map size to the input image size.
\section{Impact of Tree Density \treedense}
\label{ssec:rhodense}
\begin{figure}[ht]
\vskip 0.2in
\begin{center}
\centerline{\includegraphics[width=.5\columnwidth]{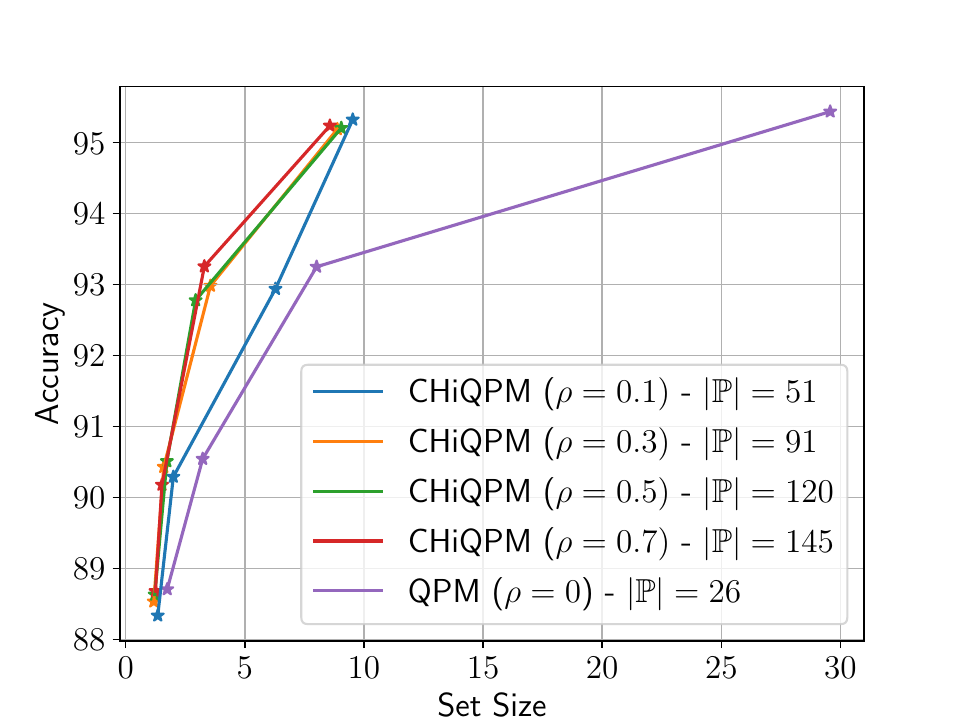}}
\caption{Accuracy over set size using the built-in conformal prediction for varying \treedense. \confDots{} Better fixed level performance, shown in \cref{fig:hieraacc}, leads to more efficient calibrated sets due to being able to limit the depth via \adalvl.}
\label{fig:conftreeacc}
\end{center}
\vskip -0.2in
\end{figure}
\begin{figure}[ht]
\vskip 0.2in
\begin{center}
\centerline{\includegraphics[width=.5\columnwidth]{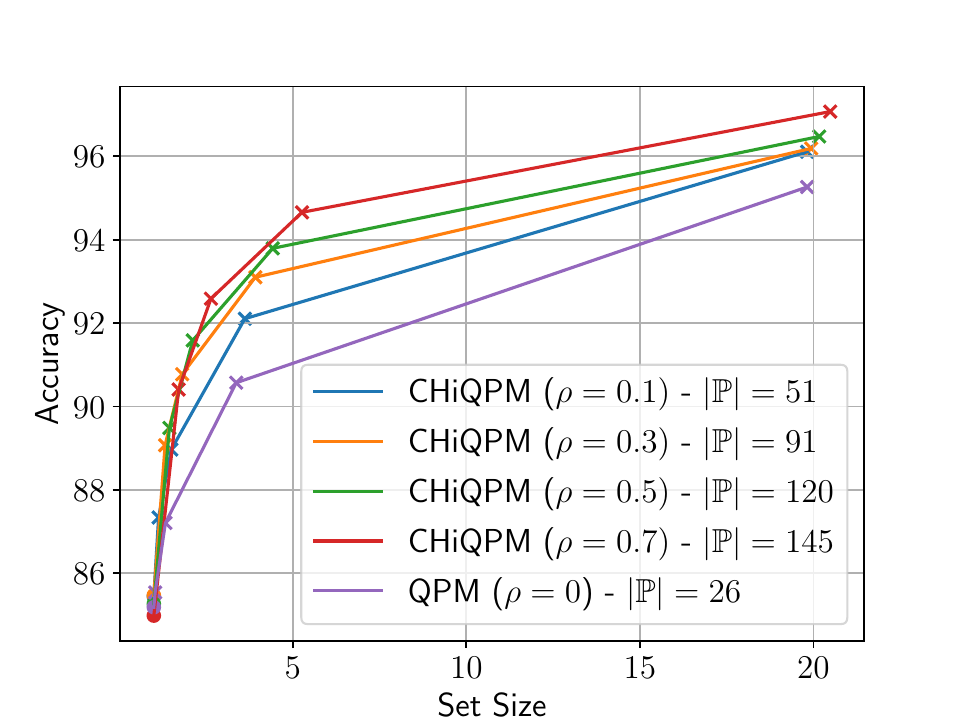}}
\caption{Set Accuracy of \hqpm{} with $\gls{nperClass}=5$ on CUB, when predicting with a fixed level in the sample specific hierarchy. Each mark represents one level.}
\label{fig:hieraacc}
\end{center}
\vskip -0.2in
\end{figure}
This section discusses the impact of \treedense{} and provides the detailed relation between \treedense{} and the efficiency of the hierarchical and calibrated predictions.
As noted in \cref{sec:abls}, \Cref{fig:SimPlot} demonstrates how a higher \treedense{} leads to an increased $\lvert\CpairsSet\rvert$, more \setmetricsim{} and efficient sets.
\cref{fig:hieraacc,fig:conftreeacc} explicitly relate \treedense{} to the efficiency of the resulting predictor and visualize the relationship between fixed level efficiency in \cref{fig:hieraacc} and calibrated efficiency in \cref{fig:conftreeacc}.
As shown in  \cref{fig:hieraacc}, in order to reach $\alpha = 0.075$, \hqpm{} can typically be limited to $\adalvl{}=2$.
However, for $\treedense= 0.1$, it does not reach $92.5\%$ accuracy with a fixed $n=2$ and therefore has to be calibrated with $\adalvl{}=1$.
This causes the significant increase in Set Size and therefore inefficiency for $\alpha = 0.075$, from 2.9 with the default configuration of   $\treedense= 0.5$ to 6.3 with $\treedense=0.1$, as clearly visualized in \cref{fig:conftreeacc}.
Therefore, optimizing \treedense{} using calibration data for a desired $\alpha$ is an effective option. 
However, note that the point-predicting performance starts to deteriorate slightly when further increasing \treedense{} beyond $\treedense=0.5$, as indicated by the start of the graphs in \cref{fig:hieraacc}.
This is likely due to classes being forced to be represented very similarly, even though they do not share $\gls{nperClass}-1$ general concepts.
The $\lvert\CpairsSet\rvert$ is further evidence for that, as the number of unenforced pairs, $\lvert\CpairsSet\rvert - \treedense * \gls{nClasses}$, declines, with just $5$ for $\treedense=0.7$.
Thus, the optimal value for \treedense{} also depends on the similarities of the classes in the dataset.



\section{Detailed Results}
\label{ssec:Results}
\begin{table}[h]
\caption{Impact of Relu and Normalization on \cubheader{} with \resnet{}}
\label{stab:ablNormRel}
\centering
\begin{tabular}{lcccc}
\toprule
Method & Acc &  C-I & Contrast. & Feature Alignment\\
\midrule
\tabOurs & \textbf{85.3 }& 94.1 & \textbf{99.9} & \textbf{3.8} \\
\midrule
w/o  \floss{} and ReLU  & 84.8 & \textbf{96.7} & 97.1 & 1.9\\
w/o Norm &  83.9 & 95.1 &  96.2 & 3.2\\

\bottomrule
\end{tabular}
\end{table}

This section contains more results to demonstrate the quantifiable high performance of our \hqpm{}.
\begin{figure}[t]
\vskip 0.2in
\begin{center}
\centerline{\includegraphics[width=.5\columnwidth]{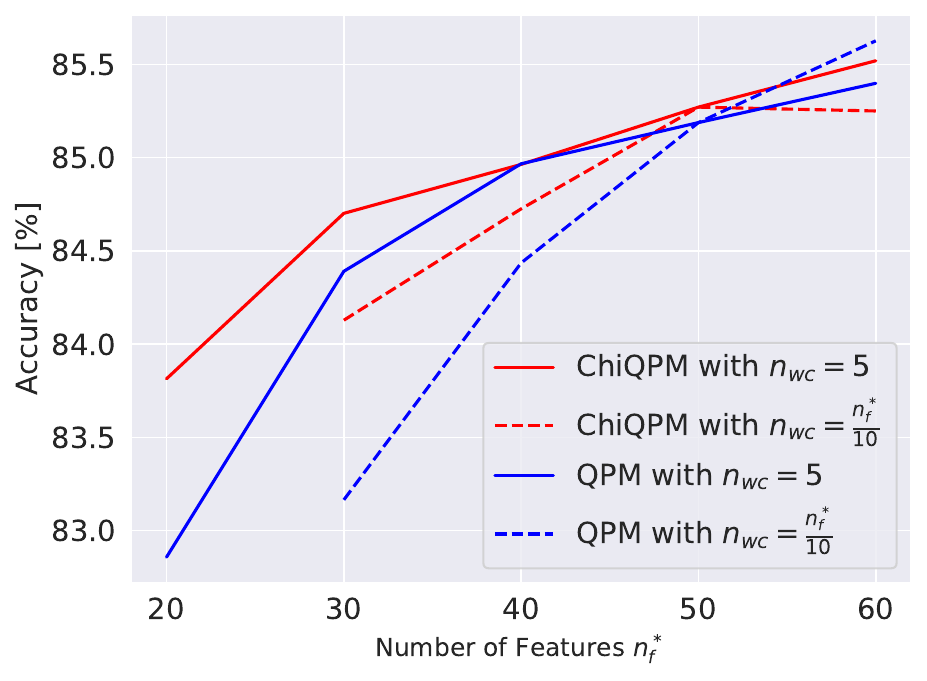}}
\caption{Accuracy on CUB in relation to Compactness: \hqpm{} uses the allowed low compactnesss more efficiently, both when scaling \gls{nperClass} with \gls{nReducedFeatures} and when not.
When \gls{nReducedFeatures} becomes higher, restricting $\gls{nperClass} =\frac{ \gls{nReducedFeatures}}{10}$ leads to a less efficient use of some features.}
\label{fig:dimVar}
\end{center}
\vskip -0.2in
\end{figure}

\begin{figure}[ht]
\vskip 0.2in
\begin{center}
\centerline{\includegraphics[width=.5\columnwidth]{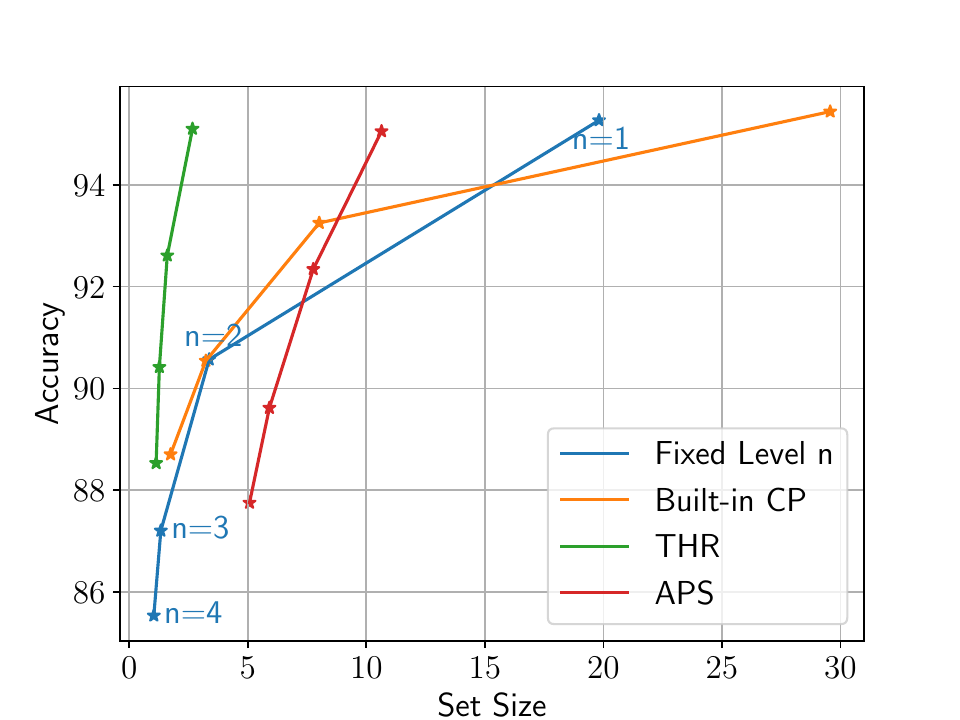}}
\caption{Results for QPM on \cubheader{} for different CP methods applied to it, comparable to \cref{fig:introComp}. \hqpm{} significantly boots the efficiency of the built-in CP.}
\label{fig:QPMintroComp}
\end{center}
\vskip -0.2in
\end{figure}

\begin{figure}[ht]
\vskip 0.2in
\begin{center}
\centerline{\includegraphics[width=.5\columnwidth]{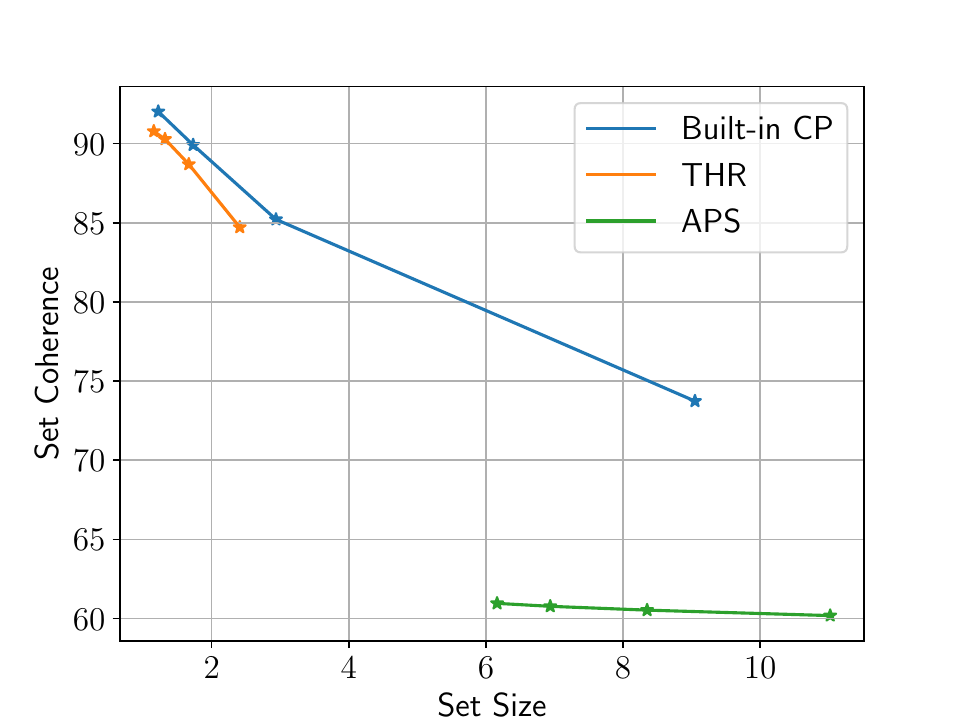}}
\caption{Average Similarity compared to other conformal prediction methods for the same \hqpm{}. \confDots{}}
\label{sfig:avgsim}
\end{center}
\vskip -0.2in
\end{figure}

First, \cref{fig:QPMintroComp} visualizes how QPM performs with different CP methods. 
When comparing it to \cref{fig:introComp}, \hqpm{}'s drastic improvement is evident, making interpretable coherent set prediction competitively efficient.
Further, \cref{sfig:avgsim} demonstrates how the built-in interpretable coherent set prediction predicts more coherent sets than conventional CP methods, even for the same \hqpm{}.
However, it is notable, that THR predicts significantly more coherent sets in that scenario, as it is applied to our \hqpm{} with \cubsim{}.

\begin{figure}[ht]
\vskip 0.2in
\begin{center}
\centerline{\includegraphics[width=\columnwidth]{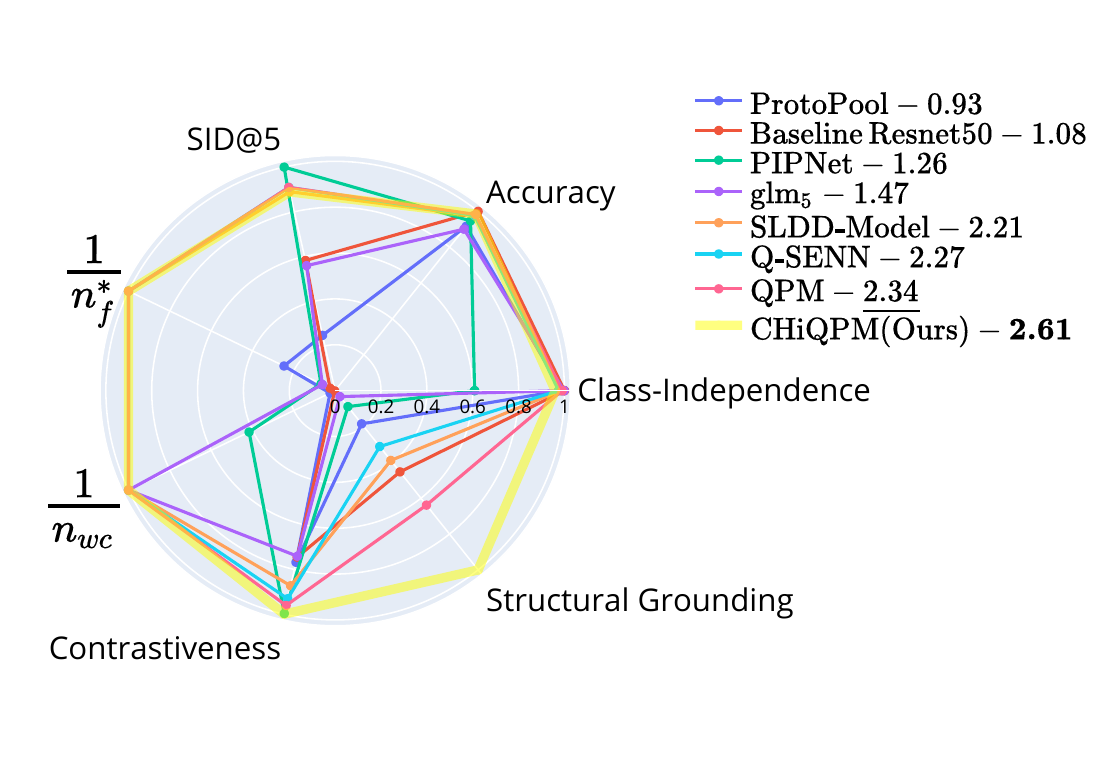}}
  \caption{Radar plot across the QPM metrics for \cubheader{} on \resnet{}. Every value is transformed to a fraction of the maximum, following QPM~\cite{qpmPaper}. }
\label{sfig:radarplot}
\end{center}
\vskip -0.2in
\end{figure}

Finally, we include the results on \resnet{}, Resnet34, Swin-Transformer-small~\cite{liu2021swin} and \gls{incv} in \cref{stab:res50Interp,stab:aCCproto-table,stab:r34aCCproto-table,stab:r34Interpproto-table,stab:incaCCproto-table,stab:incInterpproto-table,sstab:swcaCCproto-table,sstab:swInterpproto-table-adapted} with their standard deviations, computed as \textit{np.std(x)}, where x are the individual results across the different seeds. The CP efficiency  across architectures with standard deviations is shown in \cref{stab:inc,stab:r34,stab:sw,stab:r50,stab:r50img}. 
These tables demonstrate how our state-of-the-art performance is architecture independent.
Note that we recreated the Q-SENN results on \cubheader{} and \stanfordheader{} for these tables in order to get standard deviations and results on more architectures but included the reported results in \citet{qpmPaper} in the main paper.
Notably \loc{5} has slightly decreased due to the introduction of \floss{}.
Usually, \loc{5} is high due to the \gls{customLoss}. 
With the addition of \floss{}, especially on \imgnetheader{}, its relative weighting is slightly lower. 
Therefore, one might want to increase the weighting for it for \imgnetheader{}.
For easier comparison and compute limitations, we did not optimize the weighting of \gls{customLoss} at all and left it at the weight used by \gls{layerName}, \gls{qsenn} and QPM~\cite{qpmPaper}.
However, initial experiments show that one can increase it without significant impact on accuracy, shown in \cref{stab:diversity}.
The strong performance of \hqpm{} across all the criteria is also summarized in the radar plot in \cref{sfig:radarplot}, which follows QPM and thus only includes its metrics. 
Even ignoring the excellent Feature Alignment, Sparsity and novel hierarchical explanations with calibrated coherent set predictions, \hqpm{} clearly sets the state of the art.

\begin{table*}[h]
\caption{Average Set Size $\lvert\predset{}\rvert$ of \hqpm{}  with Resnet50  calibrated to reach various coverages $1-\alpha$ comparing different conformal prediction methods. All methods are very close or reach the desired coverage.}
\label{stab:r50}
\centering
\resizebox{\linewidth}{!}{
\begin{tabular}{lcccccccc}
\toprule
{$\lvert\predset{}\rvert$\arrowDown} & \multicolumn{4}{c}{CUB} & \multicolumn{4}{c}{CARS}  \\\cmidrule(lr){1-1}
\cmidrule(lr){2-5} \cmidrule(lr){6-9} 
Method & $\alpha$=0.12 & $\alpha$=0.1 & $\alpha$=0.075 & $\alpha$=0.05 & 
$\alpha$=0.075 & $\alpha$=0.05 & $\alpha$=0.0025 & $\alpha$=0.001  \\
\midrule
\ours & 1.22 $\pm$ 0.04 & 1.73 $\pm$ 0.08 & 2.94 $\pm$ 0.17 & 9.05 $\pm$ 0.20 & 1.05 $\pm$ 0.02 & 1.25 $\pm$ 0.14 & 8.25 $\pm$ 0.38 & 74.4 $\pm$ 1.83 \\ 
\midrule
THR&1.16 $\pm$ 0.04 & 1.32 $\pm$ 0.05 & 1.67 $\pm$ 0.08 & 2.41 $\pm$ 0.12 & 1.02 $\pm$ 0.01 & 1.15 $\pm$ 0.02 & 2.09 $\pm$ 0.16 & 6.93 $\pm$ 1.49 \\
APS & 6.30$\pm$ 0.41 & 7.20 $\pm$ 0.65 & 8.54 $\pm$ 0.50 & 11.3 $\pm$ 0.96 & 5.64 $\pm$ 0.30 & 6.83 $\pm$ 0.55 & 9.61 $\pm$ 0.55 & 18.6 $\pm$ 1.94 \\
\bottomrule
\end{tabular}
}
\end{table*}
\begin{table*}[h]
\caption{Average Set Size $\lvert\predset{}\rvert$ of \hqpm{}  with Resnet50 on \imgnetheader{}  calibrated to reach various coverages $1-\alpha$ comparing different conformal prediction methods. All methods are very close or reach the desired coverage.}
\label{stab:r50img}
\centering
\begin{tabular}{lcccc}
\toprule
{$\lvert\predset{}\rvert$\arrowDown} & \multicolumn{4}{c}{IMG}   \\\cmidrule(lr){1-1}
\cmidrule(lr){2-5} 
Method & $\alpha$=0.22 & $\alpha$=0.2 & $\alpha$=0.175 & $\alpha$=0.15   \\
\midrule
\ours & 1.10 $\pm$ 0.01 & 1.42 $\pm$ 0.02 & 3.25 $\pm$ 0.07 & 4.58 $\pm$ 0.11 \\ 
\midrule
THR&1.05 $\pm$ 0.00 & 1.16 $\pm$ 0.01 & 1.40 $\pm$ 0.01 & 1.87 $\pm$ 0.03 \\
APS & 16.7 $\pm$ 0.33 & 18.9 $\pm$ 0.28 & 22.1 $\pm$ 0.49 & 26.8 $\pm$ 0.504 \\
\bottomrule
\end{tabular}
\end{table*}
\begin{table*}[h]
\caption{Average Set Size $\lvert\predset{}\rvert$ of \hqpm{}  with Resnet34  calibrated to reach various coverages $1-\alpha$ comparing different conformal prediction methods. All methods are very close or reach the desired coverage.}
\label{stab:r34}
\centering
\resizebox{\linewidth}{!}{
\begin{tabular}{lcccccccc}
\toprule
{$\lvert\predset{}\rvert$\arrowDown} & \multicolumn{4}{c}{CUB} & \multicolumn{4}{c}{CARS}  \\\cmidrule(lr){1-1}
\cmidrule(lr){2-5} \cmidrule(lr){6-9} 
Method & $\alpha$=0.12 & $\alpha$=0.1 & $\alpha$=0.075 & $\alpha$=0.05 & 
$\alpha$=0.075 & $\alpha$=0.05 & $\alpha$=0.0025 & $\alpha$=0.001  \\
\midrule
\ours & 1.45 $\pm$ 0.19 & 2.22 $\pm$ 0.37 & 6.65 $\pm$ 1.96 & 26.8 $\pm$ 22.5 & 1.05 $\pm$ 0.01 & 1.47 $\pm$ 0.18 & 12.6 $\pm$ 1.19 & 96.0 $\pm$ 4.36 \\ 
\midrule
THR&1.25 $\pm$ 0.04 & 1.44 $\pm$ 0.02 & 1.88 $\pm$ 0.05 & 2.9 $\pm$ 0.22 & 1.03 $\pm$ 0.02 & 1.22 $\pm$ 0.06 & 2.57 $\pm$ 0.09 & 9.50 $\pm$ 2.15 \\
APS & 6.97 $\pm$ 0.45 & 7.88 $\pm$ 0.39 & 9.80 $\pm$ 0.62 & 12.9 $\pm$ 0.63 & 6.13 $\pm$ 0.69 & 8.20 $\pm$ 1.00 & 12.0 $\pm$ 0.86 & 21.5 $\pm$ 3.18 \\
\bottomrule
\end{tabular}
}
\end{table*}
\begin{table*}[h]
\caption{Average Set Size $\lvert\predset{}\rvert$ of \hqpm{} with \gls{incv} calibrated to reach various coverages $1-\alpha$ comparing different conformal prediction methods. All methods are very close or reach the desired coverage.}
\label{stab:inc}
\centering
\resizebox{\linewidth}{!}{
\begin{tabular}{lcccccccc}
\toprule
{$\lvert\predset{}\rvert$\arrowDown} & \multicolumn{4}{c}{CUB} & \multicolumn{4}{c}{CARS}  \\ \cmidrule(lr){1-1}
\cmidrule(lr){2-5} \cmidrule(lr){6-9} 
Method & $\alpha$=0.12 & $\alpha$=0.1 & $\alpha$=0.075 & $\alpha$=0.05 & 
$\alpha$=0.075 & $\alpha$=0.05 & $\alpha$=0.0025 & $\alpha$=0.001  \\
\midrule
\ours & 1.52 $\pm$ 0.05 & 2.34 $\pm$ 0.12 & 7.63 $\pm$ 0.15 & 61.5 $\pm$ 4.25 & 1.02 $\pm$ 0.01 & 1.47 $\pm$ 0.22 & 47.0 $\pm$ 22.7 & 81.8 $\pm$ 3.04\\ 
\midrule
THR& 1.24 $\pm$ 0.04 & 1.47 $\pm$ 0.07 & 2.09 $\pm$ 0.10 & 4.07 $\pm$ 0.38 & 1.00 $\pm$ 0.01 & 1.17 $\pm$ 0.04 & 3.17 $\pm$ 0.39 & 13.5 $\pm$ 2.68 \\
APS &7.08 $\pm$ 0.44 & 7.94 $\pm$ 0.65 & 10.5 $\pm$ 0.96 & 13.8 $\pm$ 1.35 & 5.14 $\pm$ 0.30 & 6.77 $\pm$ 0.76 & 10.8 $\pm$ 0.77 & 21.1 $\pm$ 1.82
  \\
\bottomrule
\end{tabular}
}
\end{table*}
\begin{table*}[h]
\caption{Average Set Size $\lvert\predset{}\rvert$ of \hqpm{} with Swin-Transformer-Small calibrated to reach various coverages $1-\alpha$ comparing different conformal prediction methods. All methods are very close or reach the desired coverage.}
\label{stab:sw}
\centering
\resizebox{\linewidth}{!}{
\begin{tabular}{lcccccccc}
\toprule
{$\lvert\predset{}\rvert$\arrowDown} & \multicolumn{4}{c}{CUB} & \multicolumn{4}{c}{CARS}  \\ \cmidrule(lr){1-1}
\cmidrule(lr){2-5} \cmidrule(lr){6-9} 
Method & $\alpha$=0.12 & $\alpha$=0.1 & $\alpha$=0.075 & $\alpha$=0.05 & 
$\alpha$=0.075 & $\alpha$=0.05 & $\alpha$=0.0025 & $\alpha$=0.001  \\
\midrule
\ours & 1.14 $\pm$ 0.05 & 1.44 $\pm$ 0.21 & 2.44 $\pm$ 0.32 & 8.09 $\pm$ 0.77 & 1.28 $\pm$ 0.24 & 3.3 $\pm$ 2.93 & 20.2 $\pm$ 25.6 & 75.0 $\pm$ 3.14\\ 
\midrule
THR&	1.09 $\pm$ 0.01 & 1.22 $\pm$ 0.03 & 1.49 $\pm$ 0.04 & 2.24 $\pm$ 0.13 & 1.15 $\pm$ 0.13 & 1.42 $\pm$ 0.21 & 2.44 $\pm$ 0.62 & 7.62 $\pm$ 2.36\\
APS & 1.73 $\pm$ 0.05 & 1.95 $\pm$ 0.08 & 2.49 $\pm$ 0.16 & 3.79 $\pm$ 0.39 & 2.02 $\pm$ 0.32 & 2.52 $\pm$ 0.43 & 4.64 $\pm$ 0.94 & 9.38 $\pm$ 2.05
  \\
\bottomrule
\end{tabular}
}
\end{table*}

\begin{table*}[h]
\caption{Accuracy and \loc{5} of \hqpm{} when increasing the weighting of the Feature Diversity Loss \gls{customLoss} $\beta$ beyond the default value $D$, used for Q-SENN, the SLDD-Model and QPM. Without reducing accuracy, \loc{5} can be further increased for \hqpm{}.}
\label{stab:diversity}
\centering
\resizebox{\linewidth}{!}{
\begin{tabular}{lcccccc}
\toprule
\multirow{2}{*}{Metric} & \multicolumn{3}{c}{CUB} & \multicolumn{3}{c}{IMG}  \\ &
  $\beta=D$ & $\beta=\sqrt{2}D$ & $\beta=2D$ &$\beta=D$ & $\beta=4D$ & $\beta=12D$  \\
\midrule
Accuracy&	85.3 &	85.3&85.1&	75.3 & 75.3&75.1 \\
\loc{5} &88.1 & 92.7 & 96.5 & 42.9&70.3&77.0
  \\
\bottomrule
\end{tabular}
}
\end{table*}

\section{Fine-tuning}
\label{ssec:Tune}
\paragraph{Fine-tuning}
After solving the quadratic problem, the feature selection and binary assignment are set as fixed parameters of our model.
Following \qsenntable{}, \slddtable{} and \gls{NewlayerName}, we continue training the model, so that the features adapt to the concept(s) that is or are shared by its assigned classes and also follow their normalization.
Specifically, the mean $\muf\in\mathbb{R}^{\gls{nReducedFeatures}}$ 
and standard deviation $\sigmaf\in\mathbb{R}^{\gls{nReducedFeatures}}$  
are computed on the  the training set:
\begin{align}
{f}_{j,\findex} &=  \frac{1}{\gls{featuresMapwidth} \gls{featuresMapheigth} }  
\sum_{l=1}^{\gls{featuresMapwidth}} \sum_{k=1}^{\gls{featuresMapheigth}} m_{\findex,l,k}^j \\
\muf_\findex &= \frac{\sum_{j=1}^{\gls{nTrainImages}}{f}_{j,\findex}}{\gls{nTrainImages}}\\
\sigmaf_\findex &= \sqrt{\frac{\sum_{j=1}^{\gls{nTrainImages}}({f}_{j,\findex} - \muf_\findex)^2}{\gls{nTrainImages}} },
\end{align}
where $m_{\findex,l,k}^j$ refers to the value of the feature maps at the spatial position $l, k$ for feature $\findex$ and sample $j$.
This normalization for every feature $\findex$ is maintained for the fine-tuning:
\begin{equation}
   \hat{f}_\findex = \frac{1}{\gls{featuresMapwidth} \gls{featuresMapheigth} }  \frac{\sum_{l=1}^{\gls{featuresMapwidth}} \sum_{k=1}^{\gls{featuresMapheigth}} m_{\findex,l,k} - \muf_\findex}{\sigmaf_\findex} \label{eq:norming}
\end{equation}
While this was initially done due to the reliance  of \qsenntable{} and \slddtable{} on \gls{glmsaga} for sparsification, it has also proven effective for QPM~\cite{qpmPaper}. 
While it is beneficial for the accuracy, as we show in \cref{stab:ablNormRel}, this reduces the global interpretability, as their absence, a sum of $0$ across the entire feature map $\textbf{M}$, has an effect due to varying means \muf.
Hence, we additionally apply ReLU to the features after scaling, so that \hqpm{} only reasons positively, and negligible activations are suppressed:
\begin{equation}
     \textbf{f}^* = \mathrm{ReLU}( \hat{\textbf{f}}) \label{seq:Relu}
\end{equation}
Note that this already makes \hqpm{} a set-valued predictor, as \eg when only the $\gls{nperClass}-1$ features shared by 2 classes in \CpairsSet{} activate, both would be predicted.
However, on the test datasets we used this happens very seldom and, as explained in \cref{ssec:impdeta}, we use $\mathrm{argmax}$ to break ties for evaluating point-predictive performance.

\subsection{Change during Fine-tuning}
\label{ssec:QPSolution}
\newt{
This section presents results on how much features change during the fine-tuning of CHiQPM. \Cref{tab:rho_pearson} quantifies the Pearson Correlation between the same features before and after the fine-tuning in relation to $\rho$. Evidently, a higher $\rho$ causes higher change in the features. As a consequence CHiQPM is less reliant on the perfect choice of initial features, as they undergo more changes regardless. For reference, QPM, has an average correlation of 0.855. 
}
\begin{table}[h]
\caption{Impact of $\rho$ on Average Pearson Correlation between features before and after fine-tuning.}
\label{tab:rho_pearson}
\centering
\begin{tabular}{cc}
\toprule
$\rho$ & Average Pearson Correlation \\
\midrule
0.0 & 0.825 \\
0.1 & 0.823 \\
0.3 & 0.805 \\
0.5 & 0.776 \\
0.7 & 0.743 \\
\bottomrule
\end{tabular}
\end{table}

\section{Relaxation of Hierarchical Constraint}
\label{ssec:Relax}
This section presents the iterative relaxation of the hierarchical constraint introduced in \cref{sec:constr}.
We relax the constraint of \cref{eq:o4sim} after finding the initial global solution for \wgurobi{} and \fvecgurobi{}, $\wgurobi_\mathrm{init}$ and $\fvecgurobi_\mathrm{init}$.
Specifically, we set $\fvecgurobi{}=\fvecgurobi_\mathrm{init}$ to  simplify the problem 
and allow  the addition of $\mathbf{{\achieved}}\in\{0,1\}^{\lvert\iterPairs\rvert}$ as an additional variable to optimize to the binary problem without introducing cubic constraints.
It directly optimizes which pair of classes shall share $\gls{nperClass} -1$ features with the lowest reduction on the objective for
the desired density \treedense.
Here \iterPairs{} describes the set of pairs of classes that ever were represented sharing $\gls{nperClass} -1$ features during the iterative optimization process.
Because the hierarchical constraint causes shared concepts to be represented by the shared features exclusively, other pairs in \CpairsSet{} not included in \pairsTarget{} emerge as other classes share this concept too.
Extending how the QPM~(\cref{sec:qpm})
ensures no duplicates, \iterPairs{} is continuously updated, and the optimum is found when all variables, including \iterPairs{}, have not changed for an iteration:
\begin{align}
(\boldsymbol{w}_\cindex \circ \boldsymbol{w}_{\cindex'})^T \fvecgurobi{}_\mathrm{init}\cdot\achieved_i &= (\gls{nperClass} -1)\cdot\achieved_i \quad \forall  (\cindex,\cindex') \in \iterPairs
   \label{eq:o4simCon}\\
   \sum_{i=0}^{\lvert\iterPairs\rvert} \achieved_i  &\geq \lvert\pairsTarget\rvert,
\end{align}

\section{Impact of $\lambda_\mathrm{feat}$}
\label{ssec:impactlambda}
\begin{figure*}[htb]
\centering

\begin{subfigure}[t]{.5\textwidth}
    \centering
    \includegraphics[width=\textwidth]{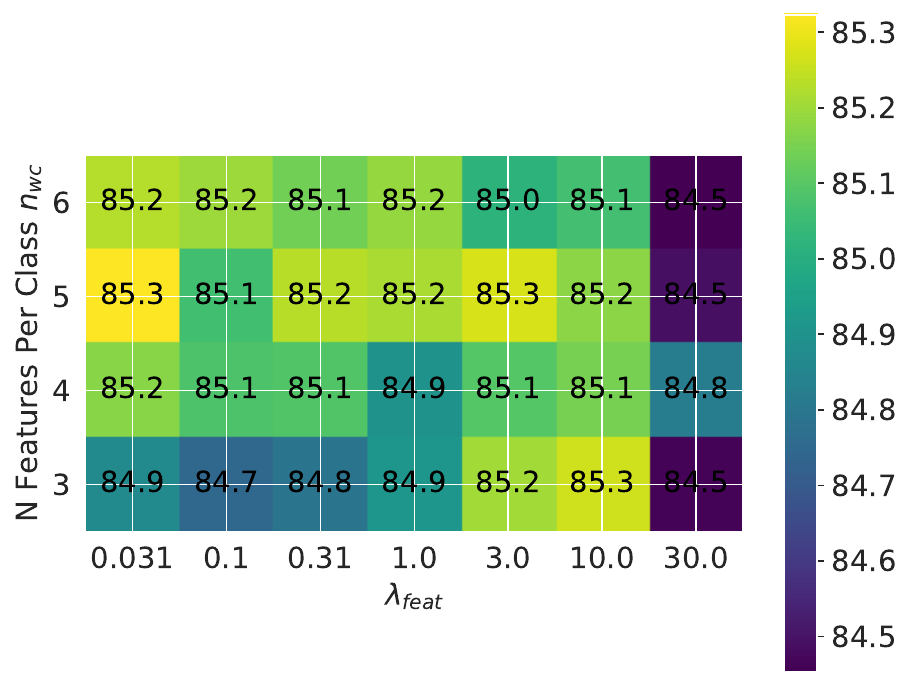}
    \caption{Accuracy} 
    \label{fig:cub_acc_qsenn}
\end{subfigure}\\
\begin{subfigure}[t]{.5\textwidth}
    \centering
    \includegraphics[width=\textwidth]{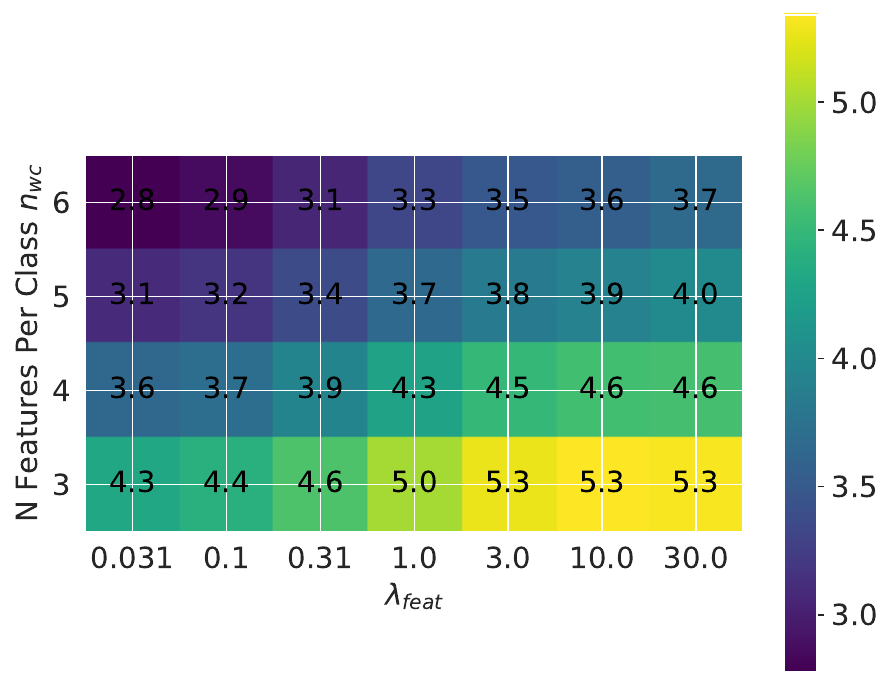}
    \caption{Feature Alignment} 
    \label{fig:cub_acc_qpm}
\end{subfigure}\\
\begin{subfigure}[t]{.5\textwidth}
    \centering
    \includegraphics[width=\textwidth]{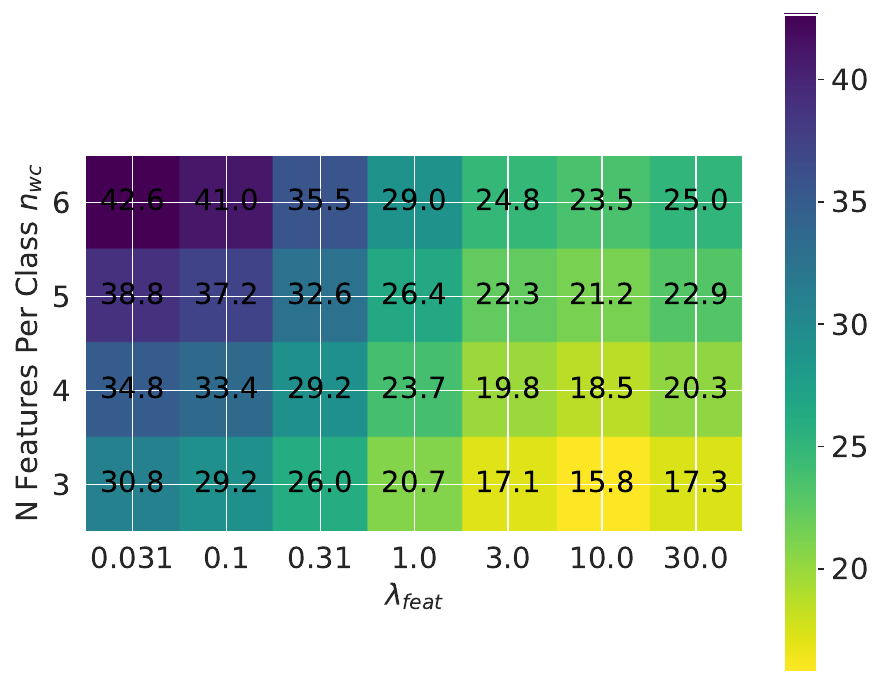}
    \caption{Sparsity} 
    \label{fig:cub_acc_eqpm}
\end{subfigure}
\caption{Accuracy, Feature Alignment and Sparsity in relation to $\lambda_\mathrm{feat}$and the number of features per class \gls{nperClass} on \cubheader{}} 
\label{sfig:grid}
\end{figure*}

\newt{
This section is concerned with the weighting factor $\lambda_\mathrm{feat}$ of the Feature Grounding Loss \floss. As described, it is set to $\lambda_\mathrm{feat}=3$, as the ablation in \cref{sfig:grid} demonstrates that a further increase harms accuracy, without significantly further improving sparsity and feature alignment. 
This ablation also highlights that features become more aligned with the attributes in \cubheader{}, as the sparsity increases. This is likely due to the shared concept between the assigned classes being less abstract.
For reference, QPM reaches a Feature Alignment of 1.9, half of \hqpm{}, and thus learns significantly less grounded features than \hqpm{}.
}

\section{Impact of ReLU}
\label{ssec:ReluAbs}



The necessity of the  normalization and the improvements through our method are summarized in \cref{stab:ablNormRel}.
First, it is clear that normalization is required to achieve state-of-the-art performance and ReLU further improves accuracy.
However, having features with a shared fixed minimum, either without normalization or due to ReLU evidently increases Feature Alignment.
Additionally, the ReLU significantly boosts \contrastiveness{}, as all the non-activating features are exactly $0$, leading to a lower distribution $\mathcal{N}_1$ (\cref{eq:contra}) with very low standard deviation and therefore an almost perfect distinction between active and inactive features.
However, the \generality{} is slightly reduced with a shared minimum, as there are no minimal inactive activations that otherwise count towards \generality.
Regardless, a \generality{} of around $95\%$ is indistinguishable from a perfect score, as a perfect feature, that equally activates on all images of its assigned classes  and never on other images would have $\generality{}=1-\frac{20}{ 200} = 90\%$.
This calculation is based on the default configuration ($\gls{nperClass}=5$ and $\gls{nReducedFeatures}=50$) for \cubheader{}, where every feature is on average assigned to $20$ out of the $200$ classes.
Hence, the feature activation would be focussed on these $10\%$ of the class samples.
A further increased value can be attributed to robust features that are only sometimes part of a classes appearance.
One example are birds that change their color during their lifetime, causing these color features to activate even without being assigned to these classes.

\section{Impact of Dynamic Hierarchy}
\label{ssec:DynHiera}

\begin{table}[h]
\caption{Comparison of dynamic and static feature ordering  across different levels in the hierarchy for CHiQPM with \resnet{} on \cubheader{}.}
\label{tab:feature_order_comparison}
\centering
\resizebox{\linewidth}{!}{
\begin{tabular}{llccccc}
\toprule
Method & Metric & Level 1 & Level 2 & Level 3 & Level 4 & Level 5 \\
\midrule
Dynamic Hierarchy (CHiQPM) & Set Accuracy & 96.5 & 93.8 & 91.6 & 89.5 & 85.3 \\
& Set Size & 20.17 & 4.43 & 2.12 & 1.45 & 1.00 \\
\midrule
Static Hierarchy (Avg. Order) & Set Accuracy & 89.7 & 87.4 & 86.2 & 85.6 & 85.2 \\
& Set Size & 18.52 & 2.85 & 1.30 & 1.09 & 1.00 \\
\bottomrule
\end{tabular}
}
\end{table}
\newt{
This section evaluates the importance of using our sample specific hierarchy instead of a static one. Our used hierarchy based on order of the features of the predicted class has significant advantages compared to other choices. Its sample specific nature lends itself to a local explanation, as one can consider the order of the features as how dominant they appear in the image. Additionally, that enables traversing up the hierarchy in a meaningful way, since the least certain features get omitted first, causing accurate and efficient sets at each step. We believe that the main alternative is a class specific fixed order of features, as that would enable a global hierarchical explanation for each class. The comparison between a fixed order of features for each class, computed based on the average order on training data and our prediction up the hierarchy is shown in ~\cref{tab:feature_order_comparison}. Evidently, predicting with such a fixed order causes less accurate inefficient sets compared to our sample specific dynamic hierarchy. For example, our dynamic hierarchy reaches 91.6\% accuracy with an average set size of 2.1, while the fixed hierarchy never reaches that accuracy. We believe that this is likely due to the unique feature, e.g. the red eye of the Bronzed Cowbird, being on average quite important to that class and thus causing prediction sets that are too specific to the top-1 prediction.  
}

\section{Localization Quality of Features}
\label{ssec:GradCam}
\newt{
This section evaluates the localization quality of the feature maps.
We evaluate a form of the Pointing Game, similar to the initial form in \cite{zhang2018top}, using segmentation masks provided for \cubheader{}~\cite{farrell_2022}. Specifically, we calculate the fraction of the activation of the GradCAM~\cite{selvaraju2020grad} saliency map that is focused on the segmented bird $S\in\{0,1\}^{w \times h}$:
\begin{equation}
    \phi = \sum \left( \frac{\text{GradCAM}}{\sum\text{GradCAM}} \odot S \right)
\end{equation}
As the bird can be considered the region of the image responsible for the classification, a higher overlap with the segmentation indicates that the saliency maps localize more faithfully. The results are shown in ~\cref{tab:model_comparison}. Importantly, one would not necessarily expect an overlap of 100\%, as the edge region is relevant to describing the shape, causing activations both on and off the segmentation. Across the entire test dataset, CHiQPM’s GradCAM focusses to 83.4\% on the bird, whereas the Dense baseline only does so with 75.8\%. Hence, CHiQPM has activation maps with improved faithfulness on CUB-2011, validating their use for saliency maps of individual interpretable features. 
}
\begin{table}[h]
\caption{Accuracy and faithfulness comparison on \cubheader{} with \resnet{}. The final column measures the average GradCAM overlap with ground truth segmentations.}
\label{tab:model_comparison}
\centering
\begin{tabular}{lcc}
\toprule
Model & Accuracy & Avg. GradCAM Overlap \\
\midrule
Dense & 86.6\% & 75.8\% \\
CHiQPM & 85.3\% & 83.4\% \\
\bottomrule
\end{tabular}
\end{table}

\section{Limitations}
\label{ssec:Limits}
This section discusses the limitations of the proposed \hqpm{}. 
\hqpm{} is a model that learns few general features which are used in a very interpretable way with a binary sparse assignment.
The learnt features already enable improved interpretability, as they tend to localize consistently on the same concepts enabling an intuitive understanding of the features and predictable local behvaiour based on global explanations. For example, given the global explanation in \cref{fig:TeaserGlob}, a user would be able to predict that the image in \cref{fig:TeaserFull} would show little activation as the eye is different. Notably, this is different to most prototypical models, which learn class-specific features~\cite{qpmPaper} that detect \textit{red eye of Bronzed Cowbird} instead of \textit{red eye} and thus lose human interpretability~\cite{kim2021hive,hoffmann2021looks}.
However, for perfect human-understandable global interpretability all of the individual features have to be understandable for humans. 
Dependent on the dataset, there are 2 to 3 remaining obstacles:
\begin{enumerate}
    \item As touched upon in \cref{sec:rlworkInt}, features can learn to detect multiple concepts, a phenomenon knows as polysemanticity. While the visualizations of features in \cref{sfig:BigTern,sfig:SmallTern,sfig:FeaturesBlack,sfig:FeaturesCover} indicate a consistent localization of the same feature across many classes on the same concept, proper metrics are missing to  even measure that. However, we believe that a model like \hqpm{} is very well suited to investitage this phenomen further, as superposition~\cite{elhage2022toy} can likely be ruled out in its final features. The issue of polysemanticity is exacerbated on \imgnetheader{}, where every feature is assigned to $\approx$ 100 classes. This connects to another limitation, where the number of selected features ~\gls{nReducedFeatures} can not be set arbitrarily high, as the time it takes to solve the QP grows exponentially with it, as shown in \cite{qpmPaper}.
    \item  While the activation maps of \hqpm{} on \cubheader{} and \stanfordheader{} seem to localize very accurately, \eg{} highlighting the red eye and not activating if it is not visible, the activation maps on \imgnetheader{}  seem to not always faithfully highlight the image region they respond to, \eg{} Feature 23 in \cref{sfig:mount1,sfig:mount2,sfig:I1Glob} consistently distributes a large portion of its activation on the same image patch. More elaborate saliency map methods or built-in methods like B-cos Networks~\cite{bohle2022b} might be required for such activation maps. Finally, with the seemingly lacking faithfulness of the activation maps for \imgnetheader{}, the use of \loc{5} for this dataset can also be questioned.
    \item \hqpm{} learns general features that are well suited to classify the dataset given the training data. 
However, we do not restrict the features to be based on concepts that humans have noticed or named before. 
Therefore, there may exist a conceptual gap between the concepts learnt by \hqpm{} and the ones known to humans. 
This Bi-directional Communication Problem~\cite{ayonrinde2025position} can however be reduced by examining the learnt general features of \hqpm{} with its faithful global interpretability, as one can \eg{} say that there is a distinctive pattern on the necks of  the black birds, that can be used to differentiate them. 
Teaching humans the general shared features as neologisms~\cite{hewitt2025we} seems promising to bridge this gap.  Again, the learnt features of \hqpm{} are uniquely suited as they are individual neurons that already detect concepts.
Finally, further work should aim at distinguishing polysemantic features  from  features that capture one concept unknown to humans. 
\end{enumerate}

The proposed method is evaluated quite thoroughly on multiple datasets with various architectures. Thus, \hqpm{} can clearly be applied to other general vision datasets.  However, these datasets need to have many classes that have shared concepts, as the method relies on learning shared concepts between the classes.

For assumptions, it is to note that the guaranteed coverage of Conformal Prediction only holds under the exchangeability assumption of test and calibration data. \newt{We designed our experiments to ensure this property. Specifically, splitting calibration off from the test data ensures i.i.d., which is a stronger property than exchangeability. Additionally, we observe that all CP methods closely match the target coverage for which we calibrate on test data.}

Finally, the built-in interpretable Conformal Prediction method of this work guarantees unconditional coverage under the given assumptions, \ie{} achieving the desired accuracy across the entire test dataset, instead of for all test samples. Achieving unconditional instead of conditional coverage can relate to fairness, as some groups might be undercovered. 
Limiting the number of levels to ascend during the conformal prediction can theoretically further negatively affect the conditional coverage, as some classes are further away from other classes. 
However, conditional coverage is challenging to even measure, as one needs a fair assessment of difficulty.
For the metrics we checked, using \adalvl{} did not negatively affect conditional coverage.
However, as the set construction is interpretable for \hqpm{}, the user can understand high uncertainty when all classes below the maximum number \adalvl{} are predicted, which may reduce the need for built-in conditional coverage.

\section{Future Work}
\label{ssec:FutureWork}
\newt{
This section discuses the future directions of this work. 
One direction is a formal analysis of the used QP. Understanding the properties in more detail might enable faster solving or scaling the model to even larger datasets that might require even more features.\\
Another very valuable direction is validating the effectiveness of the introduced explanations via human studies, which have been out of scope for this work. \\
Finally, we hypothesize that CHiQPM is well suited to handle class-imbalanced data, as the constraints on the QP ensure that every class can be predicted if their general features are present. 
}

\section{Broader Impact}
\label{ssec:BroaderImpact}
This paper presents work whose goal is to advance the field of
Machine Learning towards more interpretability. 
Globally interpretable models enable a deeper understanding of the decision-making before and during deployment, and can therefore lead to increased safety, robustness, trustworthiness and potentially even scientific discovery.
However, there are also potential negative consequences. 
Most notable, a bad actor can use a model with faithful global interpretability to deliberately and systematically ensure decision-making based on spurious or discriminating factors.
Nevertheless, only such globally interpretable models offer the transparency to robustly ensure decision-making based on the correct concepts.
\section{Compute Ressources and Runtime}
\label{sec:computeRes}
This section discusses the compute ressources used for the experiments in this paper. 
As GPU ressource, this work made use of an internal cluster composed of several NVIDIA RTX 2080 Ti. Every experiment fit on one GPU. 
As CPU ressource for solving the QP, this paper used an internal CPU cluster composed of primarily AMD EPYC 72F3 and up to 250GB of ram. 
 This reference hardware with sufficient memory is assumed for all estimates. 

\begin{table}[h]
\caption{Rough time in minutes needed to obtain a \hqpm{}, $\gls{nperClass}=5$ and $\gls{nReducedFeatures}=50$, with \resnet{} on the three datasets  used in this work.}
\label{stab:ImgNetTime}
\centering
\begin{tabular}{lcccc}
\toprule
Dataset & Dense Training (GPU) & QP (CPU) & Fine-Tuning (GPU) & Total Time (Hours)\\
\midrule
\cubheader{} & 176 & 200 & 82 & 458 (7.6)  \\

\stanfordheader{}  & 234 & 200 &110 & 544 (9.1) \\
\imgnetheader{} &  0 & 660 &  6300 & 6960 (116)\\

\bottomrule
\end{tabular}
\end{table}
\begin{table}[h]
\caption{Rough time for the 5 seeds in minutes (hours) spent on training each of the architectures for the main experiments on \cubheader{} and \stanfordheader{} (taking $\frac{4}{3}$ the time of \cubheader{} on GPU).}
\label{stab:timeMain}
\centering
\begin{tabular}{lcc}
\toprule
Architecture & GPU &CPU \\
\midrule
\resnet{} & 258 $\cdot$5 $\cdot$$\frac{7}{3}$ & 200 $\cdot$5 $\cdot$2\\
Resnet34  & 143 $\cdot$5$\cdot$$\frac{7}{3}$ & 58 $\cdot$5$\cdot$2 \\
\gls{incv}  &  76 $\cdot$5 $\cdot$$\frac{7}{3}$ & 200 $\cdot$5 $\cdot$2\\ 
Swin-Transformer-Small &478$\cdot$ 5$\cdot$$\frac{7}{3}$  & 200 $\cdot$5 $\cdot$2\\
$\sum$ & 11141 (185) & 6580 (110) \\

\bottomrule
\end{tabular}
\end{table}

\begin{table}[h]
\caption{Rough time in hours spent on training the additional models for the Ablations}
\label{stab:timeAbl}
\centering
\begin{tabular}{lccc}
\toprule
Ablation & Number of Models & GPU &CPU \\
\midrule

\cref{tab:flossAbl} & 15 & 21 & 0 \\
\cref{stab:ablNormRel}& 10 & 14 & 0 \\
\cref{fig:dimVar} & 70 & 75 & 230 \\
\cref{stab:diversity} & 12 (6 on IMG) & 352 &  0 \\
\cref{fig:QPMintroComp} (QPM) & 5 & 4 & 17  \\
\cref{fig:SimPlot} & 15 & 21 & 50 \\
$\sum$ & 127 & 487 & 297 \\
\bottomrule
\end{tabular}
\end{table}

\begin{table}[h]
\caption{Rough total time in hours spent on training with reference hardware NVIDIA RTX 2080 Ti as GPU and AMD EPYC 72F3 as CPU}
\label{stab:timeTotal}
\centering
\begin{tabular}{lcc}
\toprule
Experiment & GPU &CPU \\
\midrule
Main Experiments (\cref{stab:timeMain}) & 185 & 110\\
Ablations~(\cref{stab:timeAbl}) & 487 & 297 \\
ImageNet Main Experiments~(\cref{stab:ImgNetTime}) & 350 & 33\\
Preliminary / Failed experiments & 300 & 600 \\
Q-SENN~\cite{norrenbrock2024q} training as competitor & 60 & 0 \\
$\sum$ & 1382 & 1040 \\
\bottomrule
\end{tabular}
\end{table}

All main experiments were done on four architectures. 
The total time it takes to obtain a \hqpm{} dependes on the time spent training on the GPU, scaling with model and dataset size, and time spent solving the QP on the CPU, which scales with the dimensions of the variables and constraints. With $\gls{nperClass}=5$ and $\gls{nReducedFeatures}=50$, the CPU time depends primarily on the number of classes in the dataset \gls{nClasses} and the number of features of the black-box dense model \gls{nFeatures}.
As \stanfordheader{} has a very similar number of classes to \cubheader{} (196 to 200), we calculate with the same number of classes and $\frac{4}{3}$ times the number of training samples (8144 to 5994).
The estimated time spent per dataset is shown in \cref{stab:ImgNetTime}. How that varies between architectures is shown in \cref{stab:timeMain}.
While there is some variance in cpu time due to the ablations, we assume even cpu runtime for our estimates in \cref{stab:timeAbl}, which effectively results in an upper bound on the cpu time spent.

Generating the visualizations or explanations took negligible compute ressources as everything required is computed in one forward pass. Similarly, evaluating the models is also very fast in comparison.

Finally, most of the experiments with \resnet{} were ran on $3-5$ random seeds  with \cubheader{} before starting training for the fixed seeds and across multiple architectures. 
Similarly, further experiments of roughly the same number were run to ultimately converge to the presented method. 
Also, starting from 61GB for solving the QP, very few seeds needed more than that, which we started after the inital ones crashed. For the three solved QPs on \imgnetheader{}, we always allocated the available memory of 250G.
Additionally, we had to retrain Q-SENN to include it in our tables. Each of these runs takes roughly half an hour more on the GPU than \hqpm{}. Note that, Q-SENN did not always converge with Swin-Transformer-Small, which is why no result is reported there as one likely needs to tweak hyperparameters to ensure convergence. 
Similarly, \cref{fig:QPMintroComp} required training QPM.
We want to emphasize that we reused the same dense model or even QP solution where possible, \eg{} for \hqpm{}, QPM and Q-SENN or many of the ablations.
This puts the total amount of GPU hours to roughly 1382 and CPU hours to 1040, shown in \cref{stab:timeTotal}.

For storage, we temporarily save every model and their low-dimensional feature vectors to speed up metric calculations. While our internal clusters offers significantly more storage, the experiments of this paper needed less than 100 GB.

\section{Further Qualitative Examples}
\begin{figure}[ht]
\vskip 0.2in
\begin{center}
\centerline{\includegraphics[width=\columnwidth]{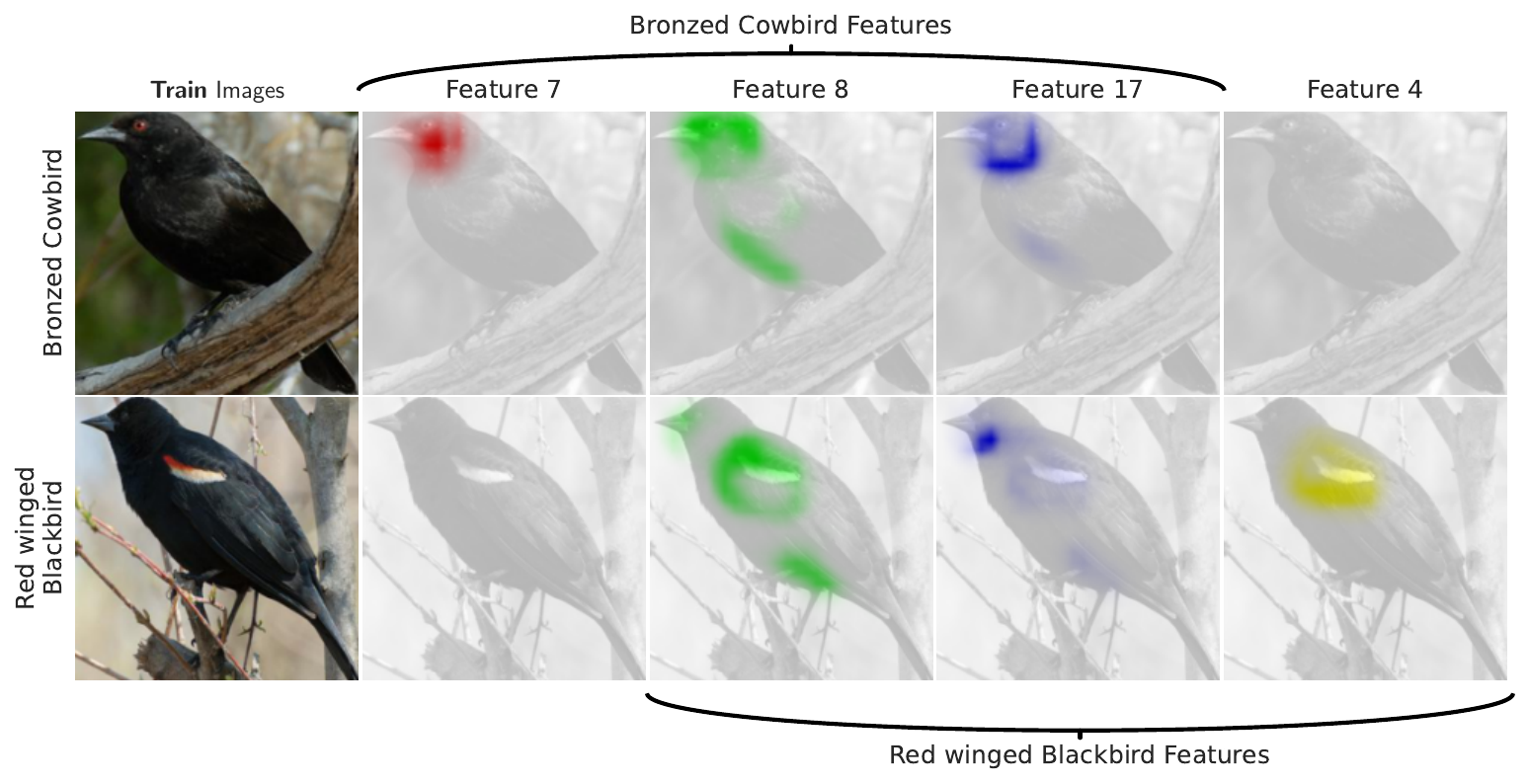}}
\caption{Global explanation comparing Bronzed Cowbird and Red Winged Blackbird. The \hqpm{} trained on \cubheader{}  with 3 features per class and 30 in total, explained in \cref{fig:TeaserFull,fig:TeaserGlob}, determined the differentiating factor, their red eye or wing, without any annotations and can communicate its behavior faithfully.}
\label{fig:globRed}
\end{center}
\vskip -0.2in
\end{figure}

\begin{figure}[ht]
\vskip 0.2in
\begin{center}
\centerline{\includegraphics[width=\columnwidth]{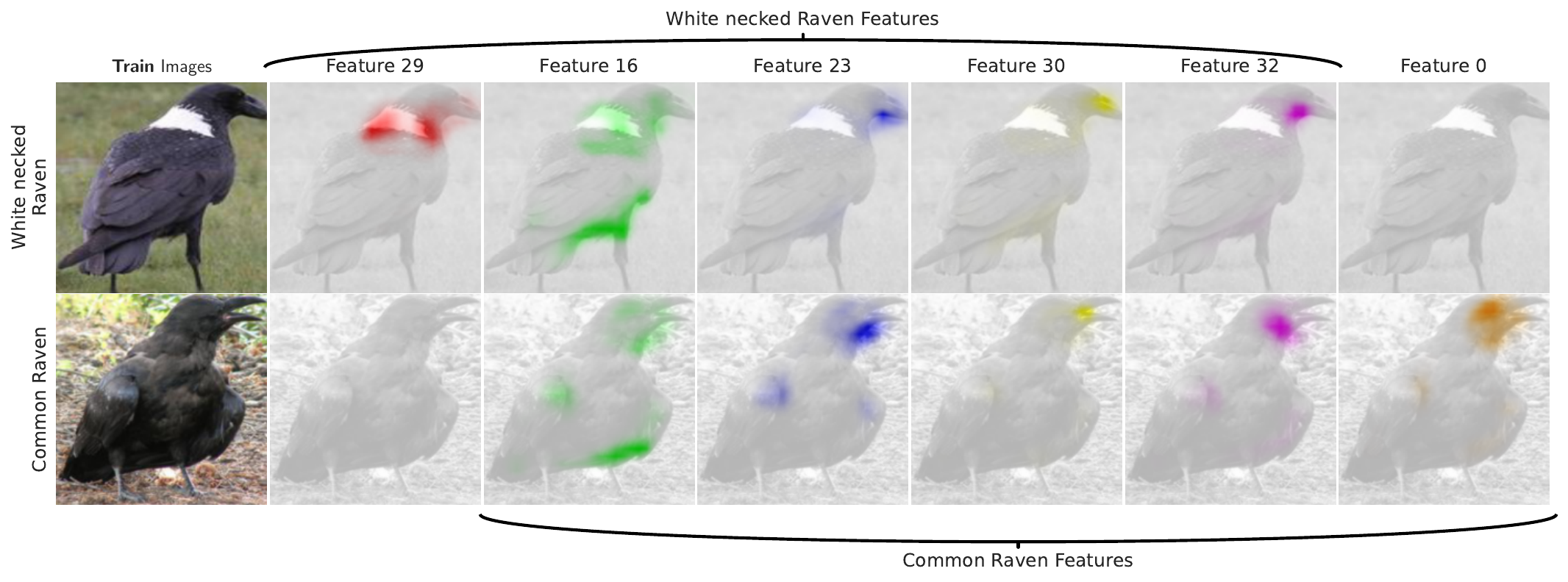}}
\caption{Global explanation comparing White necked and Common Raven. \hqpm{} trained on \cubheader{}  determined the differentiating factor, the white neck, without any annotations and can communicate its behavior faithfully. Local explanations of this model are shown in \cref{sfig:neckloc2,sfig:necklock2}.}
\label{fig:globNeck}
\end{center}
\vskip -0.2in
\end{figure}

\begin{figure}[ht]
\vskip 0.2in
\begin{center}
\centerline{\includegraphics[width=\columnwidth]{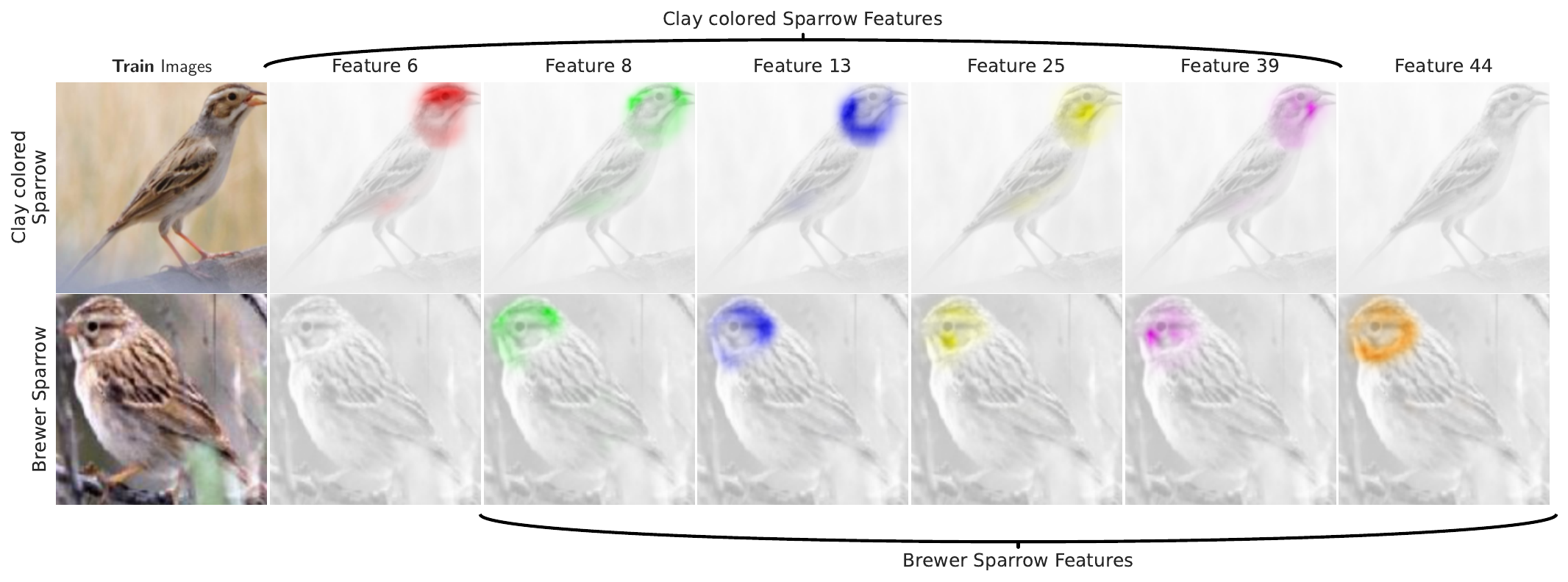}}
\caption{Global explanation comparing Clay colored and Brewer Sparrow. \hqpm{} trained on \cubheader{}  determined the differentiating factor, the distinct head patterns, \eg{} the white crown stripe for Clay colored Sparrow, without any annotations and can communicate its behavior faithfully.
Local explanations of this model are shown in \cref{sfig:spar2,sfig:spar1}.
}
\label{sfig:sparrows}
\end{center}
\vskip -0.2in
\end{figure}

\begin{figure}[ht]
\begin{center}
\centerline{\includegraphics[width=\columnwidth]{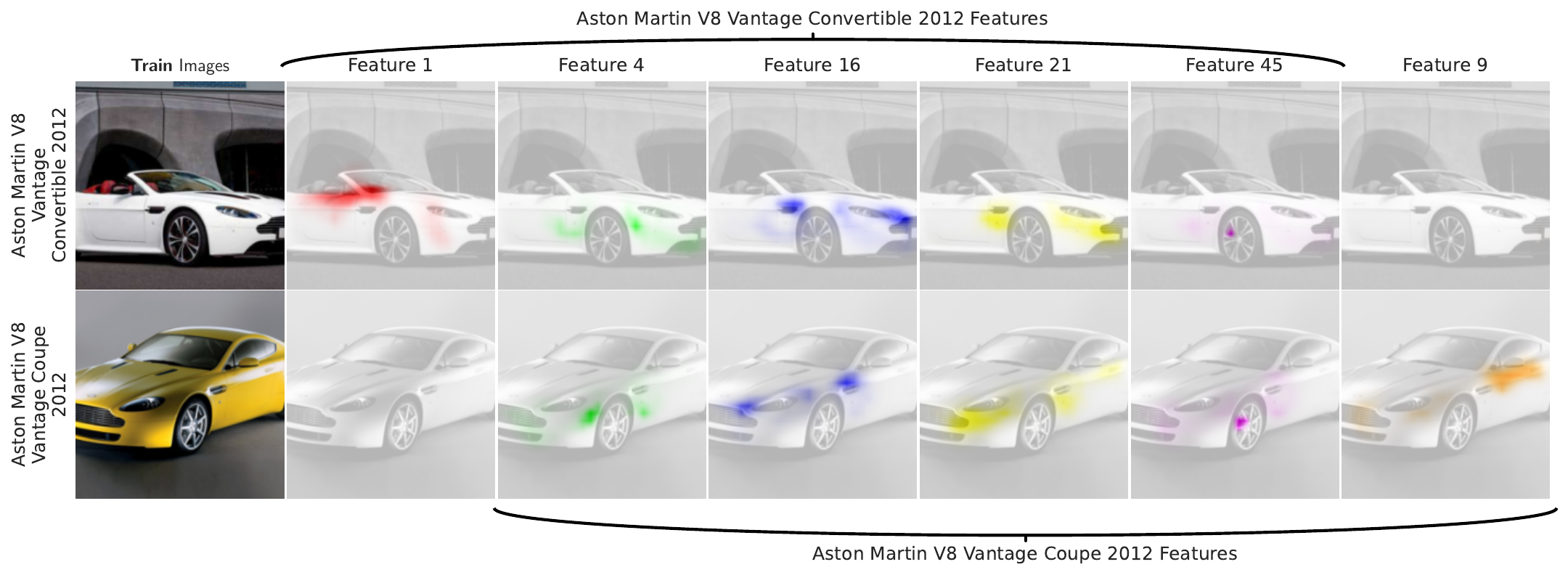}}
\caption{Contrastive global Explanation, comparing the class representations of two cars that only differ in Coupe or Convertible for \hqpm{} trained on \stanfordheader{}  that represents every class with 5 of 50 features. They are differentiated based on features activating on where the windows would be.}
\label{sfig:C1Glob}
\end{center}
\vskip -0.2in
\end{figure}
\begin{figure}[ht]
\begin{center}
\centerline{\includegraphics[width=\columnwidth]{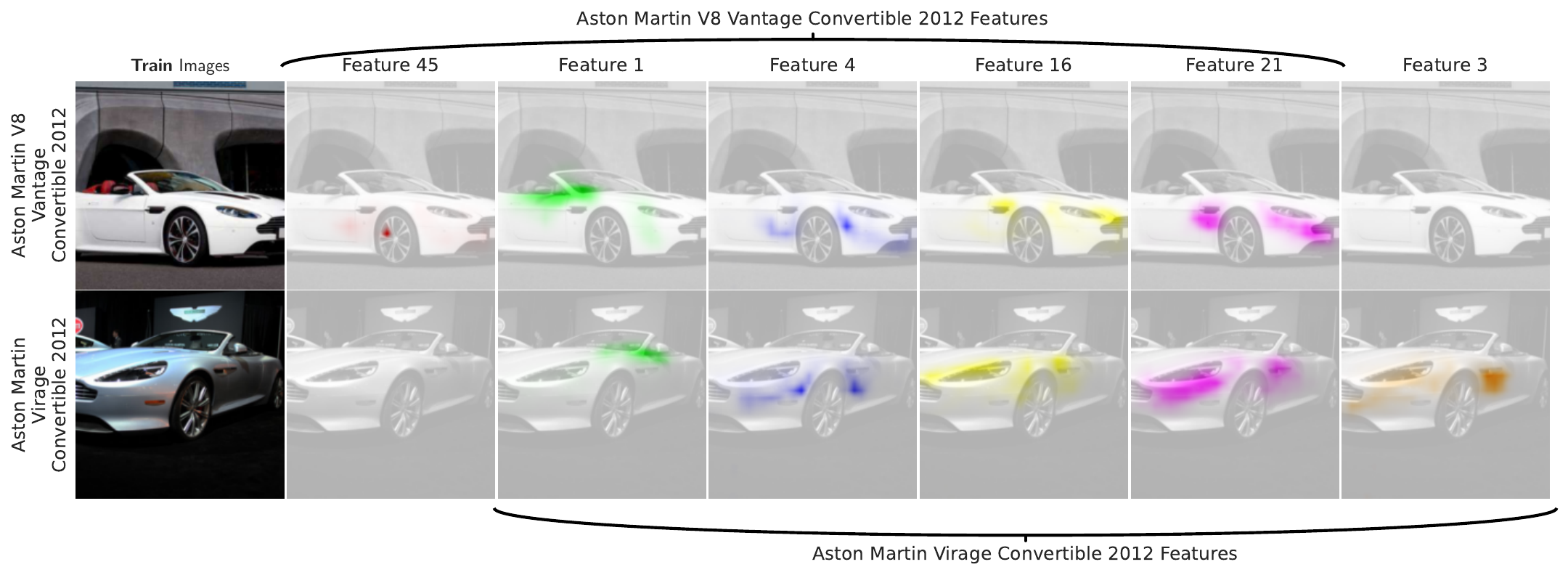}}
\caption{Contrastive global Explanation, comparing the class representations of two Convertible Aston Martins for \hqpm{} trained on \stanfordheader{} that represents every class with 5 of 50 features. They are differentiated based on human perceivable deviating features like the fender vent.}
\label{sfig:C2Glob}
\end{center}
\vskip -0.2in
\end{figure}

\begin{figure}[ht]
\begin{center}
\centerline{\includegraphics[width=\columnwidth]{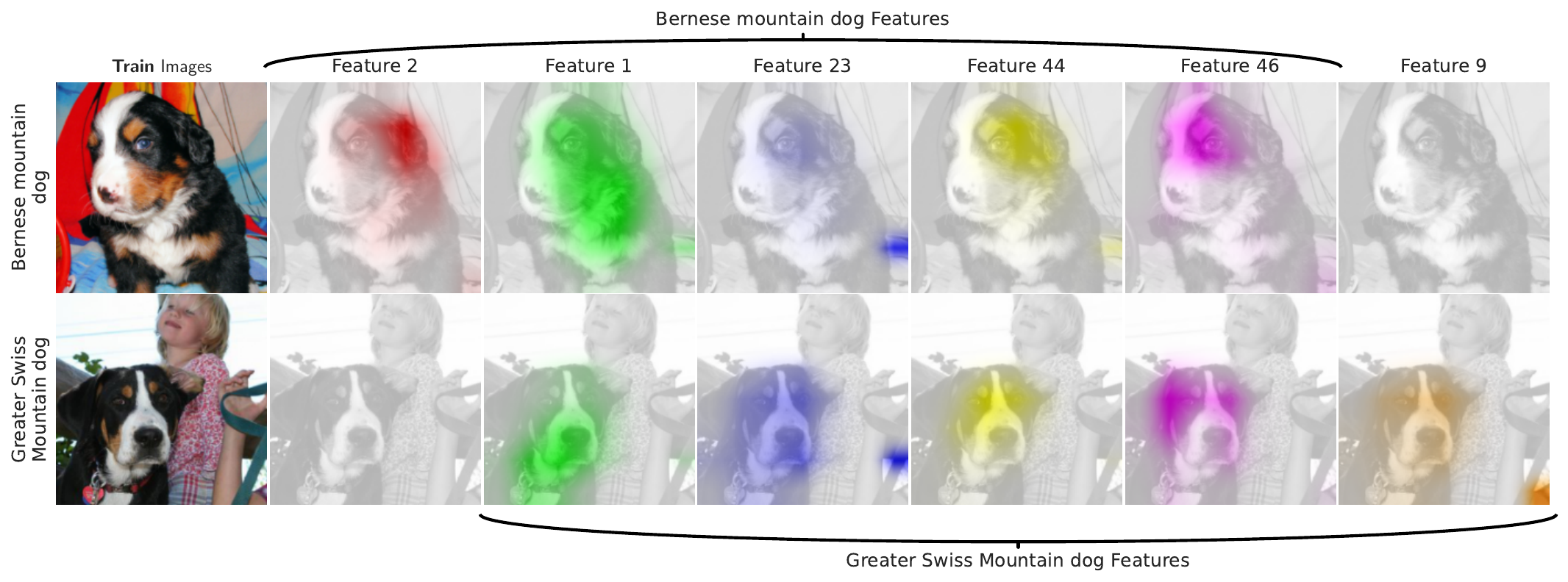}}
\caption{Contrastive global Explanation, comparing the class representations of two mountain dogs for \hqpm{} trained on \imgnetheader{} that represents every class with 5 of 50 features. They are differentiated based on human perceivable deviating features like the different ear fur. Local explanations for this model are shown in \cref{sfig:mount2,sfig:mount1}.}
\label{sfig:I1Glob}
\end{center}
\vskip -0.2in
\end{figure}
\begin{figure}[ht]
\begin{center}
\centerline{\includegraphics[width=\columnwidth]{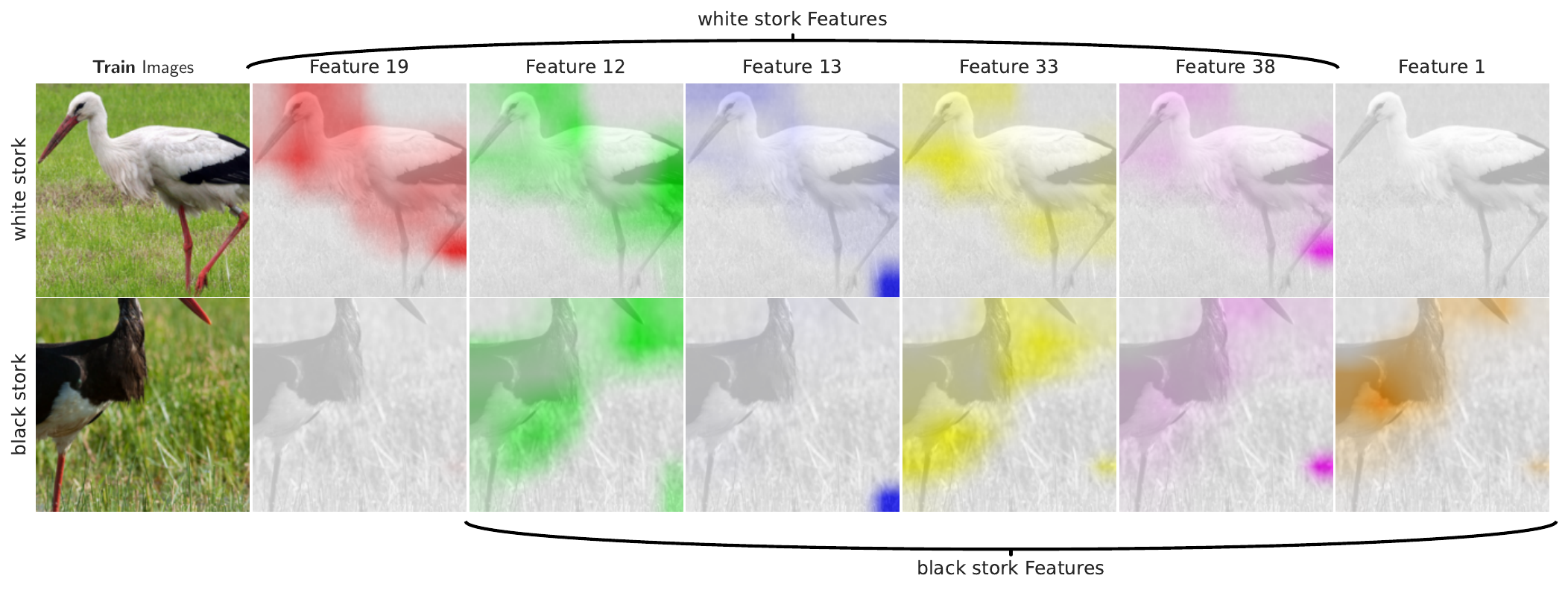}}
\caption{Contrastive global Explanation, comparing the class representations of white and black stork for \hqpm{} trained on \imgnetheader{} that represents every class with 5 of 50 features. They are differentiated based on one broadly activating feature.}
\label{sfig:I2Glob}
\end{center}
\vskip -0.2in
\end{figure}
\begin{figure}[h!t]
\begin{center}
\centerline{\includegraphics[width=\columnwidth]{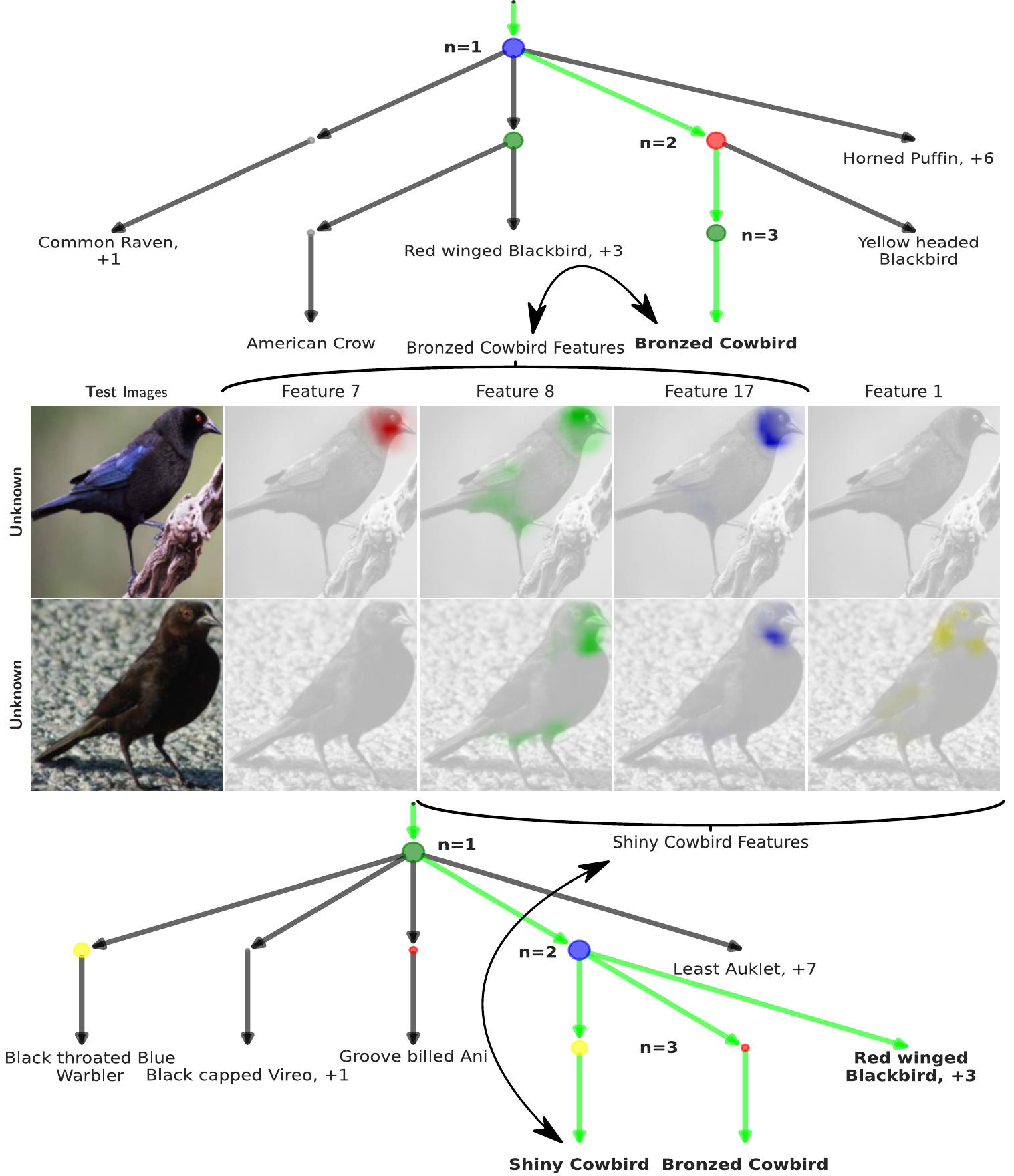}}
\caption{Exemplary local explanations provided by our \hqpm{}, with the global explanation in \cref{fig:TeaserGlob} for two test images of Bronzed Cowbird. The first row is an easy example where all $3$ features are found, including the red eye. Therefore, our calibrated model only predicts Bronzed Cowbird. 
The red eye is not visible in the second image, leading to the reasoning of our \hqpm{} along its dynamic class hierarchy identifying it as one of the black bird species and predicting all of them, including the Bronzed Cowbird.
}
\label{sfig:TeaserFull}
\end{center}
\vskip -0.2in
\end{figure}
\begin{figure}[h!t]

\begin{center}
\resizebox{.9\linewidth}{!}{
\centerline{\includegraphics[width=\columnwidth]{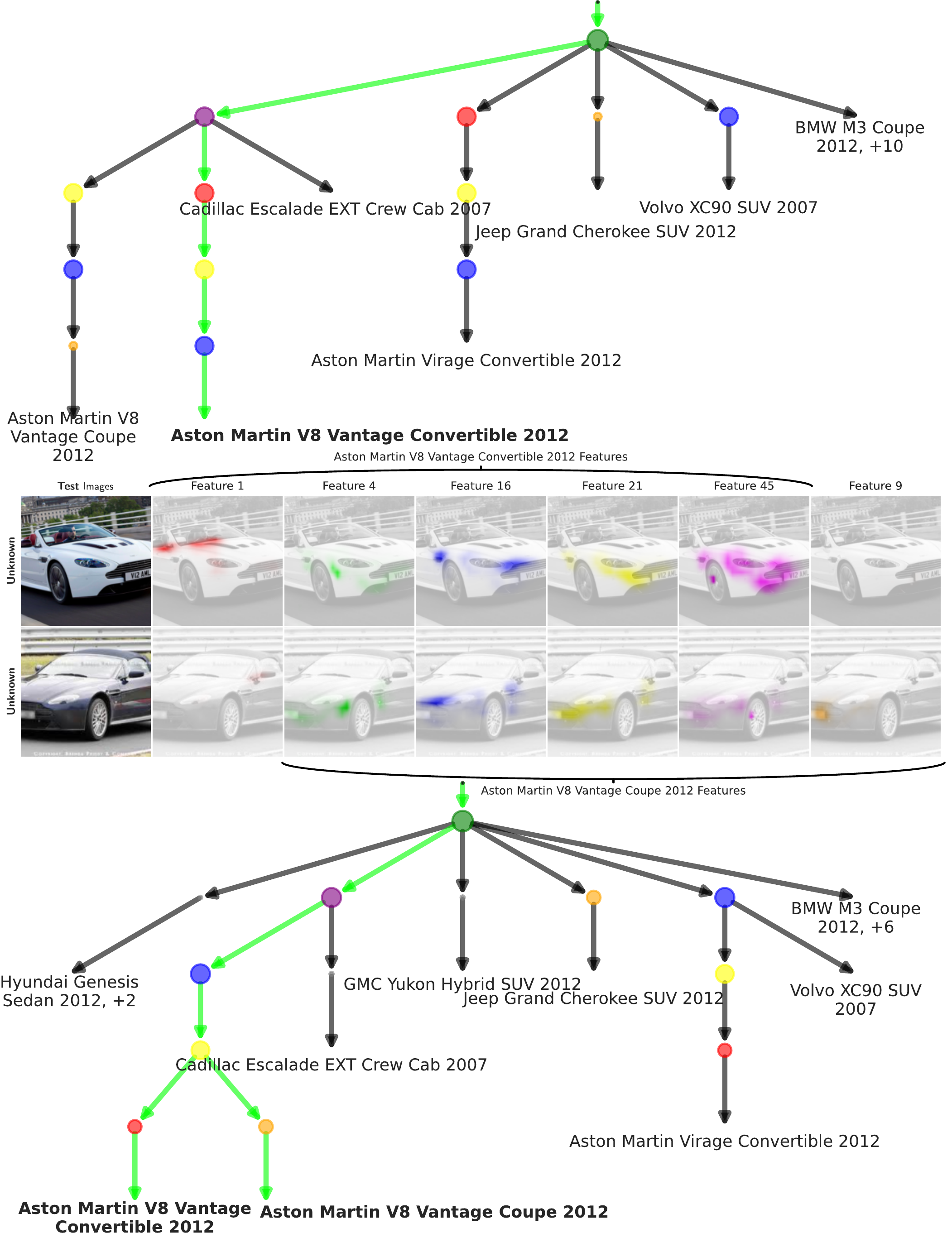}}
}
\caption{Exemplary local explanations provided by our \hqpm{}, with the global explanation in \cref{sfig:C1Glob} for two test images of the Convertible. The first row is an easy example where all $5$ features are found, because the top is down. Therefore, our calibrated model only predicts the Convertible. 
The top is up in the second image, leading to the reasoning of our \hqpm{} along its dynamic class hierarchy identifying it as one of the cars, either Coupe or Convertible because both window features are only barely activating. 
As the global explanation in \cref{sfig:C1Glob} explains, the probed \hqpm{} does not rely on the fabric top,  and hence predicts the set of both cars interpretably.
}
\label{sfig:window2}

\end{center}

\vskip -0.2in
\end{figure}

\begin{figure}[h!t]
\vskip 0.2in
\begin{center}
\centerline{\includegraphics[width=\columnwidth]{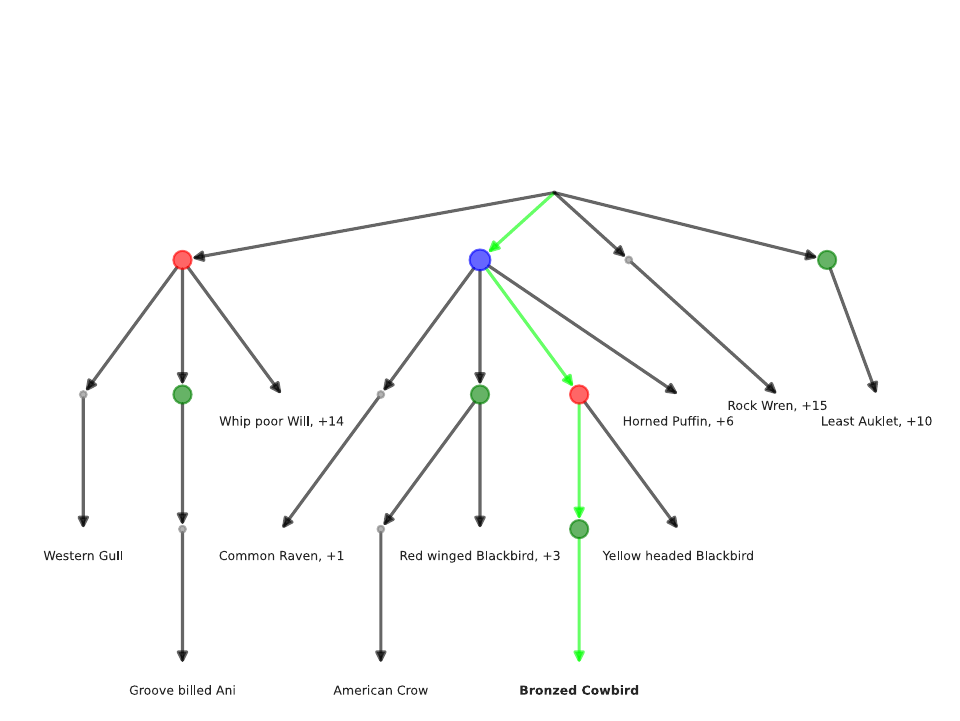}}
\caption{Full graph including all activations for top graph in \cref{sfig:TeaserFull}.}
\label{fig:globgraph1}
\end{center}
\vskip -0.2in
\end{figure}

\begin{figure}[h!t]
\vskip 0.2in
\begin{center}
\centerline{\includegraphics[width=\columnwidth]{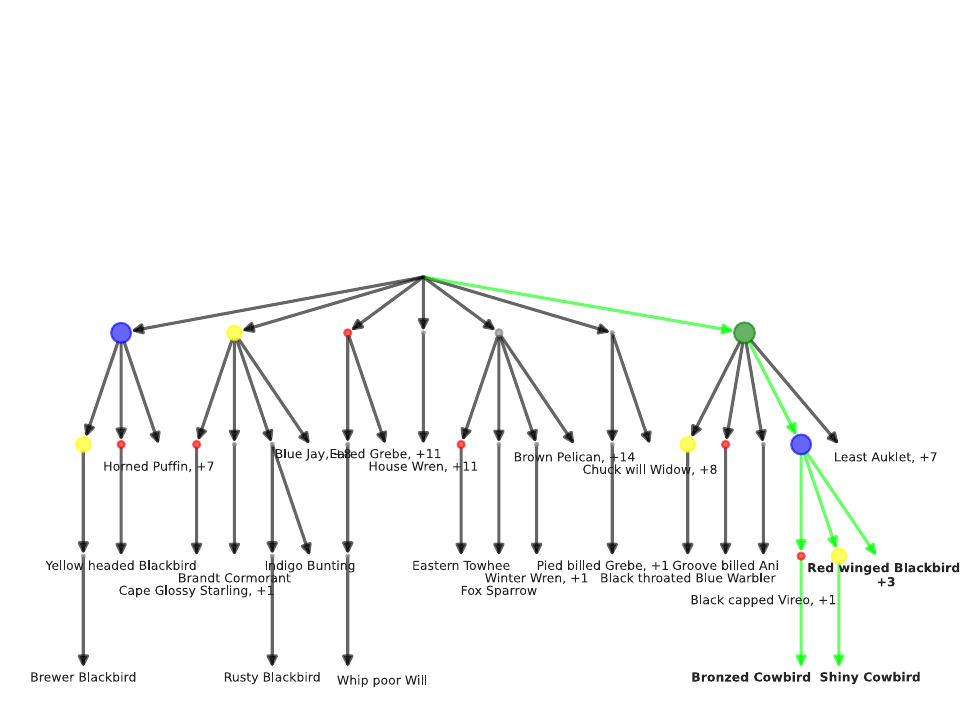}}
\caption{Full graph including all activations for graph in \cref{fig:TeaserFull}.}
\label{fig:globgraph2}
\end{center}
\vskip -0.2in
\end{figure}

\begin{figure}[h!t]
\vskip 0.2in
\begin{center}
\centerline{\includegraphics[width=\columnwidth]{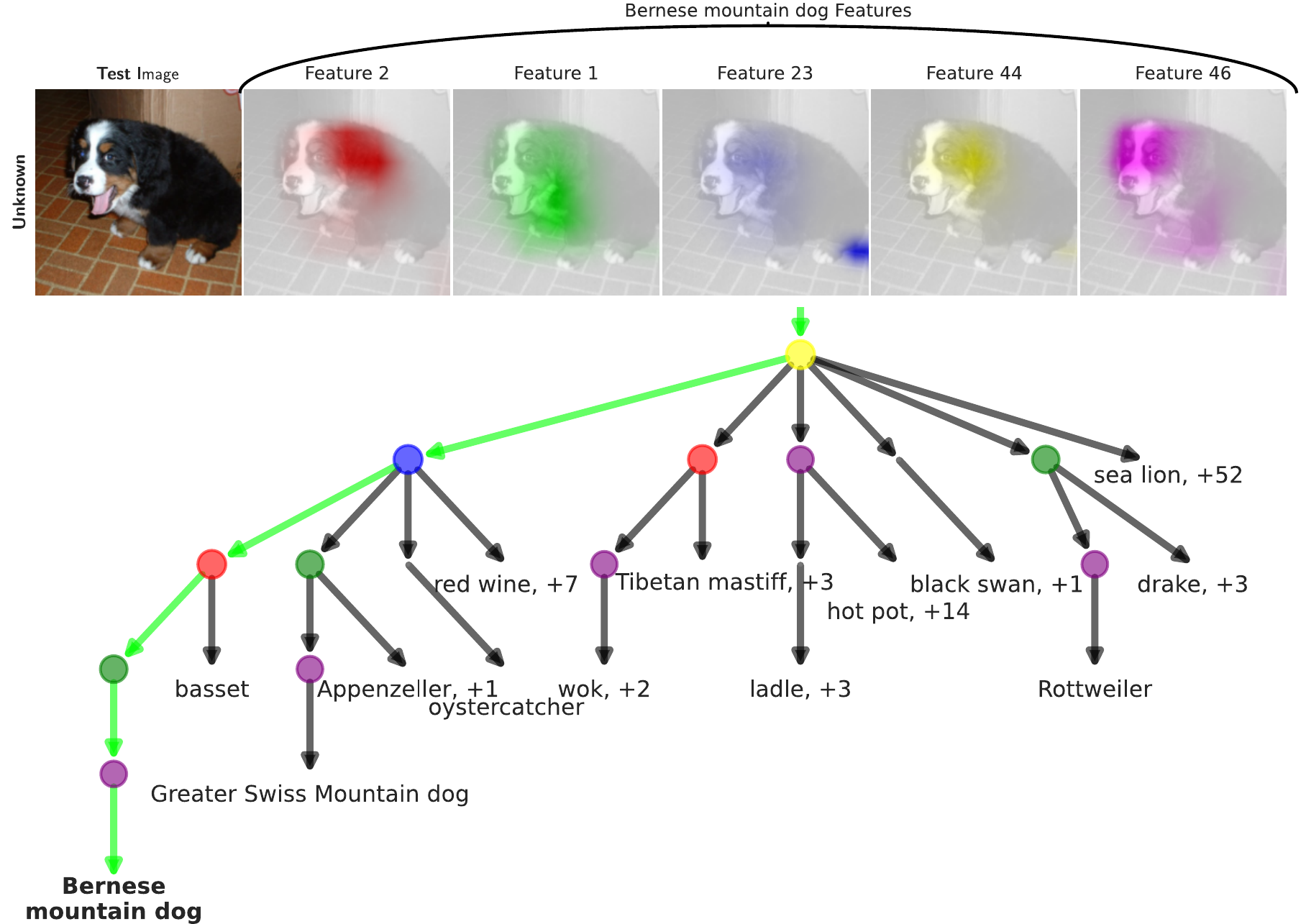}}
\caption{Exemplary local explanation for Bernese mountain dog labeled test sample of \hqpm{} with global explanation in \cref{sfig:I1Glob}. All $5$ features are recognized, hence only the true label is predicted. The tree further gives a glimpse into how the $1000$ classes of \imgnetheader{} are organized.}
\label{sfig:mount1}
\end{center}
\vskip -0.2in
\end{figure}

\begin{figure}[h!t]
\vskip 0.2in
\begin{center}
\centerline{\includegraphics[width=\columnwidth]{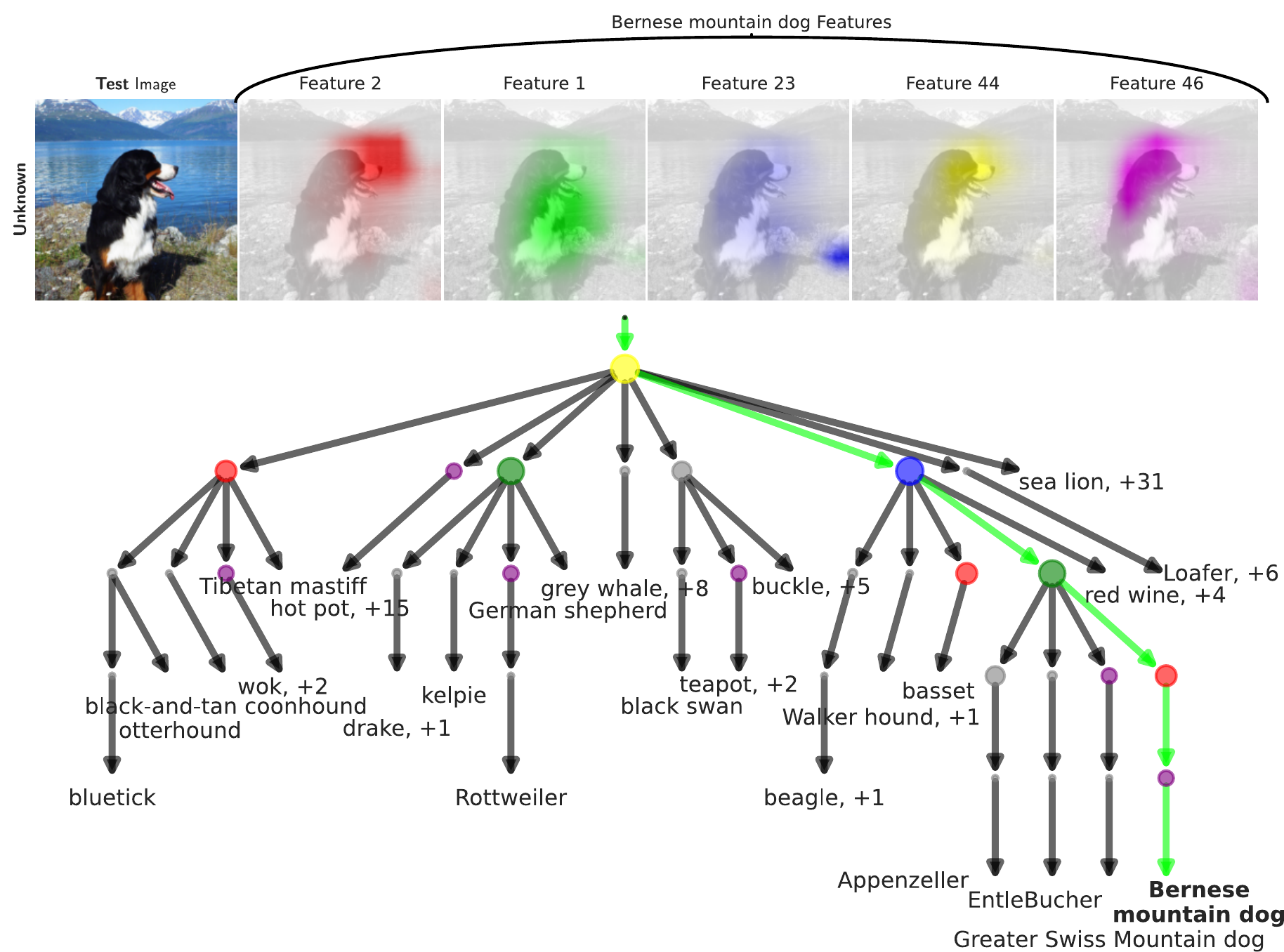}}
\caption{Exemplary local explanation for Bernese mountain dog labeled test sample of \hqpm{} with global explanation in \cref{sfig:I1Glob}. All $5$ features are recognized, hence only the true label is predicted. However, due to the sideways pose, feature $46$, focusing on the frontal face, is less activated. The tree further gives a glimpse into how the $1000$ classes of \imgnetheader{} are organized.}
\label{sfig:mount2}
\end{center}
\vskip -0.2in
\end{figure}

This section comments the extensive qualitative examples.
For all examples, we always present explanations for the same model, i.e. all explanations for a \hqpm{} on \cubheader{} with $\gls{nperClass}=5$ and $\gls{nReducedFeatures}=50$ explain the same model.
Therefore, this section discusses results of $4$ models, $3$ per dataset with the default configuration and also visualizations of \hqpm{} with $\gls{nperClass}=3$ and $\gls{nReducedFeatures}=30$ trained on \cubheader{} and explained  in \cref{fig:TeaserFull,fig:TeaserGlob}.
\Cref{fig:globRed} first contains the global class explanation between the Red-winged blackbird and Bronzed Cowbird, which are jointly predicted in \cref{fig:TeaserFull}.
The \hqpm{} faithfully communicates that it differentiates them based on their unique attributes: The red eye or wing.
Additionally, \cref{sfig:TeaserFull} contrasts the explanation of \cref{fig:TeaserFull} with one for an easy example.
Further contrastive global class explanations with human understandable differences on \hqpm{} trained on \cubheader{}, \stanfordheader{} and \imgnetheader{} are shown in \cref{sfig:C1Glob,sfig:C2Glob,sfig:I1Glob,sfig:I2Glob,fig:globNeck,sfig:sparrows}.
Additionally, \cref{sfig:TeaserFull} displays the extensive local explanations and an example where \hqpm{} dynamically predicts the appropriate set of classes based on visible evidence, as globally explained by \cref{sfig:C1Glob}.
Supporting \cref{fig:TeaserFull,fig:TeaserGlob,sfig:TeaserFull}, \cref{fig:globgraph1,fig:globgraph2} show the corresponding graphs without the limitation of only including classes that share the top feature.
Finally, \cref{sfig:mount1,sfig:mount2,sfig:neckloc2,sfig:necklock2,sfig:spar2,sfig:spar1} show more exemplary local explanations of our \hqpm{}, including the novel hierarchical explanations and saliency maps that also transport activation, as described in \cref{ssec:SaliencyViz}.
They demonstrate how \hqpm{} can provide uniquely comprehensive local interpretability for a single test image, faithfully following its global explanations.

\subsection{Feature Generalization Visualizations} 
\label{sec:appendix_feature_viz} 

\begin{figure}[h!t]
\vskip 0.2in
\begin{center}
\centerline{\includegraphics[width=\columnwidth]{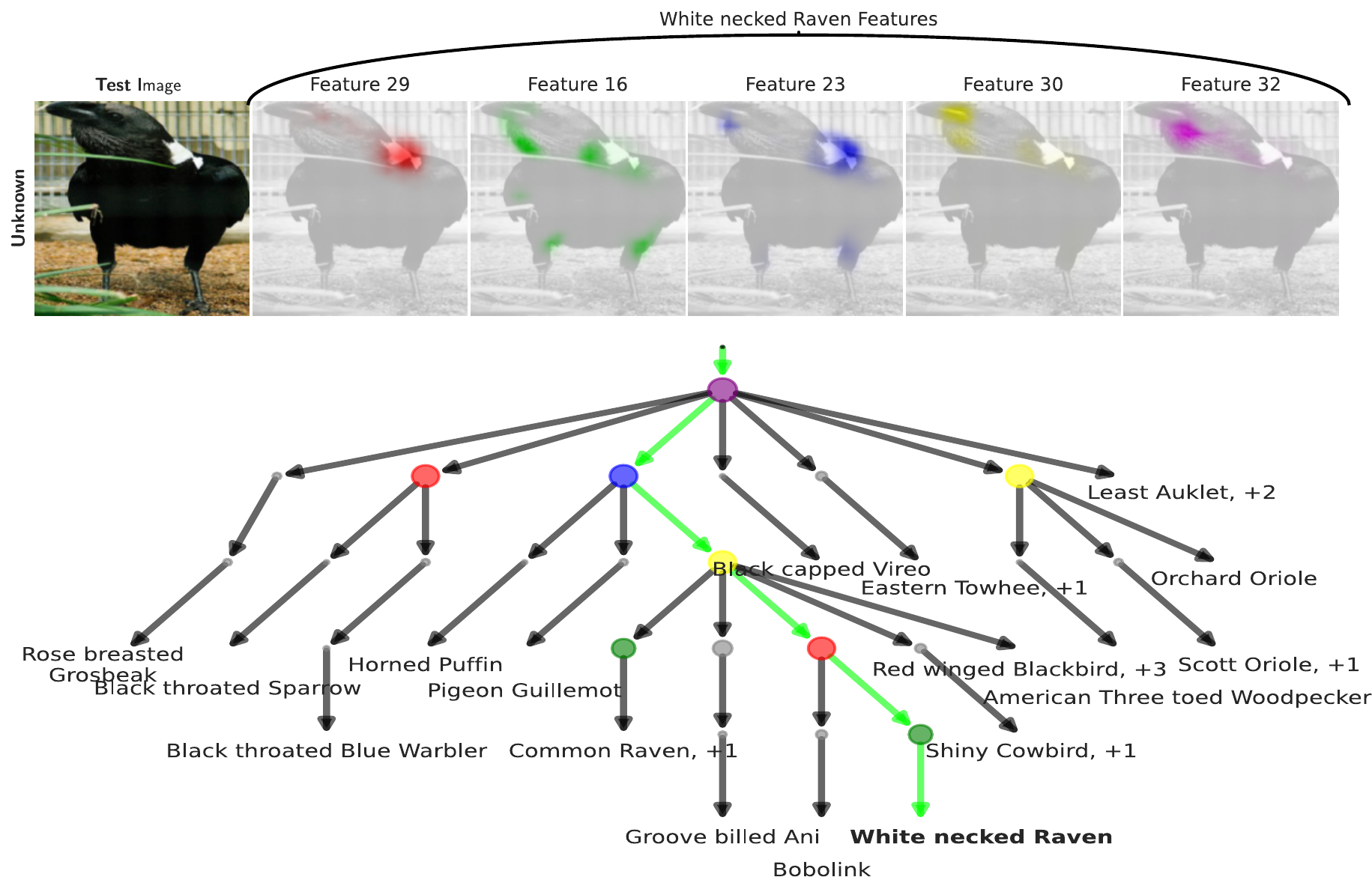}}
\caption{Exemplary local explanation for White necked Raven labeled test sample of \hqpm{} with global explanation in \cref{fig:globNeck}. All $5$ features are recognized, hence only the true label is predicted and the features of the predicted class visualized. The tree further visualizes the learned class similarities, with violet and red leading to black throated birds, whereas violet and yellow seem to indicate \textit{Oriole}.}
\label{sfig:neckloc2}
\end{center}
\vskip -0.2in
\end{figure}

\begin{figure}[h!t]
\vskip 0.2in
\begin{center}
\centerline{\includegraphics[width=\columnwidth]{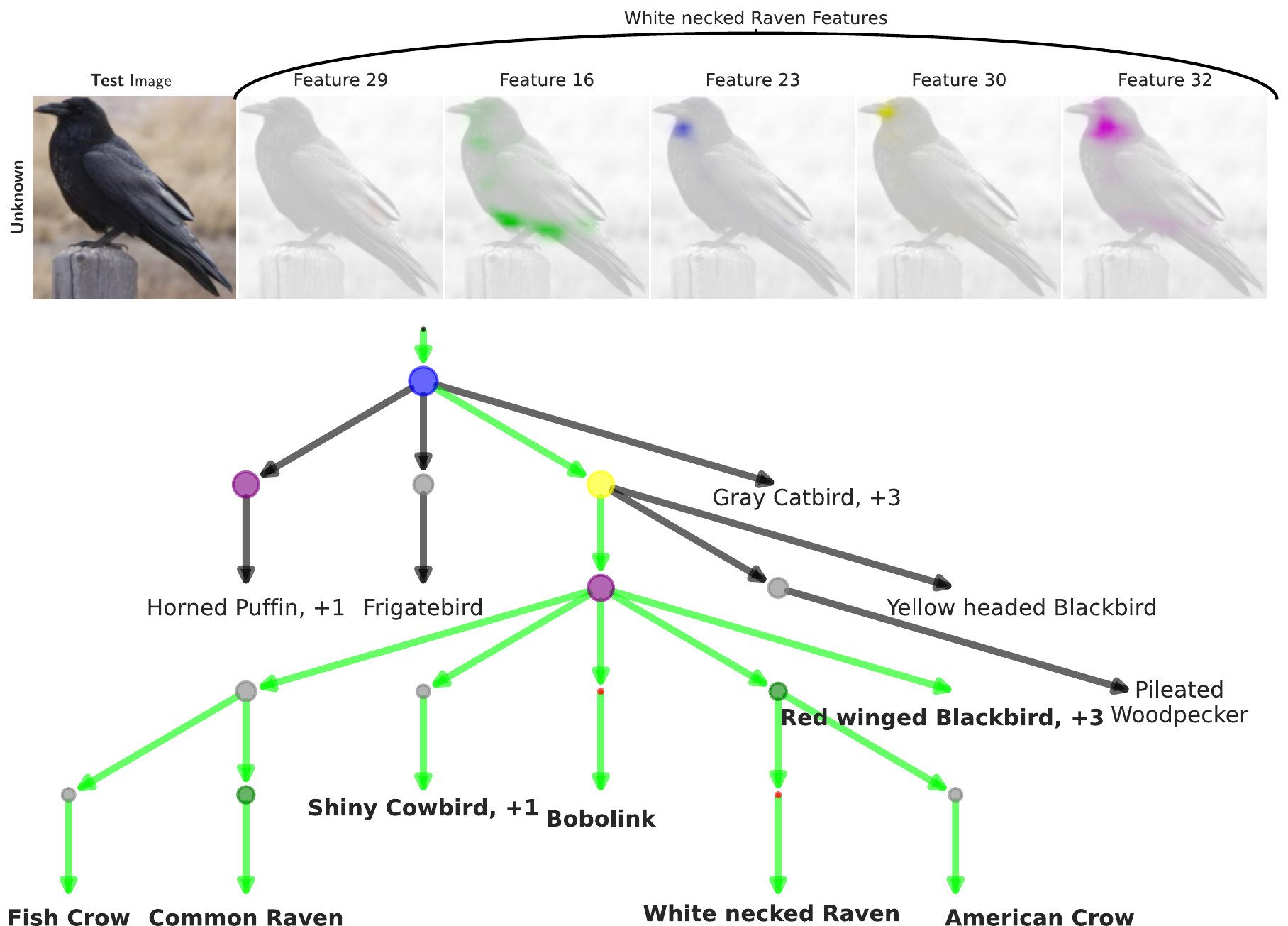}}
\caption{Exemplary local explanation for White necked Raven labeled, presumably mislabelled, test sample of \hqpm{} with global explanation in \cref{fig:globNeck}. Due to missing evidence, especially the white neck, our calibrated \hqpm{} predicts a set of black birds. We show saliency maps  for features of White necked Raven as the reader can have learned about its class representation already. Note that the gray feature for Common Raven is shown in \cref{fig:globNeck} too. When \hqpm{} determines that it needs to predict a set,  one might want to visualize all the features of all classes that are predicted in the coherent set. Alternatively, all active features could be visualized. This is ultimately up to the level of detail desired. }
\label{sfig:necklock2}
\end{center}
\vskip -0.2in
\end{figure}

\begin{figure}[h!t]
\vskip 0.2in
\begin{center}
\centerline{\includegraphics[width=\columnwidth]{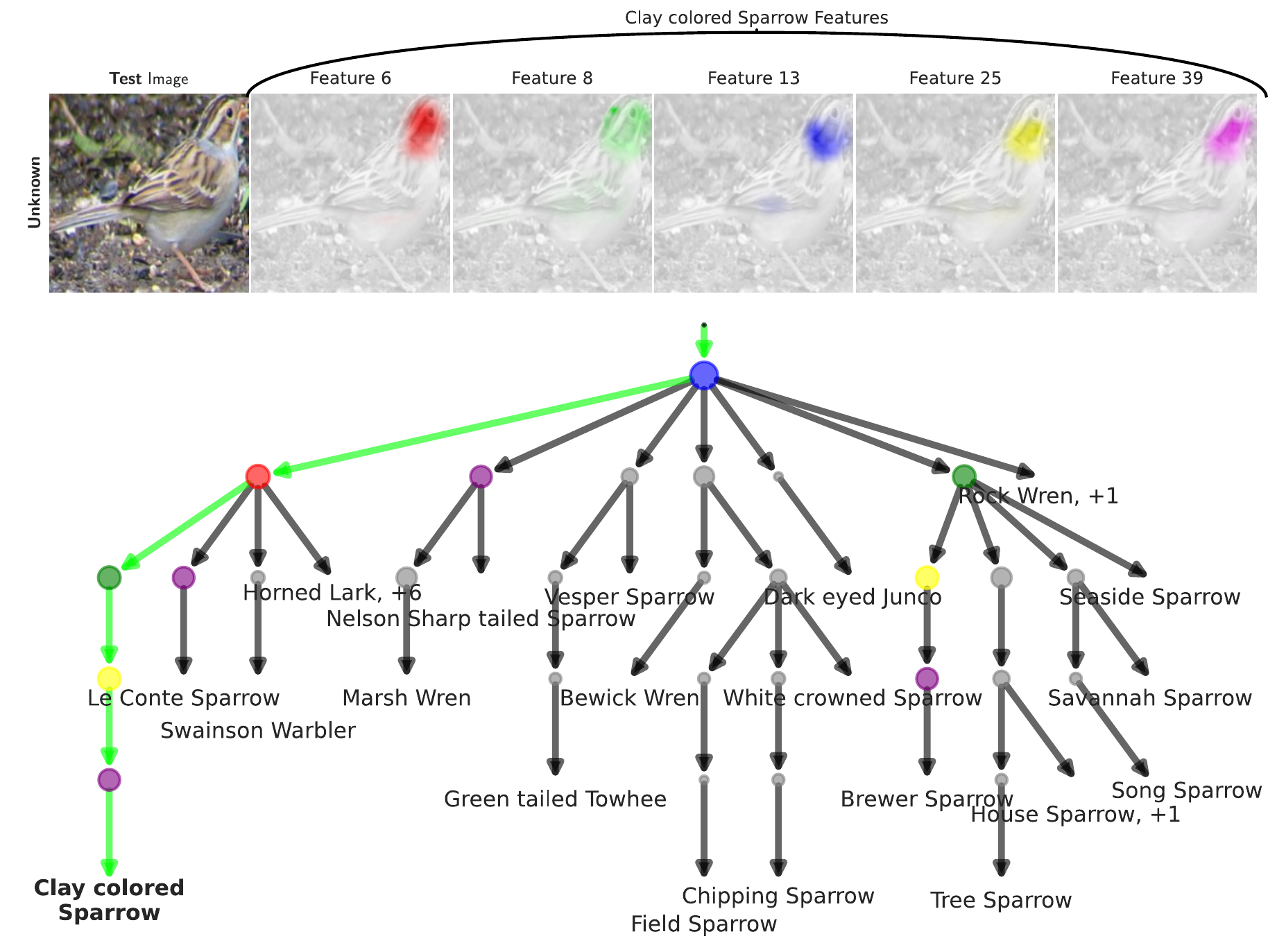}}
\caption{Exemplary local explanation for Clay colored Sparrow labeled test sample of \hqpm{} with global explanation in \cref{sfig:sparrows}.  All $5$ features are recognized, hence only the true label is predicted and the features of the predicted class visualized. Interestingly, the clearly visible white crown stripe, detected by the red feature, is sufficient for \hqpm{} to distinguish it from most other sparrows early on in the hierarchy, as opposed to \cref{sfig:spar2}, where blue and green determine a set of sparrows first.  }
\label{sfig:spar1}
\end{center}
\vskip -0.2in
\end{figure}

\begin{figure}[h!t]
\vskip 0.2in
\begin{center}
\centerline{\includegraphics[width=\columnwidth]{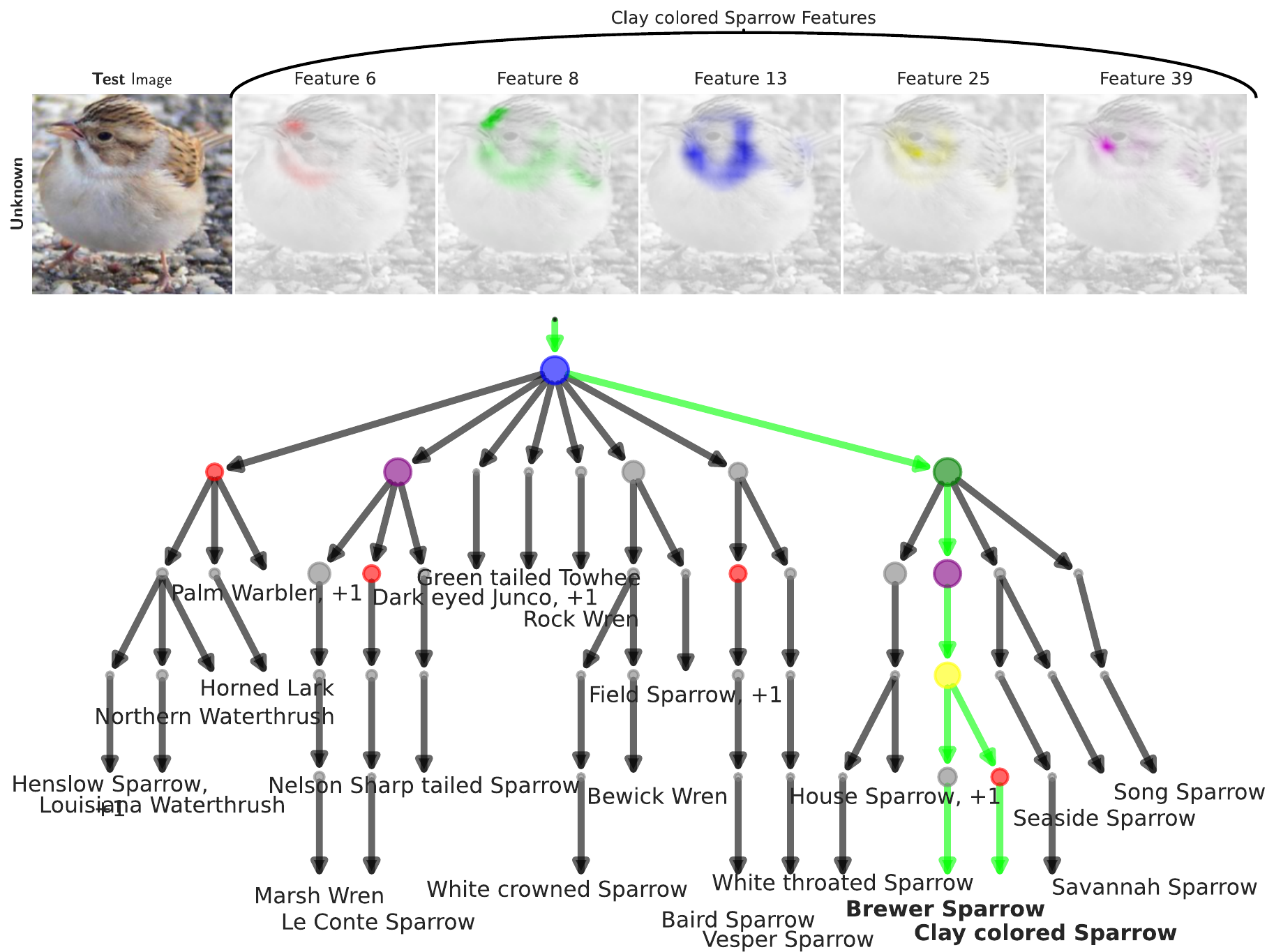}}
\caption{Exemplary local explanation for Clay colored Sparrow labeled test sample of \hqpm{} with global explanation in \cref{sfig:sparrows}.  Both head features activate evenly, thus both sparrows are predicted jointly.}
\label{sfig:spar2}
\end{center}
\vskip -0.2in
\end{figure}

This section provides supplementary visualizations demonstrating that the features learned by our \hqpm{} are not merely class-specific detectors but function as general concept detectors across a diverse range of images.
\Cref{sfig:FeaturesCover,sfig:FeaturesBlack,sfig:SmallTern,sfig:BigTern} visualize how the features of the probed CHiQPMs explained in \cref{fig:TeaserFull,fig:TeaserGlob,sfig:TeaserFull,fig:globRed,sfig:spar1,sfig:spar2,sfig:sparrows,sfig:neckloc2,fig:globNeck,sfig:necklock2} generalize across a huge range of images and classes. 
For that, we visualized all classes that share all but one feature with one class.
We choose Shiny Cowbird for \cref{sfig:FeaturesCover,sfig:FeaturesBlack}, as it has a large neighborhood in the probed \hqpm{}'s class representations and is also used in the main paper in \cref{fig:TeaserFull,fig:TeaserGlob}.
Additionally, we show the features of all classes similarly represented to Arctic Tern in \cref{sfig:BigTern,sfig:SmallTern}, as it is another class with a large neighborhood in both models.
We further supplemented \cref{sfig:FeaturesBlack} with the Raven classes from \cref{sfig:neckloc2,sfig:necklock2,fig:globNeck} to showcase that the features localize consistently, not just directly around the chosen class.
These visualizations demonstrate that similarly represented classes by  \hqpm{} are indeed similar in reality, as measured by \cubsim, and that the features are general concept detectors. 


\newcommand{\featureTFigures}{, \eg{} explained in \cref{fig:TeaserFull,fig:TeaserGlob,sfig:TeaserFull,fig:globRed}.}
\newcommand{\featureFFigures}{, \eg{} explained in \cref{sfig:spar1,sfig:spar2,sfig:sparrows,sfig:neckloc2,fig:globNeck,sfig:necklock2}.}






\begin{figure}[h!] 
    \centering
    \includegraphics[width=0.9\textwidth]{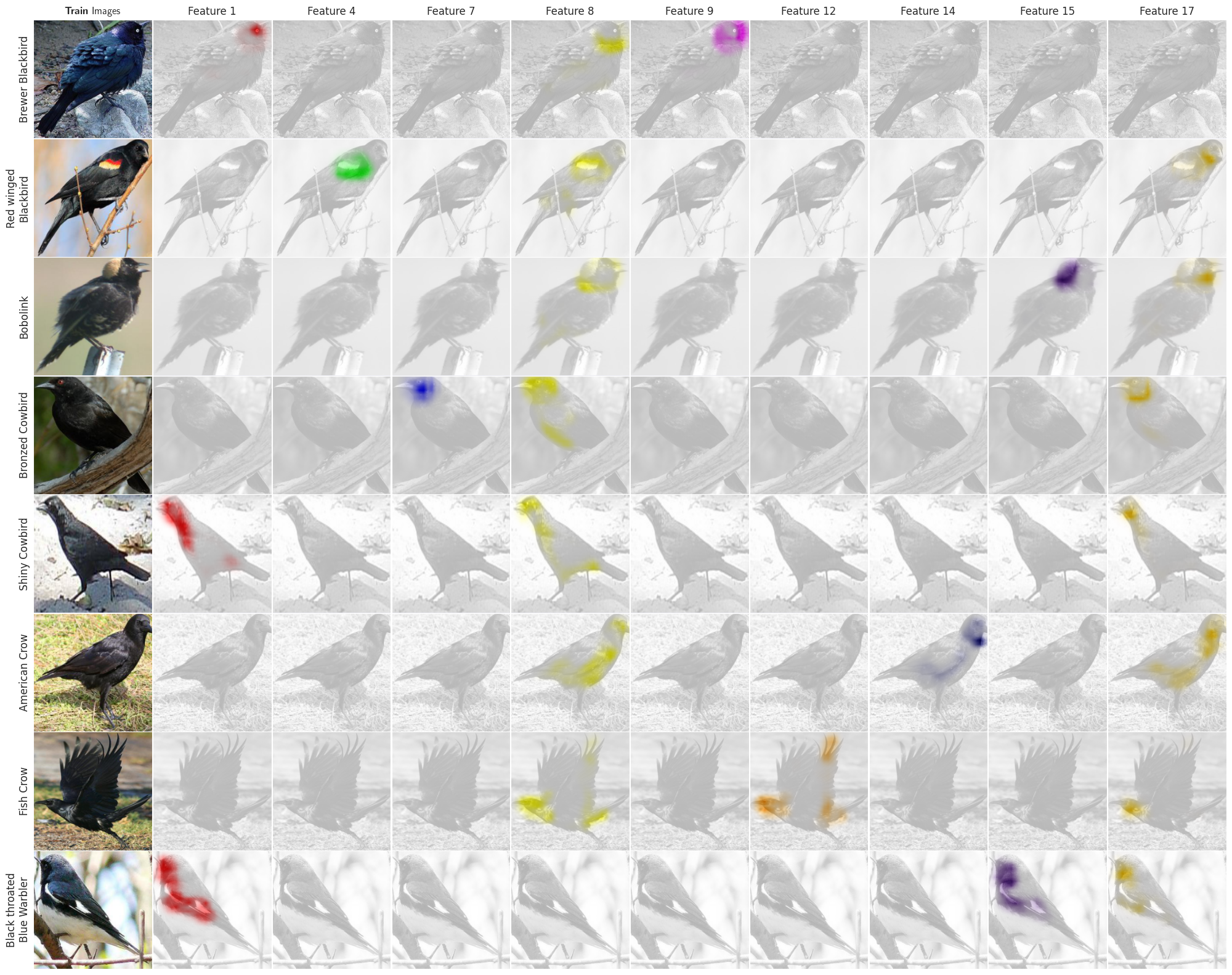} 
    \caption{Visualizations for classes similar to Shiny Cowbird using the model with 3 features per class\featureTFigures } 
    \label{sfig:FeaturesCover}
\end{figure}

\begin{figure}[h!] 
    \centering
    \includegraphics[width=0.9\textwidth]{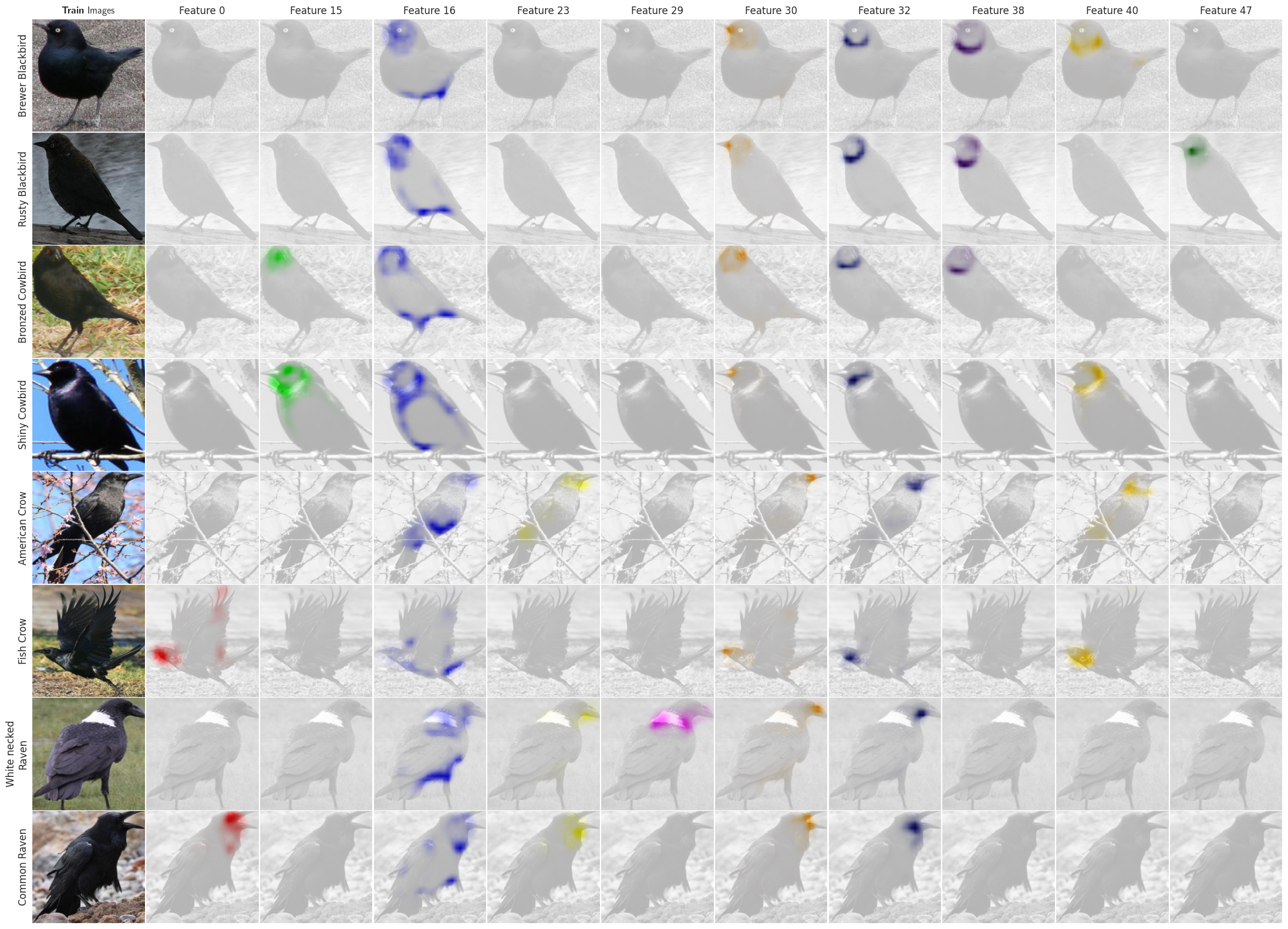}
    \caption{Visualizations for classes similar to Shiny Cowbird using the model with  $\gls{nperClass}=5$ features per class\featureFFigures We included the Raven classes from \cref{sfig:neckloc2,sfig:necklock2,fig:globNeck} for reference.}
     \label{sfig:FeaturesBlack}
\end{figure}



\begin{figure}[h!] 
    \centering
    \includegraphics[width=0.9\textwidth]{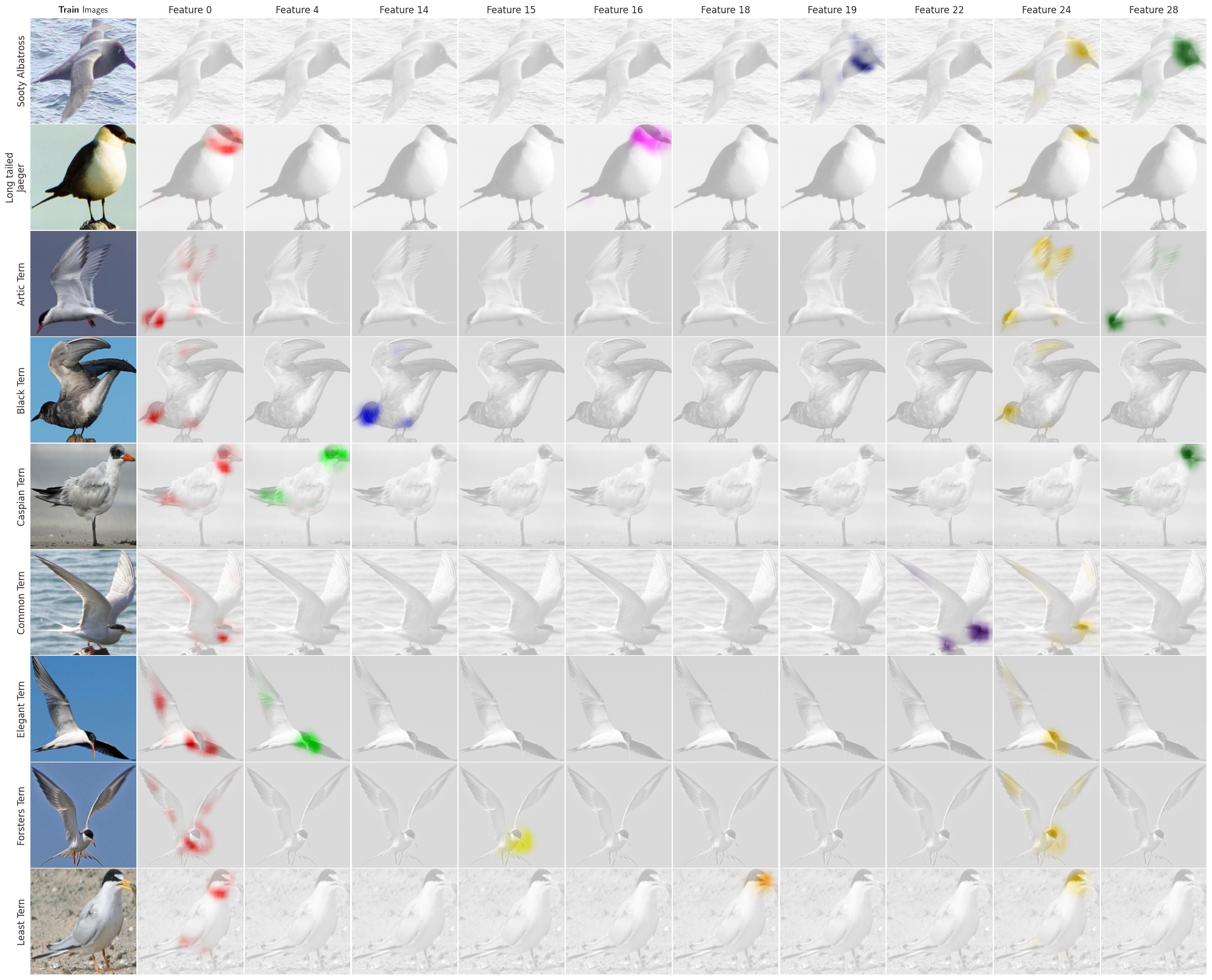} 
    \caption{Visualizations for classes similar to Arctic Tern using the model with  $\gls{nperClass}=3$ features per class\featureTFigures}
         \label{sfig:SmallTern}
\end{figure}

\begin{figure}[h!] 
    \centering
    \includegraphics[width=0.9\textwidth]{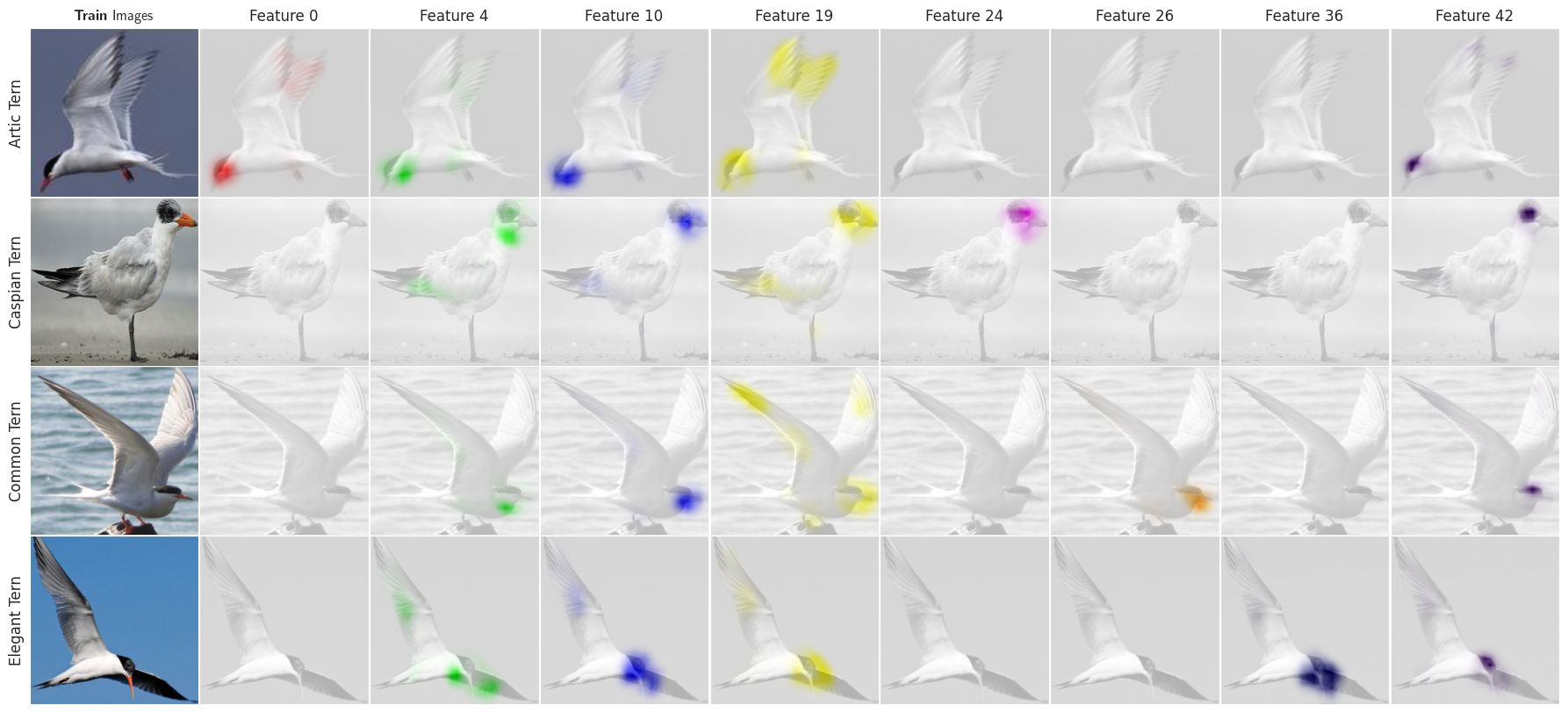} 
    \caption{Visualizations for classes similar to Arctic Tern using the model with $\gls{nperClass}=5$ features per class\featureFFigures}
    \label{sfig:BigTern}
\end{figure}
\input{secs/icmlsideways}

%% file: secs/icmlsideways.tex
\begin{table}
\centering

\caption{
\accmetricstablestart{\resnet{}}
\cref{fig:dimVar} shows the increasing gap when further raising compactness. 
\boldnessstatement
}
\label{stab:aCCproto-table}
\resizebox{\textwidth}{!}{
\begin{tabular}{lccccccccc}
\toprule
Method & \multicolumn{3}{c}{Accuracy \arrowUp} &
\multicolumn{3}{c}{Total Features\arrowDown} & \multicolumn{3}{c}{Features / Class\arrowDown} \\
& CUB& CARS & IMGNET & CUB& CARS & IMGNET & CUB& CARS & IMGNET
\\
\midrule
Baseline \resnet{} & {86.6}$\pm$0.2 & {92.1}$\pm$0.1 & 76.1 & 2048 & 2048 & 2048& 2048& 2048 &2048\\
\midrule
\glmtable{} & {78.0}$\pm$0.4 & {86.8}$\pm$0.6 & 58.0$\pm$0.0 & 809$\pm$8 & 807$\pm$10 & 1627$\pm$1 & \textbf{5}& \textbf{5} &\textbf{5} \\
\pipnettable{} & 82.0$\pm$0.3 & 86.5$\pm$0.3 &-& 731$\pm$19 &669$\pm$13 &-& 12 &11 & - \\ 
\protopooltable{} & 79.4$\pm$0.4 &87.5$\pm$0.2 &-& 202 & 195 & - &202 & 195 & - \\
\slddtable{} & {84.5}$\pm$0.2 & {91.1}$\pm$0.1 & 72.7$\pm$0.0 & \textbf{50} & \textbf{50} & \textbf{50}& \textbf{5}& \textbf{5} &\textbf{5} \\
Q-SENN  & {84.6}$\pm$0.3 & \underline{91.8}$\pm$0.3 & \underline{74.3}$\pm$0.0 & \textbf{50} & \textbf{50} & \textbf{50}& \textbf{5}& \textbf{5} &\textbf{5} \\
\gls{NewlayerName} & \underline{85.1}$\pm$0.3 & \underline{91.8}$\pm$0.3 & {74.2}$\pm$0.0 & \textbf{50} & \textbf{50} & \textbf{50}& \textbf{5}& \textbf{5} &\textbf{5} \\
\midrule
\tabOurs & \textbf{85.3}$\pm$0.3 & \textbf{91.9}$\pm$0.2  & \textbf{75.3}$\pm$0.0 & \textbf{50} &\textbf{50}& \textbf{50}& \textbf{5}& \textbf{5} &\textbf{5} \\

\bottomrule
\end{tabular}
}
\end{table}

\begin{table}

\centering
\caption{
\interpmetricstablestart{\resnet{}}}
\label{stab:res50Interp}
\resizebox{\textwidth}{!}{
\begin{tabular}{lccccccccccc}
\toprule
Method & \multicolumn{3}{c}{\loc{5} \arrowUp} &
\multicolumn{3}{c}{\generality{}\arrowUp} & \multicolumn{3}{c}{\contrastiveness{}\arrowUp} & \multicolumn{1}{c}{\cubsim{}\arrowUp} \\
& CUB& CARS & IMGNET & CUB& CARS & IMGNET & CUB& CARS & IMGNET & CUB
\\
\midrule
Baseline \resnet{} & {57.7}$\pm$0.4 & {54.4}$\pm$0.3 &37.1 & {98.0}$\pm$0.0 & {97.8}$\pm$0.0 & 99.4& {74.4}$\pm$0.1 & {75.1}$\pm$0.1 &71.6 & {34.0}$\pm$0.3 \\
\midrule
\glmtable{} & {55.4}$\pm$0.5 & {51.8}$\pm$0.3 &35.8$\pm$0.0 & \textbf{97.8}$\pm$0.0 & \textbf{97.6}$\pm$0.0 & \textbf{99.4}$\pm$0.0 & {74.0}$\pm$0.1 & {74.5}$\pm$0.1 &71.7$\pm$0.0 & {2.5}$\pm$1.0  \\
\pipnettable{} & \textbf{99.2}$\pm$0.1 & \textbf{99.0}$\pm$0.1 &- & 75.6$\pm$0.4 &62.9$\pm$0.1 &- & \underline{99.6}$\pm$0.0 &\underline{99.7}$\pm$0.0 &- & 6.7$\pm$0.9 \\ 
\protopooltable{} & 24.5$\pm$0.8 &30.7$\pm$3.4 &-& 96.9$\pm$0.1 & 96.0$\pm$0.5 &- &76.7$\pm$1.0& 78.9$\pm$2.0 &- & 13.9$\pm$0.9 \\
\slddtable{}
    & {88.2}$\pm$0.2 & {88.6}$\pm$0.6 &\textbf{64.7}$\pm$0.7 & {96.2}$\pm$0.1 & {95.5}$\pm$0.1 &{98.6}$\pm$0.0 & {87.3}$\pm$0.2 & {89.7}$\pm$0.3 &\underline{93.4}$\pm$0.1 & {29.2}$\pm$4.0 \\
 Q-SENN    &  \underline{93.2}$\pm$0.4 & \underline{94.3}$\pm$0.3 & -  & {95.5}$\pm$0.1 & {94.8}$\pm$0.1 & -   & {93.0}$\pm$0.3 & {93.9}$\pm$0.2 & -   &{20.8}$\pm$4.1 & \\
\gls{NewlayerName} & {90.1}$\pm$0.3 & {89.6}$\pm$0.4 &\underline{64.1}$\pm$0.7 & \underline{97.0}$\pm$0.0 & \underline{96.5}$\pm$0.0 &\underline{99.1}$\pm$0.0 & {96.0}$\pm$0.4 & {97.7}$\pm$0.4 &{89.3}$\pm$0.1 & \underline{47.9}$\pm$2.7 \\
\midrule
\tabOurs &  {88.1}$\pm$0.5 & {88.8}$\pm$1.0 & 42.9$\pm$0.9 &  {94.1}$\pm$0.0 & {93.5}$\pm$0.1 & 98.7$\pm$0.0 &   \textbf{99.9}$\pm$0.0 & \textbf{100.0}$\pm$0.0 & \textbf{99.9}$\pm$0.0 & \textbf{75.0}$\pm$2.2 \\
\bottomrule
\end{tabular}
}
\end{table}

\begin{table}[htbp]
  \caption{
   \accmetricstablestart{Resnet34}
   \boldnessstatement
  }
  \label{stab:r34aCCproto-table}
  \centering
  \begin{tabular}{lcccccc}
    \toprule
    Method &  \multicolumn{2}{c}{Accuracy \arrowUp} &
    \multicolumn{2}{c}{Total Features\arrowDown} & \multicolumn{2}{c}{Features / Class\arrowDown} \\
   & CUB& CARS & CUB& CARS  & CUB& CARS \\
    \midrule
    Baseline Resnet34 & 85.7$\pm$0.3 & 91.5$\pm$0.2   & 2048 & 2048 &  2048& 2048 \\
    \midrule
    \glmtable{} & {72.0}$\pm$1.0 & {82.0}$\pm$0.6 & 442$\pm$5 & 453$\pm$6  & \textbf{5}& \textbf{5} \\
    \slddtable{} & {83.2}$\pm$0.3 & {90.7}$\pm$0.3   & \textbf{50} & \textbf{50} &  \textbf{5}& \textbf{5} \\
   Q-SENN  &  \textbf{83.7}$\pm$0.2 & \underline{91.3}$\pm$0.3 & \textbf{50} & \textbf{50} &  \textbf{5}& \textbf{5} \\
     \gls{NewlayerName}  & {83.0}$\pm$0.2 & \underline{91.3}$\pm$0.0 & \textbf{50} & \textbf{50} &  \textbf{5}& \textbf{5} \\
       \midrule
\tabOurs &  \textbf{83.7}$\pm$0.2 & \textbf{91.5}$\pm$0.2& \textbf{50} &\textbf{50}& \textbf{5}& \textbf{5}\\
    \bottomrule
  \end{tabular}
\end{table}
\begin{table}[htbp]
  \caption{
  \interpmetricstablestart{Resnet34}}
  \label{stab:r34Interpproto-table}
  \centering
 \resizebox{\textwidth}{!}{
  \begin{tabular}{lccccccc}
    \toprule
    Method &  \multicolumn{2}{c}{\loc{5} \arrowUp} &
    \multicolumn{2}{c}{\generality{}\arrowUp} & \multicolumn{2}{c}{\contrastiveness{}\arrowUp} & \multicolumn{1}{c}{\cubsim{}\arrowUp} \\
  & CUB& CARS & CUB& CARS & CUB& CARS & CUB
    \\
    \midrule
    Baseline Resnet34  & {62.1}$\pm$0.3 & {56.6}$\pm$0.4  & {97.9}$\pm$0.0 & {97.7}$\pm$0.0  & {76.4}$\pm$0.1 & {77.9}$\pm$0.2  & {39.6}$\pm$0.2 \\
    \midrule
    \glmtable{}  & {59.9}$\pm$0.4 & {55.3}$\pm$0.3 
  & \textbf{97.9}$\pm$0.0 & \textbf{97.7}$\pm$0.0 & {76.5}$\pm$0.0 & {77.8}$\pm$0.2 &  {7.6}$\pm$2.2 \\
    \slddtable{}
     & \underline{90.1}$\pm$0.8 & {86.7}$\pm$2.5  & \underline{97.5}$\pm$0.0 & \underline{97.6}$\pm$0.2 & {86.0}$\pm$1.0 & {83.3}$\pm$4.6 & {24.5}$\pm$2.7  \\
     Q-SENN & {86.6}$\pm$1.4 & {80.1}$\pm$2.8  & {96.6}$\pm$0.1 & {96.0}$\pm$0.2  & {95.2}$\pm$1.0 & {94.3}$\pm$1.4  & {25.7}$\pm$2.0   \\
     \gls{NewlayerName}  & \textbf{90.5}$\pm$0.5 & \textbf{89.1}$\pm$1.1 & \underline{97.5}$\pm$0.0 & {96.9}$\pm$0.1  &  \underline{95.5}$\pm$0.2 & \underline{94.7}$\pm$1.1 & \underline{39.0}$\pm$2.9 \\
     \midrule
   \tabOurs  & {87.9}$\pm$0.5 & \underline{88.4}$\pm$1.0  & {94.1}$\pm$0.1 & {82.9}$\pm$4.5  & \textbf{99.9}$\pm$0.0 & \textbf{100.0}$\pm$0.0  & \textbf{68.5}$\pm$4.1 \\
    \bottomrule
  \end{tabular}
  }
\end{table}

\begin{table}[htbp]
  \caption{
  \accmetricstablestart{\gls{incv}}
  \boldnessstatement
  }
  \label{stab:incaCCproto-table}
  \centering
  \begin{tabular}{lcccccc}
    \toprule
    Method &  \multicolumn{2}{c}{Accuracy \arrowUp} &
    \multicolumn{2}{c}{Total Features\arrowDown} & \multicolumn{2}{c}{Features / Class\arrowDown} \\
   & CUB& CARS & CUB& CARS & CUB& CARS 
    \\
    \midrule
    Baseline  \gls{incv} & {86.1}$\pm$0.1 & {92.6}$\pm$0.2 & 2048 & 2048 & 2048& 2048\\
    \midrule
    \glmtable{} & {79.2}$\pm$0.5 & {89.3}$\pm$0.3 & 814$\pm$9 & 795$\pm$8 & \textbf{5}& \textbf{5} \\
    \slddtable{} & {83.1}$\pm$0.4 & {91.1}$\pm$0.2 &  \textbf{50} & \textbf{50} & \textbf{5}& \textbf{5} \\
    Q-SENN & {83.8}$\pm$0.4 & {91.6}$\pm$0.2 &  \textbf{50} & \textbf{50} & \textbf{5}& \textbf{5}  \\
     \gls{NewlayerName}  & \underline{84.2}$\pm$0.4 & \underline{{91.7}}$\pm$0.1 & \textbf{50} & \textbf{50} &  \textbf{5}& \textbf{5} \\ 
      \midrule
    \tabOurs   & \textbf{84.3}$\pm$0.3 & \textbf{92.1}$\pm$0.2  & \textbf{50} &\textbf{50}& \textbf{5}& \textbf{5}  \\
    \bottomrule
  \end{tabular}
\end{table}
\begin{table}[htbp]
  \caption{
  \interpmetricstablestart{\gls{incv}}}
  \label{stab:incInterpproto-table}
  \centering
  \resizebox{\textwidth}{!}{
  \begin{tabular}{lcccccccc}
    \toprule
    Method &  \multicolumn{2}{c}{\loc{5} \arrowUp} &
    \multicolumn{2}{c}{\generality{}\arrowUp} & \multicolumn{2}{c}{\contrastiveness{}\arrowUp} & \multicolumn{1}{c}{\cubsim{}\arrowUp} \\
  & CUB& CARS & CUB& CARS & CUB& CARS & CUB
    \\
    \midrule
    Baseline  \gls{incv} & {38.9}$\pm$0.3 & {33.1}$\pm$0.2  & {96.1}$\pm$0.0 & {95.7}$\pm$0.0  & {89.6}$\pm$0.2 & {91.7}$\pm$0.2 & {7.1}$\pm$9.6  \\
    \midrule
    \glmtable{}  & {39.3}$\pm$0.2 & {34.0}$\pm$0.4  & \textbf{95.4}$\pm$0.0 & \textbf{95.0}$\pm$0.0  & {91.3}$\pm$0.3 & {93.4}$\pm$0.2  & {0.3}$\pm$0.4 \\
    \slddtable{}
      & \textbf{58.1}$\pm$1.2 & \textbf{52.1}$\pm$1.5  & {92.6}$\pm$0.1 & {92.1}$\pm$0.1  & {93.0}$\pm$0.3 & {94.4}$\pm$0.2 & {24.4}$\pm$2.3  \\
     Q-SENN & \underline{55.6}$\pm$0.9 & \underline{48.2}$\pm$0.9  & {92.2}$\pm$0.2 & {91.5}$\pm$0.1  &  \underline{94.6}$\pm$0.3 &  \underline{95.1}$\pm$0.4  & {19.7}$\pm$2.6  \\
      \gls{NewlayerName}  & {48.6}$\pm$0.9 & {42.8}$\pm$0.8  & \underline{95.1}$\pm$0.1 & \underline{94.7}$\pm$0.0  & {93.4}$\pm$0.1 &{94.3}$\pm$0.1 &  \underline{34.8}$\pm$3.4 \\
       \midrule
  \tabOurs   & {45.4}$\pm$1.3 & {37.8}$\pm$2.2  & {93.4}$\pm$0.1 & {92.7}$\pm$0.2  & \textbf{100.0}$\pm$0.0 & \textbf{100.0}$\pm$0.0  & \textbf{52.8}$\pm$4.5 \\
    \bottomrule
  \end{tabular}
  }
\end{table}

\begin{table}[htbp]
  \caption{  \accmetricstablestart{Swin Transformer small}
  \boldnessstatement  }
  \label{sstab:swcaCCproto-table}
  \centering
  \begin{tabular}{lcccccc}
    \toprule    Method  &   \multicolumn{2}{c}{Accuracy \arrowUp}  &     \multicolumn{2}{c}{Total Features\arrowDown}  &  \multicolumn{2}{c}{Features / Class\arrowDown}  \\
    &  CUB &  CARS  &  CUB &  CARS  &  CUB &  CARS  \\
    \midrule
    Baseline  Swin Transformer small  &  {87.0}$\pm$0.1  &  {90.6}$\pm$0.6  &  768  &  768  &  768 &  768  \\
    \midrule
    \glmtable{}   &  {76.5}$\pm$0.4  &  {75.5}$\pm$1.2  &  572$\pm$4  &  559$\pm$8  &  \textbf{5} &  \textbf{5}  \\
    \slddtable{}  &    \underline{85.3}$\pm$0.4   &  \textbf{89.1}$\pm$0.7  &   \textbf{50}  &  \textbf{50}  &  \textbf{5} &  \textbf{5}  \\
\hline
      \gls{NewlayerName}    &  {85.0}$\pm$0.4  &   {88.7}$\pm$0.5  &  \textbf{50}  &  \textbf{50}  &   \textbf{5} &  \textbf{5}  \\
\midrule
 \tabOurs   & \textbf{85.9}$\pm$0.3 & \underline{89.0}$\pm$1.6 &  \textbf{50}  &  \textbf{50}  &   \textbf{5} &  \textbf{5}  \\
    \bottomrule
  \end{tabular}
  \end{table}
  \begin{table}[htbp]
  \caption{  \interpmetricstablestart{Swin Transformer small}}
  \label{sstab:swInterpproto-table-adapted} 
   \resizebox{\textwidth}{!}{
  \begin{tabular}{lccccccc}
    \toprule
    Method  &   \multicolumn{2}{c}{\loc{5} \arrowUp}  &     \multicolumn{2}{c}{\generality{}\arrowUp}  &  \multicolumn{2}{c}{\contrastiveness{}\arrowUp}  &  \multicolumn{1}{c}{\cubsim{}\arrowUp}  \\
    &  CUB &  CARS  &  CUB &  CARS  &  CUB &  CARS  &  CUB \\
    \midrule
    Baseline  Swin Transformer small    &  {26.4}$\pm$0.1  &  {26.0}$\pm$0.1  &  {96.8}$\pm$0.0  &  {96.6}$\pm$0.0  &  {98.3}$\pm$0.1  &  {98.8}$\pm$0.1  &  {24.5}$\pm$0.4  \\
    \midrule
    \glmtable{}  &  {26.4}$\pm$0.2  &  {26.1}$\pm$0.1  &   \textbf{96.6}$\pm$0.0  &  \textbf{96.4}$\pm$0.0  &  \underline{99.1}$\pm$0.1  &  \underline{99.6}$\pm$0.0  &  {8.8}$\pm$2.8   \\
    \slddtable{}
      &  \underline{38.0}$\pm$0.5  &  \underline{35.6}$\pm$1.1  &  {93.4}$\pm$0.1  &  {93.3}$\pm$0.2  &  {99.0}$\pm$0.2  &  {99.4}$\pm$0.2  &  {37.2}$\pm$3.4  \\
    \hline
     \gls{NewlayerName}  &  {33.6}$\pm$0.4  &  {32.0}$\pm$0.3  &  \underline{95.2}$\pm$0.0  &  \underline{94.7}$\pm$0.0  &   {98.5}$\pm$0.3  &  {99.1}$\pm$0.2  &  \underline{45.1}$\pm$3.2  \\
    \midrule
    \tabOurs  & \textbf{46.1}$\pm$0.9 & \textbf{45.8}$\pm$4.0  & {94.2}$\pm$0.1 & {93.7}$\pm$0.1   &
    \textbf{99.9}$\pm$0.0 & \textbf{99.9}$\pm$0.1   &
    \textbf{65.4}$\pm$3.6 \\
    \bottomrule
  \end{tabular}
  }
\end{table}

%% file: iclr2025_conference.bib
@String(IJCV = {Int. J. Comput. Vis.})

@String(CVPR= {IEEE Conf. Comput. Vis. Pattern Recog.})

@String(ICLR = {Int. Conf. Learn. Represent.})

@String(AAAI = {AAAI})

@String(IJCV  = {IJCV})

@String(CVPR  = {CVPR})

@String(ICLR  = {ICLR})

@inproceedings{wong2021leveraging,
  title={Leveraging sparse linear layers for debuggable deep networks},
  author={Wong, Eric and Santurkar, Shibani and Madry, Aleksander},
  booktitle={International Conference on Machine Learning},
  pages={11205--11216},
  year={2021},
  organization={PMLR}
}

@article{sawada2022concept,
  title={Concept bottleneck model with additional unsupervised concepts},
  author={Sawada, Yoshihide and Nakamura, Keigo},
  journal={IEEE Access},
  volume={10},
  pages={41758--41765},
  year={2022},
  publisher={IEEE}
}

@misc{farrell_2022, 
title={CUB-200-2011 Segmentations}, DOI={10.22002/D1.20097}, abstractNote={Segmentation masks for the CUB-200-2011 dataset.}, publisher={CaltechDATA}, author={Farrell, Ryan}, year={2022}, month={Apr} 
}

@article{zhang2018top,
  title={Top-down neural attention by excitation backprop},
  author={Zhang, Jianming and Bargal, Sarah Adel and Lin, Zhe and Brandt, Jonathan and Shen, Xiaohui and Sclaroff, Stan},
  journal={International Journal of Computer Vision},
  volume={126},
  number={10},
  pages={1084--1102},
  year={2018},
  publisher={Springer}
}

@inproceedings{ayonrinde2025position,
  title={Position: Interpretability is a Bidirectional Communication Problem},
  author={Ayonrinde, Kola},
  booktitle={ICLR 2025 Workshop on Bidirectional Human-AI Alignment}
}

@article{veale2021demystifying,
  title={Demystifying the Draft EU Artificial Intelligence Act—Analysing the good, the bad, and the unclear elements of the proposed approach},
  author={Veale, Michael and Zuiderveen Borgesius, Frederik},
  journal={Computer Law Review International},
  volume={22},
  number={4},
  pages={97--112},
  year={2021},
  publisher={Verlag Dr. Otto Schmidt}
}

@inproceedings{
dominici2025counterfactual,
title={Counterfactual Concept Bottleneck Models},
author={Gabriele Dominici and Pietro Barbiero and Francesco Giannini and Martin Gjoreski and Giuseppe Marra and Marc Langheinrich},
booktitle={The Thirteenth International Conference on Learning Representations},
year={2025},
url={https://openreview.net/forum?id=w7pMjyjsKN}
}

@inproceedings{
kaiser2025uncertainsam,
title={Uncertain{SAM}: Fast and Efficient Uncertainty Quantification of the Segment Anything Model},
author={Timo Kaiser and Thomas Norrenbrock and Bodo Rosenhahn},
booktitle={Forty-second International Conference on Machine Learning},
year={2025},
url={https://openreview.net/forum?id=G3j3kq7rSC}
}

@article{hewitt2025we,
  title={We Can't Understand AI Using our Existing Vocabulary},
  author={Hewitt, John and Geirhos, Robert and Kim, Been},
  journal={arXiv preprint arXiv:2502.07586},
  year={2025}
}

@article{selvaraju2020grad,
  title={Grad-CAM: visual explanations from deep networks via gradient-based localization},
  author={Selvaraju, Ramprasaath R and Cogswell, Michael and Das, Abhishek and Vedantam, Ramakrishna and Parikh, Devi and Batra, Dhruv},
  journal={International journal of computer vision},
  volume={128},
  pages={336--359},
  year={2020},
  publisher={Springer}
}

@inproceedings{liu2021swin,
  title={Swin transformer: Hierarchical vision transformer using shifted windows},
  author={Liu, Ze and Lin, Yutong and Cao, Yue and Hu, Han and Wei, Yixuan and Zhang, Zheng and Lin, Stephen and Guo, Baining},
  booktitle={Proceedings of the IEEE/CVF international conference on computer vision},
  pages={10012--10022},
  year={2021}
}

@misc{gurobi,
  author = {{Gurobi Optimization, LLC}},
  title = {{Gurobi Optimizer Reference Manual}},
  year = 2023,
  url = "https://www.gurobi.com"
}

@misc{wei2024torchcp,
      title={TorchCP: A Library for Conformal Prediction based on PyTorch}, 
      author={Hongxin Wei and Jianguo Huang},
      year={2024},
      eprint={2402.12683},
      archivePrefix={arXiv},
      primaryClass={cs.LG}
}

@inproceedings{
conformalINfo,
title={An Information Theoretic Perspective on Conformal Prediction},
author={Alvaro Correia and Fabio Valerio Massoli and Christos Louizos and Arash Behboodi},
booktitle={The Thirty-eighth Annual Conference on Neural Information Processing Systems},
year={2024},
url={https://openreview.net/forum?id=gKLgY3m9zj}
}

@inproceedings{papadopoulos2002inductive,
  title={Inductive confidence machines for regression},
  author={Papadopoulos, Harris and Proedrou, Kostas and Vovk, Volodya and Gammerman, Alex},
  booktitle={Machine learning: ECML 2002: 13th European conference on machine learning Helsinki, Finland, August 19--23, 2002 proceedings 13},
  pages={345--356},
  year={2002},
  organization={Springer}
}

@inproceedings{
stutz2022learning,
title={Learning Optimal Conformal Classifiers},
author={David Stutz and Krishnamurthy Dj Dvijotham and Ali Taylan Cemgil and Arnaud Doucet},
booktitle={International Conference on Learning Representations},
year={2022},
url={https://openreview.net/forum?id=t8O-4LKFVx}
}

@article{ding2024class,
  title={Class-conditional conformal prediction with many classes},
  author={Ding, Tiffany and Angelopoulos, Anastasios and Bates, Stephen and Jordan, Michael and Tibshirani, Ryan J},
  journal={Advances in Neural Information Processing Systems},
  volume={36},
  year={2024}
}

@book{vovk2005algorithmic,
  title={Algorithmic learning in a random world},
  author={Vovk, Vladimir and Gammerman, Alexander and Shafer, Glenn},
  volume={29},
  year={2005},
  publisher={Springer}
}

@article{ma2023looks,
  title={This looks like those: Illuminating prototypical concepts using multiple visualizations},
  author={Ma, Chiyu and Zhao, Brandon and Chen, Chaofan and Rudin, Cynthia},
  journal={Advances in Neural Information Processing Systems},
  volume={36},
  pages={39212--39235},
  year={2023}
}

@article{thrpaper,
  title={Least ambiguous set-valued classifiers with bounded error levels},
  author={Sadinle, Mauricio and Lei, Jing and Wasserman, Larry},
  journal={Journal of the American Statistical Association},
  volume={114},
  number={525},
  pages={223--234},
  year={2019},
  publisher={Taylor \& Francis}
}

@article{apspaper,
  title={Classification with valid and adaptive coverage},
  author={Romano, Yaniv and Sesia, Matteo and Candes, Emmanuel},
  journal={Advances in Neural Information Processing Systems},
  volume={33},
  pages={3581--3591},
  year={2020}
}

@article{oikarinen2023label,
  title={Label-free concept bottleneck models},
  author={Oikarinen, Tuomas and Das, Subhro and Nguyen, Lam M and Weng, Tsui-Wei},
  journal={arXiv preprint arXiv:2304.06129},
  year={2023}
}

@inproceedings{
oikarinen2023clipdissect,
title={{CLIP}-Dissect: Automatic Description of Neuron Representations in Deep Vision Networks},
author={Tuomas Oikarinen and Tsui-Wei Weng},
booktitle={The Eleventh International Conference on Learning Representations },
year={2023},
url={https://openreview.net/forum?id=iPWiwWHc1V}
}

@book{templeton2024scaling,
  title={Scaling monosemanticity: Extracting interpretable features from claude 3 sonnet},
  author={Templeton, Adly},
  year={2024},
  publisher={Anthropic}
}

@article{elhage2022toy,
  title={Toy models of superposition},
  author={Elhage, Nelson and Hume, Tristan and Olsson, Catherine and Schiefer, Nicholas and Henighan, Tom and Kravec, Shauna and Hatfield-Dodds, Zac and Lasenby, Robert and Drain, Dawn and Chen, Carol and others},
  journal={arXiv preprint arXiv:2209.10652},
  year={2022}
}

@article{scherlis2022polysemanticity,
  title={Polysemanticity and capacity in neural networks},
  author={Scherlis, Adam and Sachan, Kshitij and Jermyn, Adam S and Benton, Joe and Shlegeris, Buck},
  journal={arXiv preprint arXiv:2210.01892},
  year={2022}
}

@article{lipton1990contrastive,
  title={Contrastive explanation},
  author={Lipton, Peter},
  journal={Royal Institute of Philosophy Supplements},
  volume={27},
  pages={247--266},
  year={1990},
  publisher={Cambridge University Press}
}

@inproceedings{
marconato2022glancenets,
title={GlanceNets: Interpretable, Leak-proof Concept-based Models},
author={Emanuele Marconato and Andrea Passerini and Stefano Teso},
booktitle={Advances in Neural Information Processing Systems},
editor={Alice H. Oh and Alekh Agarwal and Danielle Belgrave and Kyunghyun Cho},
year={2022},
url={https://openreview.net/forum?id=J7zY9j75GoG}
}

@inproceedings{norrenbrocktake,
  title={Take 5: Interpretable Image Classification with a Handful of Features},
  author={Norrenbrock, Thomas and Rudolph, Marco and Rosenhahn, Bodo},
  year={2022},
  booktitle={Progress and Challenges in Building Trustworthy Embodied AI}
}

@article{alvarez2018towards,
  title={Towards robust interpretability with self-explaining neural networks},
  author={Alvarez Melis, David and Jaakkola, Tommi},
  journal={Advances in neural information processing systems},
  volume={31},
  year={2018}
}

@book{molnar2020interpretable,
  title={Interpretable machine learning},
  author={Molnar, Christoph},
  year={2020},
  publisher={Lulu. com}
}

@inproceedings{nauta2021neural,
  title={Neural prototype trees for interpretable fine-grained image recognition},
  author={Nauta, Meike and van Bree, Ron and Seifert, Christin},
  booktitle={Proceedings of the IEEE/CVF Conference on Computer Vision and Pattern Recognition},
  pages={14933--14943},
  year={2021}
}

@inproceedings{nori2019interpretml,
  title={Accurate intelligible models with pairwise interactions},
  author={Lou, Yin and Caruana, Rich and Gehrke, Johannes and Hooker, Giles},
  booktitle={Proceedings of the 19th ACM SIGKDD international conference on Knowledge discovery and data mining},
  pages={623--631},
  year={2013}
}

@InProceedings{pmlr-v202-kalibhat23a,
  title = 	 {Identifying Interpretable Subspaces in Image Representations},
  author =       {Kalibhat, Neha and Bhardwaj, Shweta and Bruss, C. Bayan and Firooz, Hamed and Sanjabi, Maziar and Feizi, Soheil},
  booktitle = 	 {Proceedings of the 40th International Conference on Machine Learning},
  pages = 	 {15623--15638},
  year = 	 {2023},
  editor = 	 {Krause, Andreas and Brunskill, Emma and Cho, Kyunghyun and Engelhardt, Barbara and Sabato, Sivan and Scarlett, Jonathan},
  volume = 	 {202},
  series = 	 {Proceedings of Machine Learning Research},
  month = 	 {23--29 Jul},
  publisher =    {PMLR},
  url = 	 {https://proceedings.mlr.press/v202/kalibhat23a.html}
}

@article{read1993explanatory,
  title={Explanatory coherence in social explanations: A parallel distributed processing account.},
  author={Read, Stephen J and Marcus-Newhall, Amy},
  journal={Journal of Personality and Social Psychology},
  volume={65},
  number={3},
  pages={429},
  year={1993},
  publisher={American Psychological Association}
}

@article{miller2019explanation,
  title={Explanation in artificial intelligence: Insights from the social sciences},
  author={Miller, Tim},
  journal={Artificial intelligence},
  volume={267},
  pages={1--38},
  year={2019},
  publisher={Elsevier}
}

@inproceedings{kim2021hive,
  title={Hive: evaluating the human interpretability of visual explanations},
  author={Kim, Sunnie SY and Meister, Nicole and Ramaswamy, Vikram V and Fong, Ruth and Russakovsky, Olga},
  booktitle={Computer Vision--ECCV 2022: 17th European Conference, Tel Aviv, Israel, October 23--27, 2022, Proceedings, Part XII},
  pages={280--298},
  year={2022},
  organization={Springer}
}

@article{margeloiu2021concept,
  title={Do concept bottleneck models learn as intended?},
  author={Margeloiu, Andrei and Ashman, Matthew and Bhatt, Umang and Chen, Yanzhi and Jamnik, Mateja and Weller, Adrian},
  journal={arXiv preprint arXiv:2105.04289},
  year={2021}
}

@inproceedings{koh2020concept,
  title={Concept bottleneck models},
  author={Koh, Pang Wei and Nguyen, Thao and Tang, Yew Siang and Mussmann, Stephen and Pierson, Emma and Kim, Been and Liang, Percy},
  booktitle={International Conference on Machine Learning},
  pages={5338--5348},
  year={2020},
  organization={PMLR}
}

@misc{hoffmann2021looks,
    title={This Looks Like That... Does it? Shortcomings of Latent Space Prototype Interpretability in Deep Networks}, 
    author={Adrian Hoffmann and Claudio Fanconi and Rahul Rade and Jonas Kohler},
    year={2021},
    booktitle={ICML 2021 Workshop on Theoretic Foundation, Criticism, and Application Trend of Explainable AI},
}

@article{article,
author = {Bibal, Adrien and Lognoul, Michael and Streel, Alexandre and Frénay, Benoît},
year = {2021},
month = {06},
pages = {},
title = {Legal requirements on explainability in machine learning},
volume = {29},
journal = {Artificial Intelligence and Law},
doi = {10.1007/s10506-020-09270-4}
}

@inproceedings{kim2018interpretability,
  title={Interpretability beyond feature attribution: Quantitative testing with concept activation vectors (tcav)},
  author={Kim, Been and Wattenberg, Martin and Gilmer, Justin and Cai, Carrie and Wexler, James and Viegas, Fernanda and others},
  booktitle={International conference on machine learning},
  pages={2668--2677},
  year={2018},
  organization={PMLR}
}

@article{chen2019looks,
  title={This looks like that: deep learning for interpretable image recognition},
  author={Chen, Chaofan and Li, Oscar and Tao, Daniel and Barnett, Alina and Rudin, Cynthia and Su, Jonathan K},
  journal={Advances in neural information processing systems},
  volume={32},
  year={2019}
}

@inproceedings{rymarczyk2021protopshare,
  title={Protopshare: Prototypical parts sharing for similarity discovery in interpretable image classification},
  author={Rymarczyk, Dawid and Struski, {\L}ukasz and Tabor, Jacek and Zieli{\'n}ski, Bartosz},
  booktitle={Proceedings of the 27th ACM SIGKDD Conference on Knowledge Discovery \& Data Mining},
  pages={1420--1430},
  year={2021}
}

@inproceedings{rymarczyk2022interpretable,
  title={Interpretable image classification with differentiable prototypes assignment},
  author={Rymarczyk, Dawid and Struski, {\L}ukasz and G{\'o}rszczak, Micha{\l} and Lewandowska, Koryna and Tabor, Jacek and Zieli{\'n}ski, Bartosz},
  booktitle={European Conference on Computer Vision},
  pages={351--368},
  year={2022},
  organization={Springer}
}

@inproceedings{bau2017network,
  title={Network dissection: Quantifying interpretability of deep visual representations},
  author={Bau, David and Zhou, Bolei and Khosla, Aditya and Oliva, Aude and Torralba, Antonio},
  booktitle={Proceedings of the IEEE conference on computer vision and pattern recognition},
  pages={6541--6549},
  year={2017}
}

@inproceedings{selvaraju2017grad,
  title={Grad-cam: Visual explanations from deep networks via gradient-based localization},
  author={Selvaraju, Ramprasaath R and Cogswell, Michael and Das, Abhishek and Vedantam, Ramakrishna and Parikh, Devi and Batra, Dhruv},
  booktitle={Proceedings of the IEEE international conference on computer vision},
  pages={618--626},
  year={2017}
}

@inproceedings{huang2017densely,
  title={Densely connected convolutional networks},
  author={Huang, Gao and Liu, Zhuang and Van Der Maaten, Laurens and Weinberger, Kilian Q},
  booktitle={Proceedings of the IEEE conference on computer vision and pattern recognition},
  pages={4700--4708},
  year={2017}
}

@inproceedings{straitouri2023improving,
  title={Improving expert predictions with conformal prediction},
  author={Straitouri, Eleni and Wang, Lequn and Okati, Nastaran and Rodriguez, Manuel Gomez},
  booktitle={International Conference on Machine Learning},
  pages={32633--32653},
  year={2023},
  organization={PMLR}
}

@inproceedings{10.5555/3692070.3693971,
author = {Straitouri, Eleni and Rodriguez, Manuel Gomez},
title = {Designing decision support systems using counterfactual prediction sets},
year = {2025},
publisher = {JMLR.org},
booktitle = {Proceedings of the 41st International Conference on Machine Learning},
articleno = {1901},
numpages = {23},
location = {Vienna, Austria},
series = {ICML'24}
}

@article{kuutti2020survey,
  title={A survey of deep learning applications to autonomous vehicle control},
  author={Kuutti, Sampo and Bowden, Richard and Jin, Yaochu and Barber, Phil and Fallah, Saber},
  journal={IEEE Transactions on Intelligent Transportation Systems},
  volume={22},
  number={2},
  pages={712--733},
  year={2020},
  publisher={IEEE}
}

@inproceedings{ahsan2022machine,
  title={Machine-learning-based disease diagnosis: A comprehensive review},
  author={Ahsan, Md Manjurul and Luna, Shahana Akter and Siddique, Zahed},
  booktitle={Healthcare},
  volume={10},
  number={3},
  pages={541},
  year={2022},
  organization={MDPI}
}

@inproceedings{norrenbrock2024q,
  title={Q-senn: Quantized self-explaining neural networks},
  author={Norrenbrock, Thomas and Rudolph, Marco and Rosenhahn, Bodo},
  booktitle={Proceedings of the AAAI Conference on Artificial Intelligence},
  volume={38},
  number={19},
  pages={21482--21491},
  year={2024}
}

@inproceedings{
qpmPaper,
title={{QPM}: Discrete Optimization for Globally Interpretable Image Classification},
author={Thomas Norrenbrock and Timo Kaiser and Sovan Biswas and Ramesh Manuvinakurike and Bodo Rosenhahn},
booktitle={The Thirteenth International Conference on Learning Representations},
year={2025},
url={https://openreview.net/forum?id=GlAeL0I8LX}
}

@article{imagenet15russakovsky,
    Author = {Olga Russakovsky and Jia Deng and Hao Su and Jonathan Krause and Sanjeev Satheesh and Sean Ma and Zhiheng Huang and Andrej Karpathy and Aditya Khosla and Michael Bernstein and Alexander C. Berg and Li Fei-Fei},
    Title = { {ImageNet Large Scale Visual Recognition Challenge} },
    Year = {2015},
    journal   = {International Journal of Computer Vision (IJCV)},
    doi = {10.1007/s11263-015-0816-y},
    volume={115},
    number={3},
    pages={211-252}
}

@INPROCEEDINGS{7298658,
  author={Van Horn, Grant and Branson, Steve and Farrell, Ryan and Haber, Scott and Barry, Jessie and Ipeirotis, Panos and Perona, Pietro and Belongie, Serge},
  booktitle={2015 IEEE Conference on Computer Vision and Pattern Recognition (CVPR)}, 
  title={Building a bird recognition app and large scale dataset with citizen scientists: The fine print in fine-grained dataset collection}, 
  year={2015},
  volume={},
  number={},
  pages={595-604},
  doi={10.1109/CVPR.2015.7298658}}

@inproceedings{szegedy2016rethinking,
  title={Rethinking the inception architecture for computer vision},
  author={Szegedy, Christian and Vanhoucke, Vincent and Ioffe, Sergey and Shlens, Jon and Wojna, Zbigniew},
  booktitle={Proceedings of the IEEE conference on computer vision and pattern recognition},
  pages={2818--2826},
  year={2016}
}

@misc{robustness,
   title={Robustness (Python Library)},
   author={Logan Engstrom and Andrew Ilyas and Hadi Salman and Shibani Santurkar and Dimitris Tsipras},
   year={2019},
   url={https://github.com/MadryLab/robustness}
}

@article{miller1956magical,
  title={The magical number seven, plus or minus two: Some limits on our capacity for processing information.},
  author={Miller, George A},
  journal={Psychological review},
  volume={63},
  number={2},
  pages={81},
  year={1956},
  publisher={American Psychological Association}
}

@misc{jacobgilpytorchcam,
  title={PyTorch library for CAM methods},
  author={Jacob Gildenblat and contributors},
  year={2021},
  publisher={GitHub},
  howpublished={\url{https://github.com/jacobgil/pytorch-grad-cam}},
}

@inproceedings{bohle2022b,
  title={B-cos Networks: Alignment is All We Need for Interpretability},
  author={B{\"o}hle, Moritz and Fritz, Mario and Schiele, Bernt},
  booktitle={Proceedings of the IEEE/CVF Conference on Computer Vision and Pattern Recognition},
  pages={10329--10338},
  year={2022}
}

@inproceedings{he2016deep,
  title={Deep residual learning for image recognition},
  author={He, Kaiming and Zhang, Xiangyu and Ren, Shaoqing and Sun, Jian},
  booktitle={Proceedings of the IEEE conference on computer vision and pattern recognition},
  pages={770--778},
  year={2016}
}

@article{wah2011caltech,
  title={The caltech-ucsd birds-200-2011 dataset},
  author={Wah, Catherine and Branson, Steve and Welinder, Peter and Perona, Pietro and Belongie, Serge},
  year={2011},
  publisher={California Institute of Technology}
}

@inproceedings{StanfordCars,
  title = {3D Object Representations for Fine-Grained Categorization},
  booktitle = {4th International IEEE Workshop on  3D Representation and Recognition (3dRR-13)},
  year = {2013},
  address = {Sydney, Australia},
  author = {Jonathan Krause and Michael Stark and Jia Deng and Li Fei-Fei}
}

@inproceedings{schier2025explainable,
  title={Explainable Reinforcement Learning via Dynamic Mixture Policies},
  author={Schier, Maximilian and Schubert, Frederik and Rosenhahn, Bodo},
  booktitle={2025 IEEE International Conference on Robotics and Automation (ICRA)},
  year={2025}
}

@article{rosenhahn2023optimization,
  title={Optimization of Sparsity-Constrained Neural Networks as a Mixed Integer Linear Program: NN2MILP},
  author={Rosenhahn, Bodo},
  journal={Journal of Optimization Theory and Applications},
  volume={199},
  number={3},
  pages={931--954},
  year={2023},
  publisher={Springer}
}

@inproceedings{glandorf2023hypersparse,
  title={Hypersparse neural networks: Shifting exploration to exploitation through adaptive regularization},
  author={Glandorf, Patrick and Kaiser, Timo and Rosenhahn, Bodo},
  booktitle={Proceedings of the IEEE/CVF International Conference on Computer Vision},
  pages={1234--1243},
  year={2023}
}

@inproceedings{
  glandorf2025p3b,
  title={Pruning by Block Benefit: Exploring the Properties of Vision Transformer Blocks during Domain Adaptation},
  author={Patrick Glandorf and Bodo Rosenhahn},
  booktitle={International Conference on Computer Vision Workshop},
  year={2025}
}

@article{nauta2023pipnet,
  title={PIP-Net: Patch-Based Intuitive Prototypes for Interpretable Image Classification},
  author={Nauta, Meike and Schlötterer, Jörg and van Keulen, Maurice and Seifert, Christin},
  journal={Proceedings of the IEEE/CVF Conference on Computer Vision and Pattern Recognition},
  year={2023},
}

@techreport{FGVCAircraft,
   title         = {Fine-Grained Visual Classification of Aircraft},
   author        = {S. Maji and J. Kannala and E. Rahtu
                    and M. Blaschko and A. Vedaldi},
   year          = {2013},
   archivePrefix = {arXiv},
   eprint        = {1306.5151},
   primaryClass  = "cs-cv",
}

@InProceedings{Fel_2023_CVPR,
    author    = {Fel, Thomas and Picard, Agustin and B\'ethune, Louis and Boissin, Thibaut and Vigouroux, David and Colin, Julien and Cad\`ene, R\'emi and Serre, Thomas},
    title     = {CRAFT: Concept Recursive Activation FacTorization for Explainability},
    booktitle = {Proceedings of the IEEE/CVF Conference on Computer Vision and Pattern Recognition (CVPR)},
    month     = {June},
    year      = {2023},
    pages     = {2711-2721}
}

@inproceedings{
cortes-gomez2025utilitydirected,
title={Utility-Directed Conformal Prediction: A Decision-Aware Framework for Actionable Uncertainty Quantification},
author={Santiago Cortes-Gomez and Carlos Miguel Pati{\~n}o and Yewon Byun and Steven Wu and Eric Horvitz and Bryan Wilder},
booktitle={The Thirteenth International Conference on Learning Representations},
year={2025},
url={https://openreview.net/forum?id=iOMnn1hSBO}
}

@incollection{Pytorch,
title = {PyTorch: An Imperative Style, High-Performance Deep Learning Library},
author = {Paszke, Adam and Gross, Sam and Massa, Francisco and Lerer, Adam and Bradbury, James and Chanan, Gregory and Killeen, Trevor and Lin, Zeming and Gimelshein, Natalia and Antiga, Luca and Desmaison, Alban and Kopf, Andreas and Yang, Edward and DeVito, Zachary and Raison, Martin and Tejani, Alykhan and Chilamkurthy, Sasank and Steiner, Benoit and Fang, Lu and Bai, Junjie and Chintala, Soumith},
booktitle = {Advances in Neural Information Processing Systems 32},
editor = {H. Wallach and H. Larochelle and A. Beygelzimer and F. d\textquotesingle Alch\'{e}-Buc and E. Fox and R. Garnett},
pages = {8024--8035},
year = {2019},
publisher = {Curran Associates, Inc.},
url = {http://papers.neurips.cc/paper/9015-pytorch-an-imperative-style-high-performance-deep-learning-library.pdf}
}
